\documentclass[review,preprint,12pt]{elsarticle}

\usepackage{amssymb}
\usepackage{multicol}

\usepackage{subcaption}
\usepackage{amsmath}
\usepackage{tikz}
\usetikzlibrary{shapes,arrows}
\usetikzlibrary{positioning}
\usetikzlibrary{calc}
\journal{IFAC Journal of Systems and Control}

\begin{document}

\begin{frontmatter}



\title{A Dynamic Model of a Skydiver With Validation in Wind Tunnel and Free Fall}


\author[inst1]{Anna Clarke}
\ead{anna\_shmaglit@yahoo.com}
\affiliation[inst1]{organization={Technion Autonomous Systems Program},%Department and Organization
            addressline={Technion - Israel Institute of Technology}, city={Haifa 32000},
            %postcode={32000}, 
            state={Israel}}

\author[inst2]{Per-Olof Gutman\corref{cor1}}
%\author[inst1,inst2]{Author Three}

\affiliation[inst2]{organization={Faculty of Civil and Environmental Engineering},%Department and Organization
            addressline={Technion - Israel Institute of Technology}, city={Haifa 32000},
            %postcode={32000}, 
            state={Israel}}
            
\ead{peo@technion.ac.il}
\cortext[cor1]{corresponding author}
%\author[First]{Anna Clarke} 
%\author[Second]{Per-Olof Gutman} 

%\address[First]{Technion Autonomous Systems Program, Technion - Israel Institute of Technology, Haifa 32000, Israel (e-mail: anna.clarke@tx.technion.ac.il)}
%\address[Second]{Faculty of Civil and Environmental Engineering, Technion - Israel Institute of Technology, Haifa 32000, Israel (e-mail: peo@technion.ac.il).}

\begin{abstract}
An innovative approach of gaining insight into motor skills involved in human body flight is proposed. %Body flight is the art of maneuvering during the free fall stage of skydiving  which is a rapidly developing sport. 
The key idea is the creation of a model autonomous system capable of virtually performing skydiving maneuvers.% in a virtual way, and towards the goal of turning it into a powerful tool for improving instruction methods. 

A dynamic skydiver model and simulator is developed, comprising biomechanical, aerodynamic, and kinematic models, dynamic equations of motion, and a virtual reality environment.
%Limb positions and orientations, and resulting body  angular position and velocity coordinates are measured  in skydiving experiments in a vertical wind tunnel and in free fall. 
Limb relative orientations, and resulting inertial body  angular position and velocity are measured  in skydiving experiments in a vertical wind tunnel and in free fall.
These experimental data are compared with corresponding simulation data to tune and verify the model for basic skydiving maneuvers. The model is further extended to reconstruct advanced aerial maneuvers, such as transitions between stable equilibria. The experimental data are used to estimate skydiver's conscious inputs as a function of time, via an Unscented Kalman Filter modified for this purpose.  

\end{abstract}

%%Graphical abstract was commented by peo

%%Research highlights commented out by peo


\end{frontmatter}

%\linenumbers

%\section{Introduction} see lit_survey
%\vspace{5cm}
\section{Introduction}
\label{S:1}
Aerodynamics of a human body in free-fall is a relatively unresearched area. The closest problem to a skydiver motion in free-fall is aircraft stability and control analysis. There is a vast knowledge base for aircraft stability derivatives, flow field, maneuvers, bifurcations,  etc. However, there is nothing comparable for the free-falling parachutist. The reason is that free-fall maneuvers may be very complex and have large amplitude, the parachutist is a bluff
 body, is not rigid, and has multiple control surfaces and redundant degrees-of-freedom. Nonlinear simulation of the parachutist dynamics becomes impractical due to the unreasonably large amount of computational time required to generate Computational Fluid Dynamics (CFD) data over a required range of body poses. Thus, the few existing works on simulation and analysis of skydiving focus on small perturbation linearized dynamics about the equilibrium in a basic free-fall pose (belly-to-earth) assuming the body is rigid. Even under these assumptions, the flow field computation for bluff bodies with large dynamic roughness at high Reynolds numbers is still beyond the capabilities of state-of-the-art CFD software. 

There are only two teams of researchers \cite{skydive_myers2009FAST,skydive_dietz2011cfd}, both sponsored by the US Army, that succeeded to develop a tool for free-fall analysis and simulation.
The motivation for developing such a tool was an attempt to gain a better insight into the problem of pilot chute hesitation in the MC-4 military parachute system (when the pilot chute becomes trapped in the recirculation region of the wake behind the parachutist, preventing the main parachute deployment). Both toolkits consist of a Modeling Tool that allows users to create and pose figures of a skydiver, CFD software that provides flow solutions about a skydiver's body, and a Simulation of the skydiver's dynamics. The main challenge was to find an appropriate CFD method that would address all the involved physical phenomena: massively separated unsteady turbulent 3D flow, large roughness, flapping clothing, suction and blowing through the fabric. The first research group \cite{skydive_myers2009FAST} used a Large Eddy Simulation (LES)  approach, which probably is the most suitable  for dealing with the above challenges, but is very computationally demanding - requiring hundreds of CPU hours to analyze a single parachutist pose. For this reason the resulting Free-Fall Simulator only creates a longitudinal dynamics analysis linearized around a single body configuration and orientation defined by the user in advance. The second research group \cite{skydive_dietz2011cfd} used a Reynolds Averaged Navier Stokes (RANS) approach, which has considerable shortcomings when used to analyze a skydiver in free-fall, but is the only computationally practical option for a simulator that aims to allow the user to move the skydiver's limbs during simulation. In order to support such a simulation, a parachutist body was modeled by 11 body segments each with 3 degrees of rotational freedom, and it was assumed that for a certain baseline pose the forces and moments contribution of the individual body components can be considered separately. This assumption was inspired by an analogy of flying with an aircraft: the control coefficients due to ailerons and elevators deflection are derived from an analysis of the effect of these actions on a representative aircraft model. The experiences of practicing skydivers also confirms this assumption \cite{skydive_wings}: skydivers perceive each body component as a separate control surface and focus on the angle at which each limb is presented to the relative wind. The simulator described in \cite{skydive_dietz2011cfd} allows the user to alter limb positions during the free-fall simulation and shows the resulting maneuvers of a skydiver. It still takes a few days to produce a CFD database for a specific skydiver and his equipment configuration, and the simulation is limited to small maneuvers in a belly-to-earth position and at terminal velocity. 

In addition to the above research  there are a couple of student projects concerning free-falling skydiver modeling, stability analysis, and aerodynamic testing. 
In \cite{skydive_sit} a skydiver model consisting of 15 simple geometric shapes was developed and positioned in a sit-fly pose. Dynamic movement equations were formulated assuming the body was rigid. The aerodynamic coefficients were estimated in a wind tunnel test carried out for three configurations of a skydiver's manikin. Similar work is described in \cite{skydive_cardona2011measurement}, only the manikin is positioned in the tunnel in belly-to-earth postures. The measured dimensionless aerodynamic coefficients were used to analyze stability and control effectiveness of different postures. This analysis was compared to basic maneuvers performed in a full-scale vertical wind tunnel and during actual skydives using wearable inertial sensors, magnetometers, and miniature GPS loggers. Full-scale aerodynamic forces and moments were estimated by applying a Kalman filter. In \cite{skydive_robson2010longitudinal} the authors conducted longitudinal stability analysis of a skydiver wearing a jet-powered wingsuit. They used a rigid body assumption and linearized equations of motion similar to those of a glider, thus  simplifying the problem considered in \cite{skydive_sit} and \cite{skydive_cardona2011measurement} even further.  Actual flight data was used to verify the theoretical results, using off-the-shelf Xsens Technologies \cite{sensor_roetenberg2009xsens} wearable sensors suite: accelerometers, gyros, magnetometers, GPS, and barometers. 

Summarizing all the above mentioned research projects: simulations of skydiver's dynamics have been developed to work offline and in the proximity of one of the stable equilibrium positions. Whereas, providing online solutions during unstable transitions and high amplitude maneuvers have not been attempted to date. This challenge is addressed in the present research. 

Hence, in this paper an analytic method to model the skydiver aerodynamics is proposed. The model makes it possible to simulate  a skydiver continuously altering the body posture as a function of time,  and thus simulating a wide range of skydiving maneuvers.

Limb relative orientations, and resulting inertial body linear and angular velocities were measured  in skydiving experiments in a vertical wind tunnel and in free fall, using the above  mentioned Xsens Technologies wearable motion capture suit. The main objective was to verify that all basic so called  Relative Work (RW) maneuvers, discussed in Section~\ref{sec:valid}, and described by linear and angular velocity profiles, are reconstructed by the simulator. This means that if the measured sequence of body postures is fed to the simulator, the virtual skydiver will produce velocity profiles similar to the ones measured in an actual experiment. Other equally important objectives are to gain insight into the skydiver model, and understand   the influence of its various parameters on the resulting motion. Hence, some experimental data were used to tune and estimate some of the model parameters.

The paper is organized as follows. In Section~\ref{sec:expmod}, the experimental procedure,  skydiving participants, and set-up are described. Section~\ref{sec:model} details the skydiver model formulation, including the biomechanical, kinematic, and aerodynamic models yielding the dynamic equations of motion. The initial experimental model validation, including the measurement procedure and parameter tuning is found in Section~\ref{sec:valid} while the data processing is covered in Section~\ref{sec:dat}. Damping moment coefficients connected to the skydiver's conscious control are profitably modeled  as time-varying inputs in Sections~\ref{sec:conscious} and~\ref{sec:advanced_maneuvers_ukf},   with the Modified Unscented Kalman Filter developed for their estimation is described in detail in Section~\ref{sec:modifiedukf}. Further experimental results, also for advanced maneuvers, compared with the skydiver model outputs are reported in Section~\ref{sec:experimentalresults}. Insights into how the skydiver body produces the damping moment coefficients can be learnt in Section~\ref{sec:advanced_maneuvers_mech}. Section~\ref{sec:discussion} concludes the paper with a summary and discussion about the goodness of fit, whether the model is parsimonious, possible uses of the model, and possible avenues for further investigations.

This paper is based on a part of \cite{Clarke:2021}. A preliminary version of the skydiver model was presented at a conference, \cite{clarkeGutman2017}.

\section{Experimental modalities} \label{sec:expmod} %Participants and Procedures
Before describing the skydiver model and its development, an overview of the experiments are given in this section. Further details are reported together with the experiment results in Section~\ref{sec:valid}.

The initial validation of the skydiver model involved performing seven basic maneuvers in a belly-to-earth orientation by five participants. Of the seven maneuvers, listed just below, the first five were initially performed in a vertical wind tunnel, and then repeated in free fall. Vertical wind tunnels are widely used skydiving simulators (diameter around 4-4.5 m, height 10-15 m), where the air is blown upwards at around 60 m/sec which is the average terminal vertical velocity of skydivers in a belly-to-earth pose in free fall. The human body floats inside the tunnel replicating the physics of body flight.
\begin{enumerate}
    \item \textit{Neutral Fall}: falling straight down while maintaining the Neutral Pose, and performing continuous adjustments following from the interaction with the airflow
    \item \textit{Rotations}: right and left 360 degrees turns
    \item \textit{Side Slides}: moving to the left and to the right while preserving the initial heading
    \item \textit{Longitudinal Movement}: moving forwards and backwards while preserving the initial heading
    \item \textit{Vertical Movement}: changing the fall rate (the terminal velocity), i.e. falling faster and slower than the nominal fall rate of a given skydiver, which matches his Neutral Pose. In the wind tunnel this maneuver looks like going upwards and downwards in the flying chamber. 
    \item \textit{Barrel Roll}: rolling around the longitudinal body axis while starting and finishing in a belly-to-earth orientation, and preserving the initial heading
    \item \textit{Back Loop}: rotating around the body frontal axis while starting and finishing in a belly-to-earth orientation, and preserving the initial heading
\end{enumerate}
The two last maneuvers were performed only in free-fall, as they require considerable space, especially when executed by skydivers who are less experienced in the wind tunnel flying. 

Each experiment involved performing one maneuver by one skydiver, while (s)he was instructed to repeat the maneuver as many times as possible during the available free-fall time or wind tunnel session, typically 30-50 seconds and 1-2 minutes, respectively. The collected data included the anthropometric measurements (body shape, size, and weight), a sequence of postures (expressed by 23 quaternions describing the relative orientation of body segments at each instant of time) recorded by the Xsens body movement tracking system \cite{sensor_roetenberg2009xsens}, and a video of the maneuvers. Most of the collected data is publicly available via \cite{dataset}, along with Matlab code for the initial data processing and graphical representation.

%\vspace{0.2cm}
%\textbf{\underline{Experimental Procedure:}}
%\vspace{0.2cm}
\subsection{Experimental Procedure}
\begin{enumerate}
    \item Giving introduction and instructions to the participant
    \item Collecting anthropometric measurements, and keeping a record of the participant's specific equipment configuration: type of jumpsuit, helmet, shoes, parachute container, and weight belt  
    \item Providing the participant with the Xsens suit, and performing the calibration procedure defined by Xsens: standing still and walking for a few seconds.
    \item Videoing how the participant performs the predefined maneuver
    \item Post processing the Xsens data by the means of the Xsens software (MVN Studio), while providing it with the specific anthropometric parameters
    \item Running the skydiving simulation, while providing it with a sequence of recorded postures as the time varying input, and anthropometric parameters, and gear specifications as simulation parameters.
\end{enumerate}

The result of each experiment is the comparison between the skydiver's inertial motion recorded via Xsens and reconstructed by the skydiving simulation, as presented in Section \ref{sec:valid}. 

%\vspace{0.2cm}
%\textbf{\underline{Description of the Participants:}}
%\vspace{0.2cm}
\subsection{Description of the Participants}
The participants were informed of the aims and procedures of the experiments which were approved by the Technion Institutional Review Board and Human Subjects Protection, and signed an informed consent form.

The three female and two male skydivers participating in the model validation experiments were between the age of 35-55, height 160-175 cm, weight 50-85 kg, and possessed different body shapes, such that the three major somatotypes (ectomorph, mesomorph, and endomorph) were represented.
Their skill level varied from Elite Competitor to an Intermediate Student. The gear variations included three types of jumpsuits, three types of helmets, and an option of wearing a weight belt. 

The jumpsuits used in the experiments were professional skydiving suits particular to the following disciplines: RW (Relative Work), free fly, and wind tunnel flying. The RW suits have booties that add a very large aerodynamic surface attached to the shins. The wind tunnel suits have the most tight fit, leaving no flapping cloth on any of the body segments. The free-fly suits used in the experiments were more baggy than the wind tunnel suits and made from a different material. 

In this way, it was possible to verify that the skydiving simulation is sufficiently generic and can accurately reconstruct maneuvers performed during a number of representative study cases. In each study case the following factors were different:
\begin{enumerate}
    \item participant's body type, height, and weight
    \item participant's equipment configuration (suit, helmet, weight belt, parachute container)
    \item participant's skill level
    \item environment (free-fall/ wind tunnel)
\end{enumerate}
Notice that the first two factors constitute parameters in the Biomechanical model which is a part of the skydiver model, as explained in Section \ref{sec:model}. Each of the 16 body segments in the modes is configured according to the measurements collected for each participant. The head segment is configured according to the helmet type. The weight belt and the parachute container (if present) are taken into account when defining the torso segments. The booties (in the case of RW suit) are represented by a triangular surface attached to the legs segments. 

The last two factors are testing the skydiver model robustness, since they are not represented directly. There are six tuning parameters related to the Aerodynamic model: damping moment coefficients, maximum moment coefficient, and maximum drag coefficients. The damping moment coefficients represent the skydiver's muscle tension, and are assumed constant during a given type of maneuver. It was however discovered that this assumption is appropriate for skilled skydivers only, whereas for truthful reconstruction of maneuvers performed by students these parameters must be estimated as a function of time, as shown in Section \ref{sec:advanced_maneuvers_ukf}.   

The maximum values of the  moment/drag coefficients reflect the amount of turbulence in the environment (airflow in wind tunnels is expected to be more turbulent), but also depend on the amount of flapping clothing, which increases the drag. In professional RW suits different types of material are used for different suit segments. For this reason the model has an option of assigning different maximum moment/drag coefficients to different body segments. However, for the investigated study cases reported here this was not necessary, since the skydiving simulation was sufficiently robust, as explained in Section \ref{sec:aero_params}.

%\textbf{\textit{Experimental Set-Up}}
\subsection{Experimental Set-Up}
\begin{figure}
    \centering
     \includegraphics[width=\textwidth]{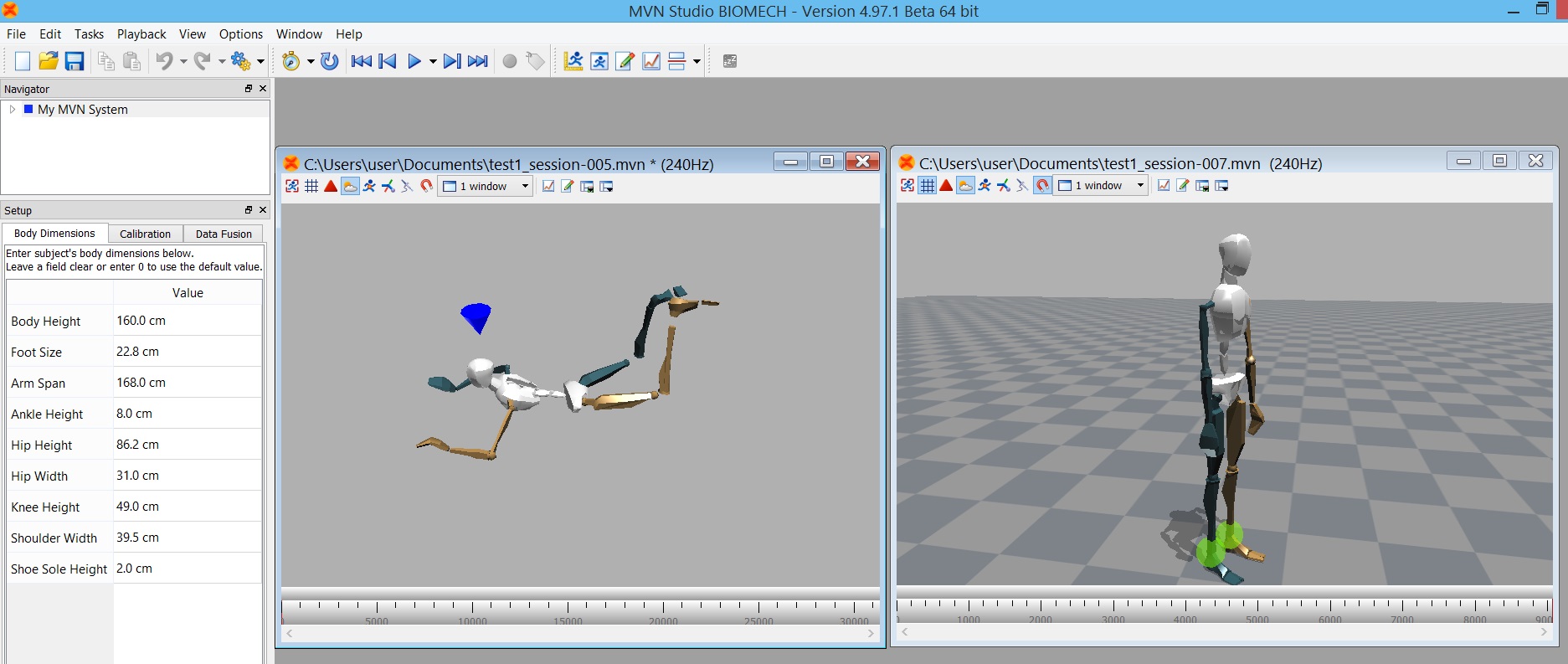}
    \caption{MVN Studio: Xsens software for displaying and post-processing the posture measurements.}
    \label{fig:studio}
\end{figure}
As mentioned, an Xsens body movement tracking system \cite{sensor_roetenberg2009xsens} was chosen for getting an accurate measurement of the full body posture as a function of time. Xsens provides a suit with 16 miniature inertial sensors that are fixed at strategic locations on the body. Each unit includes a 3D accelerometer, 3D rate gyroscope, 3D magnetometer, and a barometer. The participants wore the motion tracking suit underneath their conventional skydiving gear. The suit has a battery and a small computer located on the back that is not restricting the skydiving-specific movements. The Xsens hardware includes appropriate pre-sampling filters for the output recorded at 240 Hz. Each measurement set at a given sampling instant includes the orientation of 23 body segments (pelvis, four spine segments, neck, head, shoulders, upper arms, forearms, hands, upper legs, lower legs, feet, toes) relative to the inertial frame, expressed by quaternions. The measurements accuracy is less than 5 degrees RMS of the dominant joint angles \cite{schepers2018xsens}. The processing of Xsens data, as well as the transformation between the skydiver model and the Xsens body model, is given in Section \ref{sec:dat}. 

 All performed experiments, in the wind tunnel as well as in free fall, have video documentation and data files that include all the measurements. The data files can be viewed via the provided Xsens software, see Figure \ref{fig:studio}.

\section{Skydiver Model Formulation} \label{sec:model}
\subsection{Dynamic Simulation of Body Flight}

In this section we describe the dynamic simulation of the human body in free-fall. It receives as input a sequence of body postures and computes position, orientation, and linear and angular velocities of the skydiver model in a 3D world. The inputted postures can be recorded, transmitted in real-time, as well as synthetically generated, or even inputted via a keyboard using a GUI (Graphical User Interface), specifically designed for this purpose. The simulation is implemented in Matlab. It has a continuous graphical output, using the Virtual Reality Modeling Language (VRML), which shows a figure of a skydiver in its current pose moving through the sky. The sky has a grid of equally spaced half-transparent dots, so that the skydiver’s maneuvers can be easily perceived by the viewer. The modules comprising our skydiving simulator are described below.

\begin{figure}[ht!]
\begin{center}
\includegraphics[width=1\textwidth]{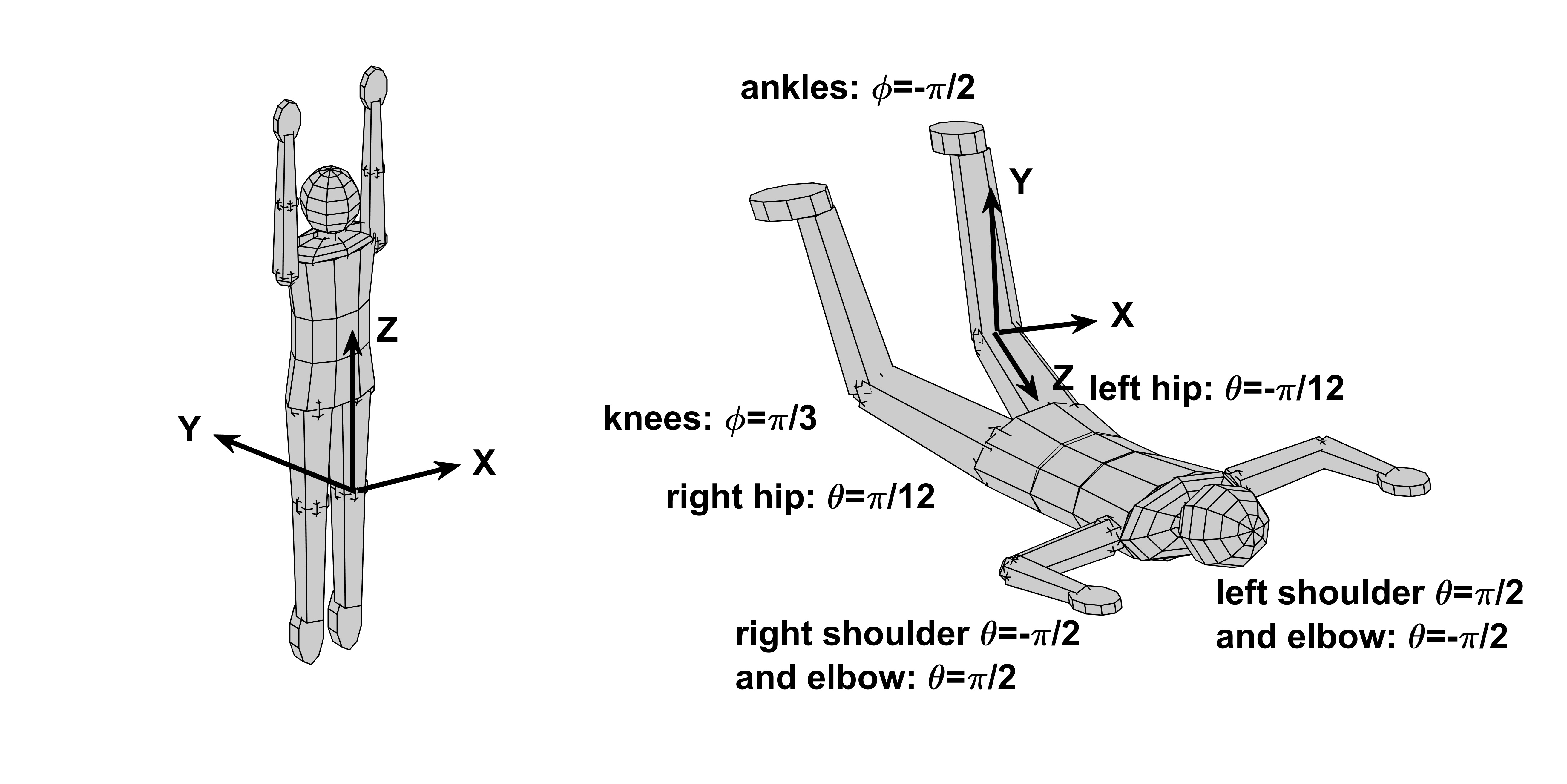} 
\caption{Pose defined by all zero Euler angles (left), and a standard neutral pose (right). For each limb the Euler angles $\psi,\theta,\phi$ are rotations around $Z, Y, X$ axis of the coordinate system attached to the parent limb.}
\label{fig:coord_axis}
\end{center}
\end{figure}  
%\nomenclature{$\psi,\theta,\phi$}{[rad] Euler angles: yaw, pitch, and roll, respectively}

\subsection{The Biomechanical Model}
\textbf{The Biomechanical Model} 
defines the skydiver body configuration by  means of 16 rigid segments (pelvis, abdomen, thorax, head, upper arms, forearms, hands, upper legs, lower legs, and feet), and 15 joints (lumbar, thorax, neck, shoulders, elbows, wrists, hips, knees, ankles). The relative orientation of connected segments is defined by three Euler angles. The sequence of rotation is shown in Figure \ref{fig:coord_axis}. The body model has 45 DOFs which are expressed by a set of rotation quaternions for convenience of computations.   
The body segments are modeled by  means of simple geometrical shapes, such as truncated cones. This way, the local center of gravity and principal moments of inertia of each segment can be computed according to conventional Body Segment Parameters (BSP) equations \cite{bsd_kwon1994kwon3d}.  The overall center of gravity and inertia tensor are computed taking into account the masses of the individual segments, and utilizing the similarity transformation and the parallel axis theorem. 

The center of gravity and inertia tensor of each body segment is initially expressed in the Local Frame, and then transformed into the Body Frame, arbitrarily chosen to be attached to the pelvic joint.
For example, Eq. (\ref{eq:chain}) shows the transformation chain for the right hand, needed for computing its center of gravity in the Body frame $\vec{r}_{cg_{Hand}}^{\,Body}$. 
\begin{equation}  \label{eq:chain}
\begin{split}
\vec{q}_{Body}^{\,Hand}=&\vec{q}_{Forearm}^{\,Hand} \otimes 
 \vec{q}_{Upperarm}^{\,Forearm} \otimes 
 \vec{q}_{Thorax}^{\,Upperarm} \otimes \vec{q}_{Abdomen}^{\,Thorax} \otimes 
 \vec{q}_{Body}^{\,Abdomen} \\ 
 \vec{d}_{Hand}^{\,IN \, Body}=&\vec{q}_{Body}^{\,Abdomen} \otimes ( \vec{d}_{Thorax}^{\,IN \, Abdomen} + \vec{q}_{Abdomen}^{\,Thorax} \otimes ( \vec{d}_{Upperarm}^{\,IN \, Thorax}
 + ... \\ &\vec{q}_{Thorax}^{\,Upperarm}
 \otimes ( \vec{d}_{Forearm}^{\,IN  \, Upperarm} + \vec{q}_{Upperarm}^{\,Forearm} \otimes  \vec{d}_{Hand}^{\,IN \, Forearm}
)
)
)\\
 \vec{r}_{cg_{Hand}}^{\,Body}=&\vec{d}_{Hand}^{\,IN \, Body}+\vec{q}_{Body}^{\,Hand} \otimes \vec{r}_{cg_{Hand}}^{\,local} 
 \end{split}
\end{equation}
where $\vec{q}^{\,Frame 1}_{Frame 2}$ is the rotation quaternion from $Frame1$ to $Frame2$, $\vec{d}_{Limb}^{\,IN \, Frame}$ is the origin of the coordinate system attached to $Limb$ expressed in coordinate system $Frame$, and the symbol $\otimes$ denotes a rotation of a vector by a quaternion.

%\nomenclature{$\vec{r}_{cg_{Limb}}^{\,Frame}$}{center of gravity of \textit{Limb} expressed in coordinate system $Frame$}
%\nomenclature{$\vec{q}^{\,Frame 1}_{Frame 2}$}{rotation quaternion from $Frame1$ to $Frame2$}
%\nomenclature{$\vec{d}_{Limb}^{\,IN \, Frame}$}{origin of the coordinate system attached to $Limb$ expressed in coordinate system $Frame$}

Conducting similar computations for each segment makes it possible to obtain the body instantaneous center of gravity $\vec{r}_{cg}$ and inertia tensor $I$, as shown in Eqs. \eqref{eq:cg1}, \eqref{eq:cg2}.
\begin{equation} \label{eq:cg1}
     \vec{r}_{cg}=\frac{\sum_{i=1}^{N_{Limbs}} \vec{r}_{cg_i}^{\,Body} m_i}{\sum_{i=1}^{N_{Limbs}} m_i}
\end{equation}
\begin{equation}  \label{eq:cg2}
\begin{split}
 I=&\sum_{i=1}^{N_{Limbs}}  \underbrace{ DCM_{Body}^{Limb_i} I_{local_i} (DCM_{Body}^{Limb_i})^T}_{the\ similarity\ transformation}  + ... \\ & \sum_{i=1}^{N_{Limbs}} \underbrace{ \left[ \begin{smallmatrix}  \Delta Y^2+\Delta Z^2 & -\Delta X \Delta Y & -\Delta X \Delta Z \\
-\Delta X \Delta Y & \Delta X^2+\Delta Z^2 & -\Delta Y \Delta Z \\  -\Delta X \Delta Z & -\Delta Y\Delta Z &  \Delta X^2+\Delta Y^2 
\end{smallmatrix} \right]_i m_i}_{parallel\ axis\ theorem} 
\end{split}
\end{equation}
where $m_i$ is the mass of the Limb $i$,  $[\Delta X ,\ \Delta Y ,\ \Delta Z]^T$ is the distance between local and global center of gravity, $I_{local_i}$  is the inertia tensor of the Limb $i$ in its Local Frame, and $DCM$ is the  direction cosine matrix computed from the relevant quaternion.
%\nomenclature{$\vec{r}_{cg}=\left[ \begin{matrix} x_{cg} & y_{cg} & z_{cg} \end{matrix} \right]^T, \, \vec{r_{cg}^i}$}{[m] center of gravity of the body and of the Limb $i$ in Body Frame}
%\nomenclature{$I$}{[kg $\cdot$ m$^2$] body inertia tensor}
%\nomenclature{$[\Delta X ,\ \Delta Y ,\ \Delta Z]^T$}{[m] distance between local and global center of gravity}
%\nomenclature{$m, \, m_i$}{[kg] total body mass, mass of the Limb $i$}
% \newline

The model has to be provided with a set of parameters, expressing body size, shape, and weight of the skydiver under investigation. The model can also take into account the size, weight, and shape of specific equipment: the rig, weight-belt, helmet, and booties, i.e. material connecting toes and knees, and thus creating a large aerodynamic surface.
%\newline

\subsection{The Kinematic Model}
\textbf{The Kinematic Model} computes the body inertial orientation, body angles of attack and sideslip ($\alpha$, $\beta$ [rad]), and the local angles of attack, sideslip, and roll of each segment $i$ relative to the airflow ($\alpha_i$, $\beta_i$, $\gamma_i$ [rad]). 
%\nomenclature{$X_I,Y_I,Z_I$}{axes of the inertial coordinate system: north, west, up, respectively}
%\nomenclature{$X_{Body},Y_{Body},Z_{Body}$}{axes of the Body coordinate system: left, dorsoventral, anteroposterior, respectively}

%\begin{wrapfigure}{l}{0.45\textwidth}
%\vspace{-20pt}
\begin{figure}
\begin{center}
\includegraphics[width=0.45\textwidth]{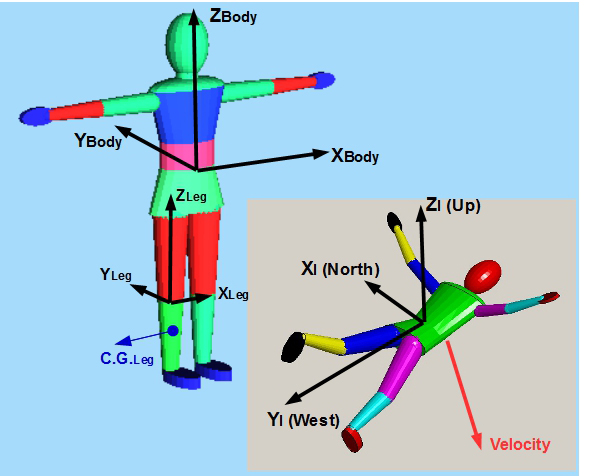} 
\caption{Coordinate systems} 
\label{fig:coord}
\end{center}
\end{figure}
%\vspace{-18pt}
%\end{wrapfigure}  
Additionally, the Kinematic Model contains transformations between all the coordinate systems (see Figure \ref{fig:coord}) involved in further computations: 

\begin{enumerate}
  \item \textit{Inertial Frame}: defined as North, West, Up
  \item \textit{Body Frame}: coincides with Inertial Frame when standing and facing east
  \item \textit{Global Wind Frame}:  transformation from Body to Wind frame includes two Euler rotations: $\alpha$ about X-axis, and then $-\beta$ about Y-axis
  \item \textit{Local Limb Frame}: has origin in the relevant joint, see e.g. Leg Local Frame shown in Figure \ref{fig:coord}
  \item \textit{Local Wind Frame}: transformation from Local Limb to Wind frame defines local angles of attack, sideslip, and roll.
\end{enumerate}

%\nomenclature{$\alpha, \, \beta$}{[rad] angles of attack and sideslip of the body}

The body coordinate system was chosen such that it is consistent with standard anatomical terms of location. Axes: X - left-right, Y - dorsoventral, Z - anteroposterior. Planes: YZ - sagittal plane, XY - transverse (horizontal) plane, XZ - coronal (frontal) plane. The Kinematic Model includes the following computations:
\begin{enumerate}
\item The Euler angles $[ \begin{matrix}\psi & \theta & \phi \end{matrix}]$ defining the transformation from Inertial to Body frame: Eq. (\ref{eq:euler})
\item Body angles of attack $\alpha$ and sideslip $\beta$: Eq. (\ref{eq:alfabeta})
\item Local angles of attack, sideslip, and roll $\alpha_i$, $\beta_i$, $\gamma_i$ of each limb relative to the airflow: Eq. (\ref{eq:alfabetaI})
\end{enumerate}

\begin{equation}  \label{eq:euler}
\begin{split}
&\vec{q}_{Body}^{\,I}=\left[ \begin{matrix} q_0&q_1&q_2&q_3 \end{matrix} \right]^T \\ 
&\psi=atan \frac{2(q_0q_3+q_1q_2)}{1-2(q_2^2+q_3^2)} \qquad
\theta=asin 2(q_0q_2-q_1q_3) \\
&\phi=atan \frac{2(q_0q_1+q_3q_2)}{1-2(q_2^2+q_1^2)}\\
\end{split}
\end{equation}
\begin{equation}  \label{eq:alfabeta}
\alpha=-atan \frac{V}{W}
\quad \beta=-asin\frac{U}{\sqrt{(U^2+V^2+W^2)}}
\end{equation}
\begin{equation}  \label{eq:alfabetaI}
\begin{gathered}
\vec{q}_{W}^{\,B}=\left[ \begin{matrix} cos \frac{\alpha}{2}&sin  \frac{\alpha}{2}&0&0 \end{matrix} \right]^T \otimes \left[ \begin{matrix} cos  \frac{-\beta}{2}&0&sin  \frac{-\beta}{2}&0 \end{matrix} \right]^T \\
\vec{q}_{Wind}^{\,Limb_i}=\vec{q}_{Body}^{\,Limb_i} \otimes \vec{q}_{W}^{\,B}=\left[ \begin{matrix} q_0&q_1&q_2&q_3 \end{matrix} \right]^T \\
 \alpha_i=atan \frac{2(q_0q_3-q_1q_2)}{q_0^2-q_1^2-q_2^2+q_3^2} \qquad
 \beta_i=-asin 2(q_0q_2+q_1q_3) \\
\gamma_i=atan \frac{2(q_0q_3-q_1q_2)}{q_0^2-q_2^2-q_3^2+q_1^2}
\end{gathered}
\end{equation}
where $\vec{q}_{W}^{\,B}$ and $\vec{q}_{Wind}^{\,Limb_i}$ are rotation quaternions from Body to Wind and from i-th Limb to Wind frames, respectively.
%\newline
\subsection{Dynamic Equations of Motion}
\textbf{The Dynamic Equations of Motion}, derived following the Newton-Euler method, provide six equations: 3D forces ($\vec{F}$ [N]) and moments ($\vec{M}$ [Nm]):
\begin{equation}
\begin{split}
    \vec{F} &=m\dot{\vec{V}} +  m\dot{\vec{\Omega}} \times \vec{r_{cg}} + m\vec{\Omega} \times \dot{\vec{r_{cg}}}+ \vec{\Omega} \times \left( m\vec{V}+m\vec{\Omega} \times \vec{r_{cg}} \right) \\
    \vec{M} &=I\dot{\vec{\Omega}} +\dot{I}\vec{\Omega}+  \vec{r_{cg}} \times m\dot{\vec{V}} + \dot{\vec{r_{cg}}} \times  m\vec{V} +\vec{V} \times \left( m\vec{\Omega} \times \vec{r_{cg}} \right) + \\
 & + \vec{\Omega} \times (I\vec{\Omega}) 
+ \vec{\Omega} \times (\vec{r_{cg}}  \times m\vec{V} )
\end{split}
\label{eq:eq_F_M}
\end{equation}
where $\vec{V}=\left[ \begin{matrix} U & V & W \end{matrix} \right]^T$ [m/s] - linear velocity, $\vec{\Omega}=\left[ \begin{matrix} P&Q&R \end{matrix} \right]^T$ [rad/s] - angular velocity, $I$ [kg $\cdot$ m$^2$] - inertia tensor, $\vec{r_{cg}}$ [m] - center of gravity, all expressed in Body frame, and $m$ [kg] - total body mass.

The center of gravity vector and inertia tensor are functions of time and their derivatives are not neglected (as in the case of aircraft models), since the body posture is continuously changing.
The decomposition of the vector form given in Eq. \eqref{eq:eq_F_M} into six equations involving the derivatives of $U, V, W$ and $P,Q,R$ is given below.

The equations for 3D forces and moments are derived from taking the derivatives of the linear and angular momentum, respectively, as summarized in Eqs. \eqref{eq:force1}, \eqref{eq:force2} and Eqs. \eqref{eq:moment1}, \eqref{eq:moment2}. 
\begin{equation}  \label{eq:force1}
\begin{split}
&\text{Linear momentum: } \vec{p}= m\vec{V}+m\vec{\Omega} \times \vec{r_{cg}} \\
& \vec{F}=\dot{\vec{p}} + \vec{\Omega} \times \vec{p} \\
& \vec{F}=m\dot{\vec{V}} +  m\dot{\vec{\Omega}} \times \vec{r_{cg}} + m\vec{\Omega} \times \dot{\vec{r_{cg}}}+ \vec{\Omega} \times \left( m\vec{V}+m\vec{\Omega} \times \vec{r_{cg}} \right) 
\end{split}
\end{equation}
where $ \vec{F}=\left[ \begin{matrix} X & Y & Z \end{matrix} \right]^T $
\begin{equation}  \label{eq:force2}
\begin{split}                                                   
& \frac{X}{m}=\dot{U}+\dot{Q}z_{cg}-\dot{R}y_{cg}+Q\dot{z_{cg}}-R\dot{y_{cg}}+QW-RV+... \\ & \qquad Q(Py_{cg}-Qx_{cg})- R(Rx_{cg}-Pz_{cg})                                                    \\
& \frac{Y}{m}=\dot{V}+\dot{R}x_{cg}-\dot{P}z_{cg}+R\dot{x_{cg}}-P\dot{z_{cg}}+RU-PW+... \\ & \qquad R(Qz_{cg}-Py_{cg})-P(Py_{cg}-Qx_{cg})
\\
&  \frac{Z}{m}=  \dot{W}+ 
 \dot{P}y_{cg}-\dot{Q}x_{cg}+ 
  P\dot{y_{cg}}-Q\dot{x_{cg}}+ PV-QU+ ... \\ & \qquad
  P(Rx_{cg}-Pz_{cg})- Q(Qz_{cg}-Ry_{cg})
 \end{split}
\end{equation}
%\nomenclature{$\vec{F}=\left[ \begin{matrix} X & Y & Z \end{matrix} \right]^T$}{[N] overall force acting on the body}
%\nomenclature{$\left[ \begin{matrix} X & Y & Z \end{matrix} \right]^T$}{components of the force $\vec{F}$}
%\nomenclature{$\vec{M}=\left[ \begin{matrix} L & M & N \end{matrix} \right]^T$}{[Nm] overall moment acting on the body}
%\nomenclature{$\left[ \begin{matrix} L & M & N \end{matrix} \right]^T$}{components of the moment $\vec{M}$}

\begin{equation}  \label{eq:moment1}
\begin{split}
& \text{Angular momentum: } \vec{l}=I\vec{\Omega}+\vec{r_{cg}} \times m\vec{V} \\
& \vec{M}=\dot{\vec{l}}+\vec{\Omega} \times \vec{l}+\vec{V} \times \left( m\vec{\Omega} \times \vec{r_{cg}} \right) \\
& \vec{M}=I\dot{\vec{\Omega}} +\dot{I}\vec{\Omega}+  \vec{r_{cg}} \times m\dot{\vec{V}} + \dot{\vec{r_{cg}}} \times  m\vec{V} +\vec{V} \times \left( m\vec{\Omega} \times \vec{r_{cg}} \right) 
 + ... \\ & \qquad \vec{\Omega} \times (I\vec{\Omega}) +
\vec{\Omega} \times (\vec{r_{cg}}  \times m\vec{V} )
\end{split}
\end{equation}
where $  \vec{M}=\left[ \begin{matrix} L&M&N \end{matrix} \right]^T $
\begin{equation}  \label{eq:moment2}
\begin{split}
&L =\dot{I_{xx}}P-\dot{I_{xy}}Q-\dot{I_{xz}}R+I_{xx}\dot{P}- I_{xy}(\dot{Q}-RP)-I_{xz}(\dot{R}+PQ) ... \\ & \qquad
-I_{yz}(Q^2-R^2)+(I_{zz}
 -I_{yy})QR  +m\dot{y_{cg}}W -m\dot{z_{cg}}V+... \\ & \qquad my_{cg}(\dot{W}+VP-QU)+mz_{cg}(-\dot{V}+WP-RU) \\
&M =\dot{I_{yy}}Q-\dot{I_{yx}}P-\dot{I_{yz}}R+I_{yy}\dot{Q}-I_{xy}(\dot{P}+QR)-I_{yz}(\dot{R}-PQ)... \\ & \qquad
-I_{xz}(R^2-P^2)+(I_{xx}
 -I_{zz})PR +m\dot{z_{cg}}U-m\dot{x_{cg}}W+... \\ & \qquad mx_{cg}(-\dot{W}+QU-PV)+mz_{cg}(\dot{U}+WQ-RV) \\
&N =\dot{I_{zz}}R-\dot{I_{xz}}P-\dot{I_{yz}}Q+I_{zz}\dot{R}+I_{xz}(-\dot{P}+QR)-I_{yz}(\dot{Q}+PR)... \\ & \qquad
+I_{xy}(Q^2-P^2)+(I_{yy}
 -I_{xx})PQ  +m\dot{x_{cg}}V-m\dot{y_{cg}}U+... \\ & \qquad mx_{cg}(\dot{V}+UR-PW)+my_{cg}(-\dot{U}+VR-QW)
 \end{split}
\end{equation}
where $\vec{p}$ and $\vec{l}$ are the linear and angular momentum, expressed in Body frame, and the remaining symbols are consistent with definitions given in Eq. \eqref{eq:eq_F_M}.

%\nomenclature{$\vec{p}$}{[kg $\cdot$ m/s] linear momentum}
%\nomenclature{$\vec{l}$}{[kg $\cdot$ m$^2$/s] angular momentum}

%\nomenclature{$\vec{V}=\left[ \begin{matrix} U & V & W \end{matrix} \right]^T$}{[m/s] body linear velocity}
%\nomenclature{$\left[ \begin{matrix} U & V & W \end{matrix} \right]^T$}{components of the linear velocity $\vec{V}$}
%\nomenclature{$\vec{\Omega}=\left[ \begin{matrix} P & Q & R \end{matrix} \right]^T$}{[rad/s] body angular velocity}
%\nomenclature{$\left[ \begin{matrix} P & Q & R \end{matrix} \right]^T$}{components of the angular velocity $\vec{\Omega}$}
%\nomenclature{$\left[ \begin{matrix} x_{cg} & y_{cg} & z_{cg} \end{matrix} \right]^T$}{components of the center of gravity vector $\vec{r_{cg}}$}
The inertial orientation of the skydiver is represented by a rotation quaternion ($\vec{q}_{Body}^{\,I}$), and propagated in time as in Eq. (\ref{eq:quat}).

\begin{equation}  \label{eq:quat}
\quad \dot{\vec{q}}_{Body}^{\,I}=0.5\left[ \begin{matrix} 0&-P&-Q&-R\\P&0&R&-Q\\Q&-R&0&P\\R&Q&-P&0 \end{matrix} \right]\vec{q}_{Body}^{\,I}
\end{equation}

%\nomenclature{$\alpha_i,\beta_i,\gamma_i$}{[rad] angles of attack, sideslip, and roll of the Limb $i$}
%\newline
\subsection{The Aerodynamic Model}
\textbf{The Aerodynamic Model} is formulated as a sum of forces and moments acting on each individual segment, modeled similar to aircraft aerodynamics - proportional to velocity squared and to the area exposed to the airflow, see Eq. \eqref{eq:limbf}, \eqref{eq:limbm} . The forces and moments computed for each segment $i$ depend on the local angles of attack, sideslip, and roll, and three aerodynamic coefficients: $Cl_i^{max}$, $Cd_i^{max}$, $Cm_i^{max}$, which were tuned in experiments.
The total aerodynamic force and moment together with the gravity forces (given in Eqs. (\ref{eq:sum_f}), (\ref{eq:sum_m})) are substituted into the equations of motion Eq. (\ref{eq:eq_F_M}).

Additionally, the aerodynamic model can receive two types of skydiver conscious input, represented by damping moment and input moment coefficients. These inputs are significant when simulating advanced skydiving maneuvers, as in Section \ref{sec:advanced_maneuvers_ukf}, while for all basic maneuvers in a belly-to-earth orientation the input moments can be neglected (and thus don't appear in Eq. \eqref{eq:limbf}, \eqref{eq:limbm}) and the damping moment coefficients ($\vec{Cm}_{damp}$ in Eq. \eqref{eq:sum_m}) can be tuned as constant values.

\begin{equation}  \label{eq:sum_f}
\begin{split}
&\vec{F}=  \displaystyle\sum_{i=1}^{N_{limbs}}\vec{F_a}^i + \vec{q}_{Body}^{\,I} \otimes\left[ \begin{matrix} 0\\0\\-mg \end{matrix} \right]
\end{split}
\end{equation}
\begin{equation}  \label{eq:sum_m}
\begin{split}
 \vec{M}=& \displaystyle\sum_{i=1}^{N_{limbs}}\left(\vec{r}_{cg}^{\,i} \times \vec{F_a}^i + \vec{M_a}^i \right)+\vec{r_{cg}} \times \left(\vec{q}_{Body}^{\,I} \otimes\left[ \begin{matrix} 0\\0\\-mg \end{matrix} \right]  \right)...\\& -0.005\rho A  \left\|\vec{V}\right\|^2 H \vec{Cm}_{damp} \vec{\Omega} 
\end{split}
\end{equation}
where $g$ [m/s$^2$] is  the gravity constant, $\rho$ [kg/m$^3$] is  air density, $N_{limbs}$ is the number of body segments, $A$ [m$^2$] is the overall area exposed to the airflow, $H$ [m] is the skydiver's height, $\vec{Cm}_{damp}$ [-] are the estimated damping moment coefficients, and $\vec{F_a}^i, \vec{M_a}^i, \vec{r}_{cg}^{\,i} $ are the
aerodynamic force and moment vectors acting on limb $i$ and its center of gravity expressed in the Body frame.
%The symbol $\otimes$ denotes a rotation of a vector by a quaternion, and 
The operator $\left\|\cdot\right\|^2$ denotes the square norm of a vector.

%\nomenclature{$g$}{[m/s$^2$] gravity constant}
%\nomenclature{$\rho$}{[kg/m$^3$] air density}
%\nomenclature{$N_{limbs}$}{number of body segments}
%\nomenclature{$A$}{[m$^2$] overall body area exposed to the airflow}
%\nomenclature{$H$}{[m] skydiver's height}
%\nomenclature{$\vec{Cm}_{damp}$}{[-] damping moment  coefficients}
%\nomenclature{$\vec{F_a}^i$}{[N] aerodynamic force acting on the Limb $i$}
%\nomenclature{$\vec{M_a}^i, \vec{M}^i_{atot}$}{[Nm] aerodynamic moment: acting on the Limb $i$, and total (summing up moments acting on all limbs) }

The last term in Eq. (\ref{eq:sum_m}) is the aerodynamic damping moment that occurs due to the changes in the orientation of the local wind vector with rotation rates across the skydiver. This moment can be measured in wind tunnel experiments, as in \cite{skydive_myers2009FAST}, and is expected to be small. For this reason the term is scaled by 0.01: for the convenience of tuning of the damping moment coefficients. Their tuning and role in skydiving maneuvers are further discussed in Section \ref{sec:aero_params}.

%\nomenclature{$\vec{L}_i$}{[N] aerodynamic force perpendicular to the airflow acting on the Limb $i$ expressed in Body Frame}
%\nomenclature{$\vec{D}_i, \vec{D}$}{[N] aerodynamic force parallel to the airflow expressed in Body Frame, acting on the Limb $i$, and on the body in total}
%\nomenclature{$A_i$}{[m$^2$] limb characteristic area (local $XZ$ plane)}
%\nomenclature{$Area_i$}{[m$^2$] total limb area exposed to the airflow}
%\nomenclature{$(Cl_{\alpha})_i, \, (Cl_{\beta})_i$}{[-] aerodynamic coefficients for the Limb $i$ expressing the dependence of forces on the angles of attack and sideslip}
%\nomenclature{$(Cm_{\alpha})_i, \, (Cm_{\beta})_i$}{[-] aerodynamic coefficients for the Limb $i$ expressing the dependence of moments on the angles of attack and sideslip}
%\nomenclature{$ Cl_i ^{max},  Cd_i ^{max}, Cm_i^{max}$}{[-] maximum values of aerodynamic coefficients for the Limb $i$}

The two components of the aerodynamic force acting on an individual limb $i$:  perpendicular $\vec{L_i}$ [N] and parallel $\vec{D_i}$ [N] to the local wind direction are given in Eq. (\ref{eq:limbf}). The aerodynamic moment acting on a body segment which is angled relative to the air flow can be approximated as shown in Eq. (\ref{eq:limbm}).
\begin{equation}  \label{eq:limbf}
\begin{split}
& \vec{F_a}^i = \vec{L_i} +\vec{D_i} \\
& \vec{L_i}= \vec{q}_{B}^{\,W} \otimes \left( \left[ \begin{matrix} cos \frac{\gamma_i}2\\0\\0\\sin \frac{\gamma_i}2 \end{matrix} \right] \otimes \left[ \begin{matrix} 0.5\rho A_i \left\|\vec{V}\right\|^2(Cl_{\beta})_i\\ 0.5\rho A_i  \left\|\vec{V}\right\|^2(Cl_{\alpha})_i\\0 \end{matrix} \right] \right) \\
& \vec{D_i}=\vec{q}_{B}^{\,W} \otimes \left[ \begin{matrix} 0& 0& 0.5\rho Area_i \left\|\vec{V}\right\|^2Cd_i^{max} \end{matrix} \right]^T 
\end{split}
\end{equation}
where $A_i$ [m$^2$] is the limb characteristic area (in local $xz$ plane), $Area_i$ [m$^2$] is the total limb area exposed to the airflow approximated according to Eq. (\ref{eq:area}), $\vec{q}_{B}^{\,W}$ is the rotation quaternion from Wind Frame to Body Frame, and $(Cl_{\alpha})_i, \, (Cl_{\beta})_i$ [-] are the aerodynamic coefficients approximated according to Eq.  (\ref{eq:coeffCl}), while $ Cl_i ^{max}$, $ Cd_i^{max}$ [-] are the estimated drag coefficients.
\begin{equation}  \label{eq:area}
\begin{split}
Area_i=max  \begin{cases} A_i^{xz}\left| cos\beta_i sin \alpha_i\right| \\ A_i^{xy}\left| cos\beta_i cos\alpha_i\right| \\ A_i^{yz}\left| sin\beta_i \right| \end{cases}
\end{split}
\end{equation}
\begin{equation}  \label{eq:coeffCl}
\begin{split}
&(Cl_{\alpha})_i = Cl_i ^{max}sin (2\alpha_i) \quad
(Cl_{\beta})_i = Cl_i ^{max}sin (2\beta_i) 
\end{split}
\end{equation}
\begin{equation}  \label{eq:limbm}
\begin{split}
&M_i=\vec{q}_{B}^{\,W} \otimes \left( \left[ \begin{matrix} cos \frac{\gamma_i}2\\0\\0\\sin \frac{\gamma_i}2 \end{matrix} \right] \otimes \left[ \begin{matrix} \frac{1}{2}\rho A_i \left\|\vec{V}\right\|^2l_i(Cm_{\alpha})_i\\ \frac{1}{2}\rho A_i  \left\|\vec{V}\right\|^2l_i(Cm_{\beta})_i\\0 \end{matrix} \right] \right)
\end{split}
\end{equation}
where $l_i$ [m] is limb characteristic length, and the moment coefficients $(Cm_{\alpha})_i, (Cm_{\beta})_i$ [-] can be approximated according to Eq. (\ref{eq:cm}) , while $Cm_i ^{max}$ [-] are their maximal values, estimated from experiments. 
\begin{equation}  \label{eq:cm}
\begin{split}
&(Cm_{\alpha})_i = -Cm_i ^{max}sin (2\alpha_i) \quad
(Cm_{\beta})_i = -Cm_i ^{max}sin (2\beta_i)
\end{split}
\end{equation}

The aerodynamic coefficients $Cl_i^{max}$, $Cd_i^{max}$, $Cm_i^{max}$ can be different for each body segment $i$ in case that a skydiver is wearing a suit composed of different materials, such that its type and the amount of flapping cloth used for each segment is optimized for a specific skydiving discipline. If the test jumpers wear general purpose suits, as in the experiments described next, these coefficients can be assumed equal for all segments. Hence, only three aerodynamic coefficients have to be estimated:  $Cl^{max}$, $Cd^{max}$, $Cm^{max}$.

\subsection{Reformulation of the Aerodynamic Model}
The aerodynamic model requires calculation of the angles of attack and sideslip of every body segment. These angles are derived from the quaternion describing the rotation between the local segment and Wind systems. This quaternion is decomposed into three Euler angles with a pre-set order. In the synthetic simulation the final positions of end-effectors are normally reached by assigning specific joints rotation angles. Conveniently, the minimal values for all angles are chosen. Thus, the relation for calculation of aerodynamic angles, given in Eq. \eqref{eq:alfabetaI}, holds true. However, the actual measurements expressed by quaternions and thereafter decomposed into Euler angles can produce erroneous results due to the non-uniqueness of Euler angles. For example, it is easy to see that $DCM(\phi,\theta,\psi)=DCM(\phi+\pi,-\theta,\psi-\pi)$. From experiments with the Xsens body measurement system it was noticed that the same position of end-effectors in a synthetic simulation and in reality produced different values for the angles of attack and sideslip of some limbs. This required to formulate the Aerodynamic Model in a more generic way. 

The angles of attack and sideslip can be computed from the unity vector $\vec{v}_i$ in the direction of velocity in a local Limb $i$ frame, defined in Eq. \eqref{eq:unityVloclimb}.
\begin{equation} \label{eq:unityVloclimb}
\vec{v}_i=\vec{q}^{\, Body}_{Limb_i}\otimes \vec{u} 
\end{equation}
where $\vec{u}$ is the unity vector in the direction of velocity in the Body frame (attached to the pelvis). The drag force acting on Limb $i$ in the perpendicular direction to the local wind $\vec{L}_i$ is defined by two components: $\vec{L_{\alpha i}}$ and $\vec{L_{\beta i}}$. The magnitude of these components is calculated as before (see Eqs. \eqref{eq:limbf}, \eqref{eq:coeffCl}) from the $\alpha_i$ and $\beta_i$ angles which are now defined as:
\begin{subequations}
\begin{equation}
|\alpha_i|=acos(\vec{v}_{iz}) 
\end{equation}
\begin{equation}
|\beta_i|=acos(\vec{v}_{ix})
\end{equation}
\end{subequations}
The direction of the two components of $\vec{L}_i$ is defined as follows: $\vec{L_{\alpha i}}$ lies in the plane defined by vectors $\vec{v}_i$ and $[0 \quad 0 \quad 1]^T$, perpendicular to $\vec{v}_i$, and the cosine of an angle between $\vec{v}_i$ and $[0 \quad 0 \quad 1]^T$ and an angle between $\vec{L_{\alpha i}}$ and $[0 \quad 0 \quad 1]^T$ have the same sign. The component $\vec{L_{\beta i}}$ lies in the plane defined by vectors $\vec{v}_i$ and $[1 \quad 0 \quad 0]^T$, perpendicular to $\vec{v}_i$, and the angle between $\vec{L_{\beta i}}$ and $[0 \quad 1 \quad 0]^T$ is less than 90 degrees. These observations can be formalized as:
\begin{subequations}
\begin{equation}
\vec{n_{\alpha i}}=\vec{v}_i \times \left( \vec{v}_i \times \left[ \begin{matrix} 0 \\ 0 \\ 1 \end{matrix} \right] \right)
\end{equation}
\begin{equation}
\vec{n_{\beta i}}=\vec{v}_i \times \left( \vec{v}_i \times \left[ \begin{matrix} 1 \\ 0 \\ 0 \end{matrix} \right] \right)
\end{equation}
\end{subequations}
while the signs of $\vec{n_{\alpha i}}$ and $\vec{n_{\beta i}}$ are reversed if the following conditions hold:
\begin{subequations}
\begin{equation}
\begin{gathered}
condition_{\alpha i}=\left( |\alpha_i|>\pi/2 \quad and \quad acos(\vec{n_{\alpha i}}_z)<\pi/2 \right) \quad or \\ \quad \left( |\alpha_i|<\pi/2 \quad and \quad acos(\vec{n_{\alpha i}}_z)>\pi/2 \right)
\end{gathered}
\end{equation}
\begin{equation}
condition_{\beta i}=\vec{n_{\alpha i}}_y<0
\end{equation}
\end{subequations}
The direction of matching aerodynamic moments is defined as:
\begin{subequations}
\begin{equation}
\vec{m_{\alpha i}}=\vec{v}_i \times \vec{n_{\alpha i}}
\end{equation}
\begin{equation}
\vec{m_{\beta i}}=\vec{v}_i \times \vec{n_{\beta i}}
\end{equation}
\end{subequations}
Multiplying the magnitudes of the Drag and Moment components by the normalized direction vectors provides the Drag and Moment expressed in the local limb frame. Subsequently, the local Drags and Moments associated with each limb are transferred to the Body frame and summarized.   
%\nomenclature{$\vec{L_{\alpha i}}, \vec{L_{\beta i}}$}{[N] components of the drag force acting on Limb $i$ in the perpendicular direction to the local wind $\vec{L}_i$}
%\nomenclature{$\vec{u}, \, \vec{v}_i$}{[-] unity vector in the direction of velocity in the Body Frame, unity vector in the direction of velocity in a local Limb $i$ Frame}
%\nomenclature{$\vec{n_{\alpha i}}, \, \vec{n_{\beta i}}, \, \vec{m_{\alpha i}}, \, \vec{m_{\beta i}}$}{[-] unity vectors in the direction of forces $\vec{L_{\alpha i}}, \vec{L_{\beta i}}$ and their matching moments}
%\newline

%\begin{wrapfigure}{l}{0.45\textwidth}
%\vspace{-20pt}
\begin{figure}
\begin{center}
\includegraphics[width=0.45\textwidth]{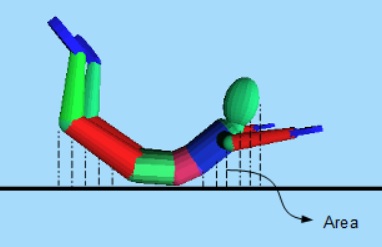} 
\caption{Area calculated by the Arch Model} 
\label{fig:archA}
\end{center}
\end{figure}
%\vspace{-18pt}
%\end{wrapfigure}
\subsection{The Arch Model}
\begin{figure}[!htb] 
 \centering
    \includegraphics[width=0.6\textwidth]{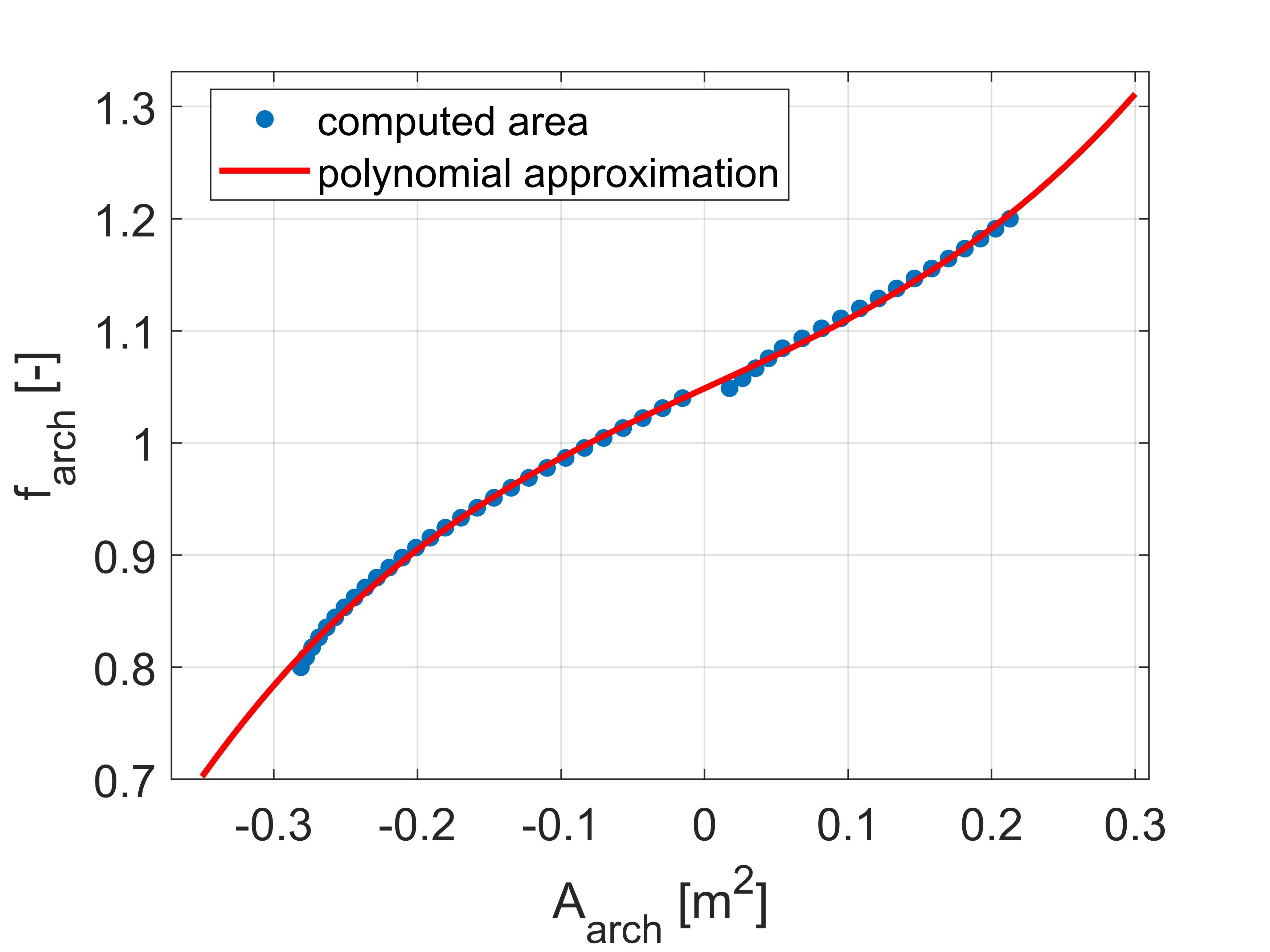}
    \caption{
       Arch model approximated according to Eq. \eqref{eq:arch_area}.
    } \label{fig:arch_model}
   %\vspace{-20pt} 
\end{figure}

\textbf{The Arch Model}
 reflects the fact that arching decreases aerodynamic drag since the airflow is distributed around the body more efficiently, consequently the skydiver falls faster. De-arching increases the drag by making the airflow path more complex, causing the skydiver to fall slower. Skydivers feel it as a turbulent flow that accumulates in the abdomen area. Thus, the drag force $\vec{D}$ acting on the body parallel to the wind direction (expressed in Body Frame) can be updated according to Eq. (\ref{eq:arch_eq}), before $\vec{D}$ is substituted into Eq. (\ref{eq:sum_f}).
\begin{equation}  \label{eq:arch_eq}
\begin{split}
&\vec{D}= \left( \displaystyle\sum_{i=1}^{N_{limbs}}\vec{D}_i \right)f_{arch}
\end{split}
\end{equation}
where $\vec{D}_i$ is the drag force, acting on individual body segments calculated in Eq. (\ref{eq:limbf}), and $f_{arch}$ is a factor that reflects the amount of arching in a body posture. This amount can be expressed in terms of \textit{Area}, shown in Figure \ref{fig:archA}. The value of this area (noted by $A_{arch}$) is signed: negative for the arch pose (as in Figure \ref{fig:archA}), positive for de-arching. The relation between $A_{arch}$ and $f_{arch}$ is approximated according to Eq. (\ref{eq:arch_area}), which provides a good fit to our arching model, see Figure \ref{fig:arch_model}. 
\begin{equation}  \label{eq:arch_area}
\begin{split}
&f_{arch}=2.8828A_{arch}^3-0.0039A_{arch}^2+0.5281A_{arch}+1.055
\end{split}
\end{equation}
\begin{figure}[!htb] 
 \centering
    \includegraphics[width=0.7\textwidth]{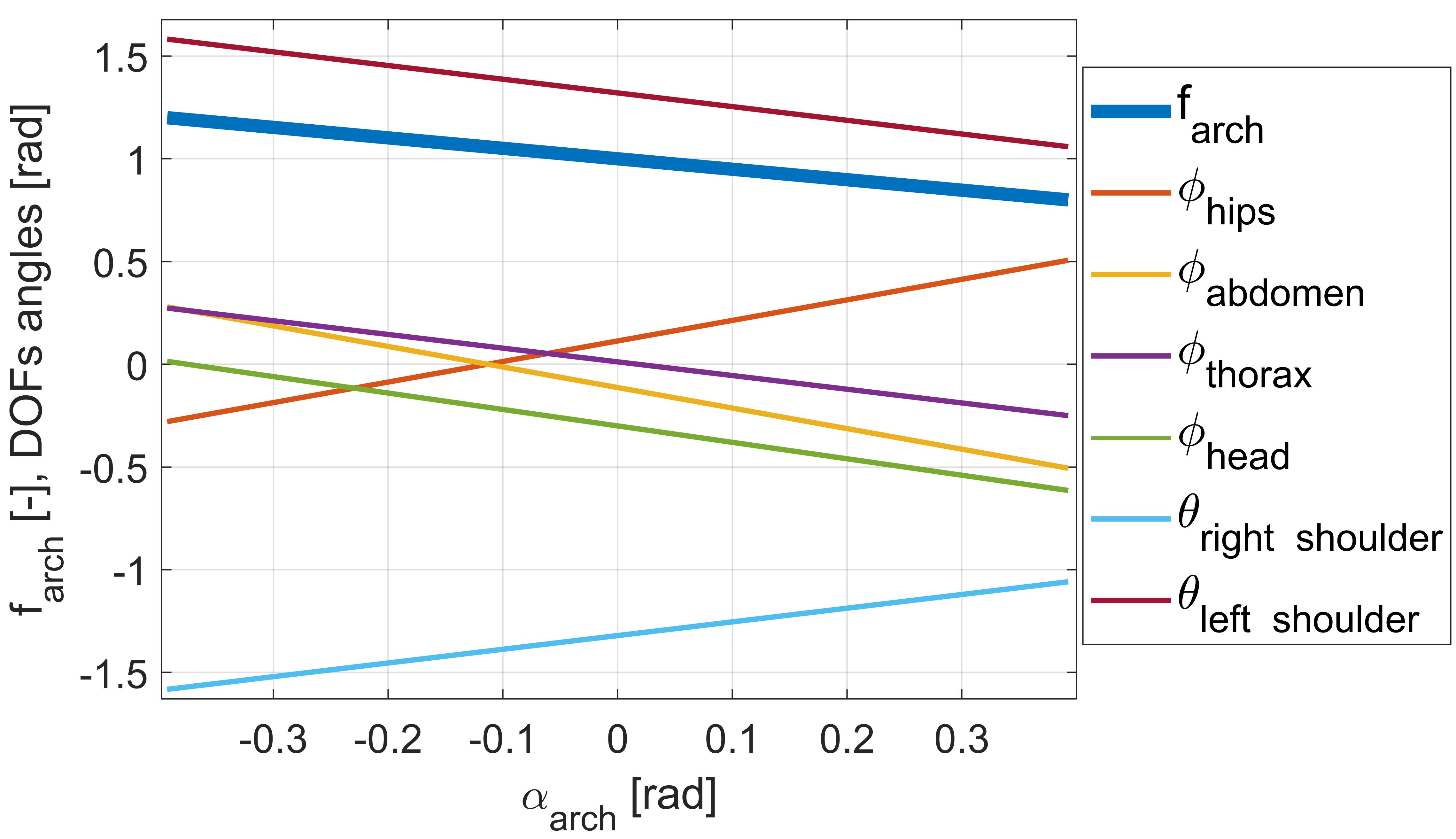}
    \caption{
       Body Degrees-of-Freedom as a function of an arch pattern angle.
    } \label{fig:arch_dof}
   %\vspace{-20pt} 
\end{figure}
%\nomenclature{$f_{arch}$}{[-] factor reflecting the amount of body arch}
%\nomenclature{$A_{arch}$}{[m$^2$] virtual area that reflects arching: area contained between the horizon and the body projection into sagittal plane}
The factor $f_{arch}$ defines the arching movement pattern $\alpha_{arch}$ that controls seven DOFs, as shown in Eq. \eqref{eq:arch_pat}. This pattern  was constructed from video observations of typical skydiving postures. Notice from Eq. \eqref{eq:arch_pat} and Figure \ref{fig:arch_dof} that the relation between $f_{arch}$ and $\alpha_{arch}$, which follows from Eq. \eqref{eq:arch_area}, is linear. Also, when the posture is neutral (not arching or de-arching) and $\alpha_{arch}=0$, notice that $f_{arch}=1$.
\begin{equation}
    \begin{gathered}
    f_{arch}=1-\frac{0.2}{\pi/8}\alpha_{arch} \\
    \phi_{hips} = (\phi_{hips})_{Neutral}-\alpha_{arch} \\
    \phi_{abdomen} = (\phi_{abdomen})_{Neutral}+\alpha_{arch} \\
    \phi_{thorax} = (\phi_{thorax})_{Neutral}+\frac{2}{3}\alpha_{arch} \\
    \phi_{head} = (\phi_{head})_{Neutral}+\frac{4}{5}\alpha_{arch} \\
    \theta_{right \, shoulder} = (\theta_{right \, shoulder})_{Neutral}-\frac{2}{3}\alpha_{arch} \\
    \theta_{left \, shoulder} = (\theta_{left \, shoulder})_{Neutral}+\frac{2}{3}\alpha_{arch} 
    \end{gathered}
    \label{eq:arch_pat}
\end{equation}
where the Euler angles defining the body DOFs are defined as in Figure \ref{fig:coord_axis}.

\section{Experimental Validation and Model Tuning} \label{sec:valid}
\subsection{Model Validation in Experiments}
The skydiving simulator output was experimentally verified in two stages: 
\begin{enumerate}
    \item \textbf{Basic Maneuvers:} in a belly-to-earth pose (turning right/ left, moving forwards/ backwards, sliding sideways, changing fall rate) were performed by different skydivers in a wind tunnel and in free-fall. The test-jumpers were wearing variable types of equipment and possessed different levels of skill: intermediate, advanced, and elite.  
     \item \textbf{Advanced Maneuvers:} A test-jumper in free-fall performed a variety of transitions between belly-to-earth and back-to-earth orientations (flips, rolls, and layouts). Additionally, in both orientations the skydiver performed angle-flying: extreme horizontal movement achieved by maintaining a steep angle between the torso and the airflow.    
\end{enumerate}
The first stage was aimed at tuning the six parameters related to the aerodynamic model: the aerodynamic coefficients ($Cl^{max}$, $Cd^{max}$, $Cm^{max}$), and roll, pitch, and yaw damping moment coefficients ($\vec{Cm}_{damp}$). The coefficients were chosen such that all the basic maneuvers are closely reconstructed by the simulator, fed by the recorded posture sequences. It was important to verify that the model is sufficiently generic, i.e. can represent different body types/ equipment configurations, and is not overly sensitive to the jumpers' skill level and variations that might occur in the free-fall environment. For this purpose, the experiments took place in both existing types of wind tunnels (non-recirculating and recirculating), and various drop-zones. This way, the free-fall experiments were performed in  hot and cold climates, in humid and dry places, sunny, cloudy, windy weather, and exiting from various types of aircraft. 

The second stage of experimental validation was concerned with extending the skydiving simulator to advanced aerial maneuvers, performed in other than belly-to-earth orientations. During such maneuvers the roll, pitch, and yaw damping moment coefficients can no longer be assumed constant. Additionally, the skydivers consciously impose pressure on the airflow in order to maintain the advanced body orientations and transition between equilibria. These additional inputs and the damping moment coefficients were estimated by the means of an Unscented Kalman Filter (UKF), \cite{julier1997new}, at every time instant during the experiments, while the aerodynamic coefficients were set to constants tuned from the experiments of the first stage (basic maneuvers). Next, the measured body posture sequences and the estimated skydiver's conscious inputs were fed into the skydiving simulator and it was verified that the skydiver's inertial motion were accurately reconstructed. 

 The RMS errors in angular and linear (horizontal and vertical) velocities were 0.15 rad/s, 0.45 m/s (horizontal), and 1.5 m/s (vertical), while the velocities amplitudes were 7 rad/s, 15 m/s, and 65 m/s, respectively.

\subsection{Basic Maneuver Validation: Aerial Rotations}%{Measurement procedure}
\begin{figure}
    \centering
     \includegraphics[width=\textwidth]{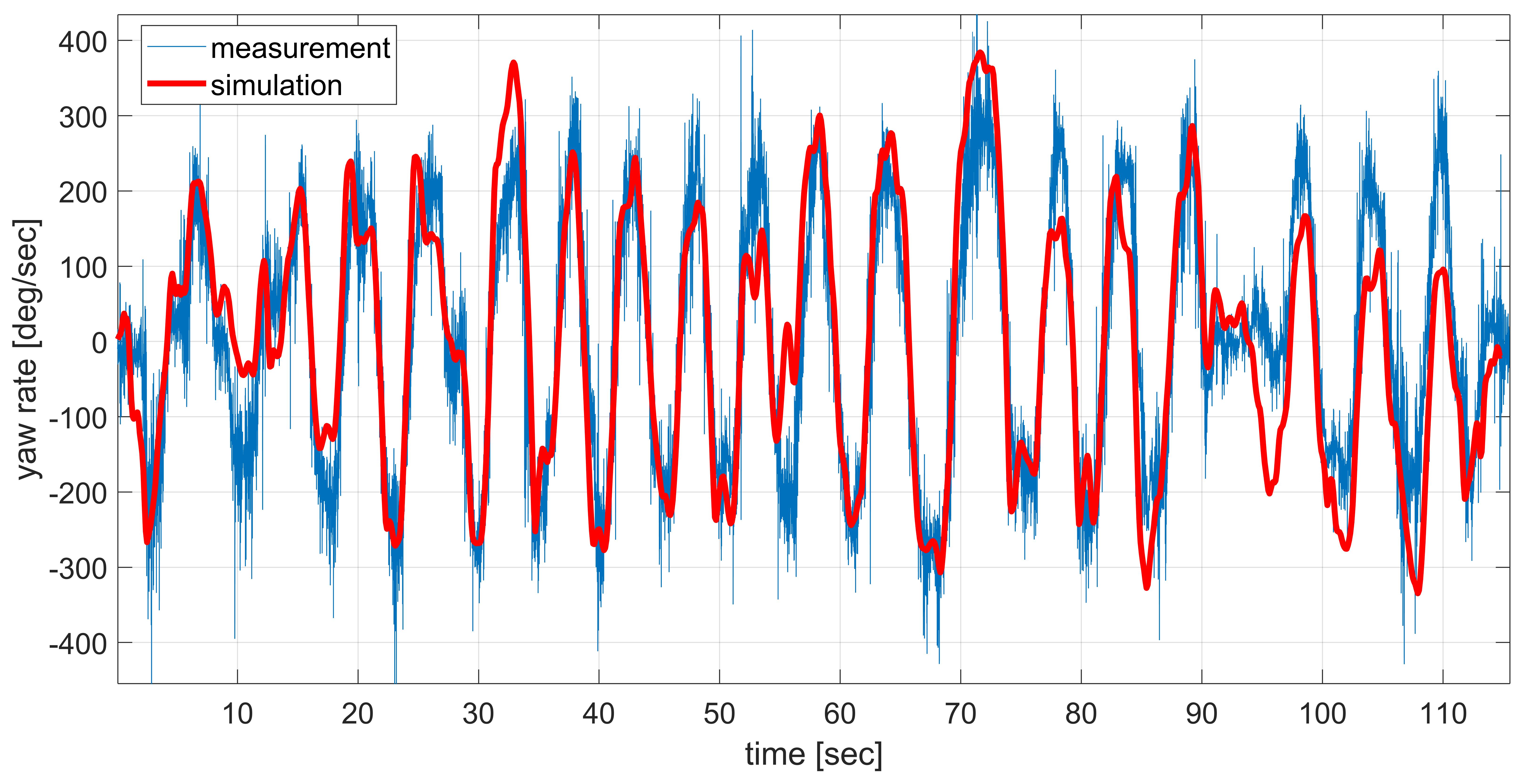}
    \caption{Comparison of the yaw rate measured during the rotations experiment in the wind tunnel and reconstructed by the skydiving simulator. }
    \label{fig:tunnel_yr}
\end{figure}
\begin{figure}
    \centering
     \includegraphics[width=\textwidth]{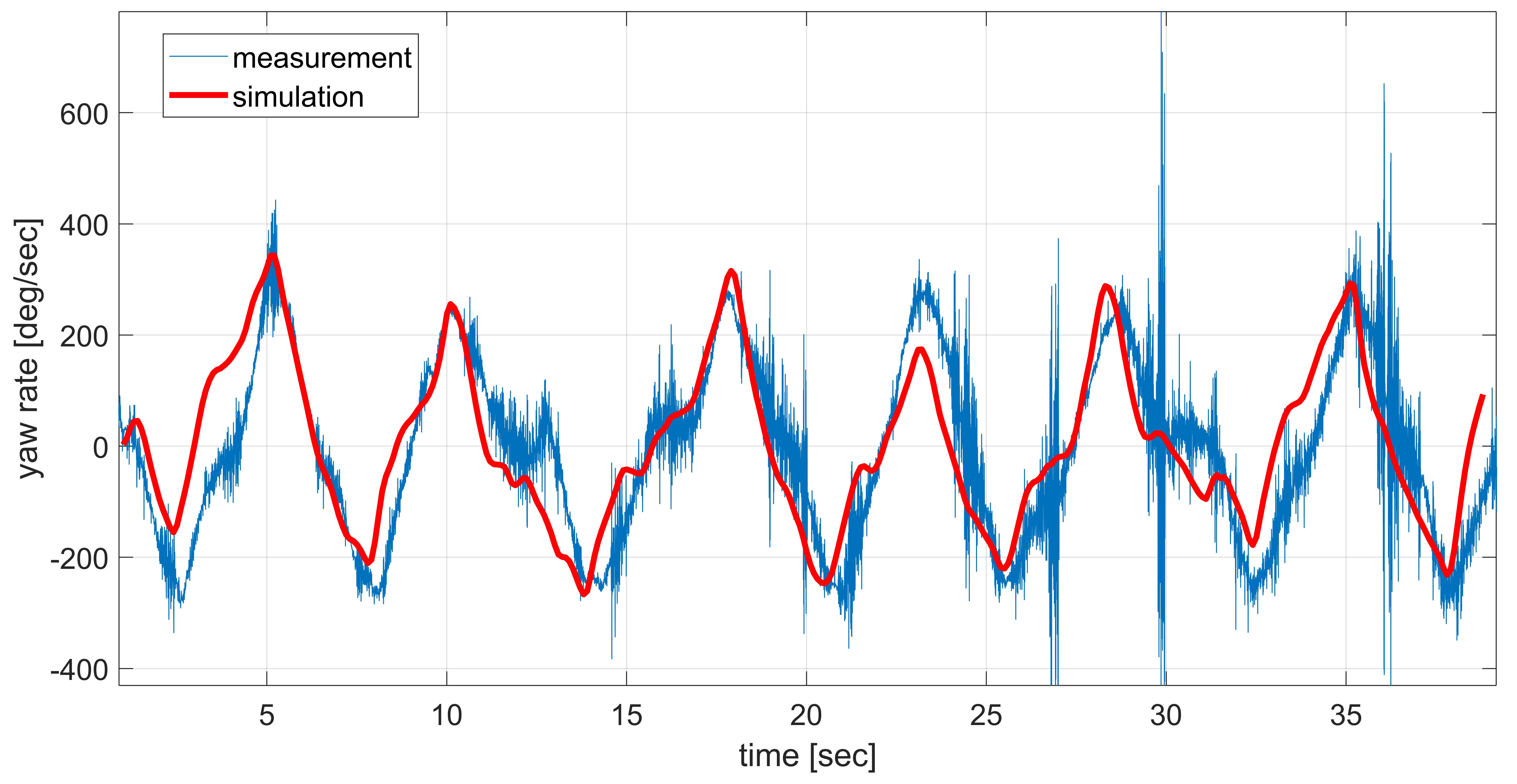}
    \caption{Comparison of the yaw rate measured during the rotations experiment in free-fall and reconstructed by the skydiving simulator. }
    \label{fig:ff_yr}
\end{figure}
%Each participant was equipped with the Xsens suit and performed the calibration procedures, defined by Xsens, that are necessary for the convergence of the measurement system. The participants' body segments and height were measured and inputted into the Xsens software, as well as into the skydiving simulator, along with weight, helmet size, jumpsuit type, and size and weight of the parachute container when relevant.

%Next, the participants were instructed what maneuvers have to be performed during their free-fall or wind tunnel session. For example, 
One of the basic skydiving skills is performing aerial rotations. Thus, one of the participants was instructed to enter the wind tunnel and perform 360 degrees turns to the left and to the right in a belly-to-earth pose during two minutes – a typical wind tunnel session. Additionally, the participant was instructed to prevent any horizontal or vertical displacement relative to the initial body position in the center of the tunnel. Analogously, the participant performing rotations in free-fall, was instructed to exit the airplane, reach terminal velocity, face the video operator, and perform 360 degrees right and left turns using the video operator as a visual reference. After about 35 seconds both skydivers performed their normal brake-off and parachute deployment procedures. 

The Xsens postures recorded in both environments were then inputted into the skydiving simulator and the yaw rate measured during the turns and reconstructed in the simulator was compared, as shown in Figures \ref{fig:tunnel_yr} and \ref{fig:ff_yr}. Additionally, it was verified that the simulator exhibits a correct behavior for the neutral posture, as explained below.

\subsection{The Trim Condition}
\begin{figure}
    \centering
     \includegraphics[width=0.4\textwidth]{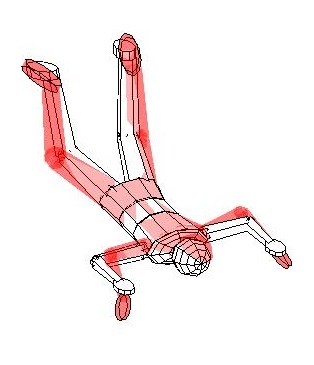}
    \caption{Skydiver body in a standard neutral posture (white) with an overlaid half-transparent posture (red) measured in the tunnel when the skydiver was neutral.}
    \label{fig:trim}
\end{figure}

The Trim Condition of the skydiver model describes a neutral posture such that the skydiver falls vertically with terminal velocity, with no horizontal movement and no rotations. A neutral posture in a belly-to-earth orientation is normally described as follows: slight arch in the torso, the head is slightly upwards, the arms are at 90 degrees to the torso and the elbows are bent at 90 degrees, the legs are slightly apart and the knees are bent. Such a neutral posture is defined in Figure \ref {fig:coord_axis}, and shown in Figure \ref{fig:trim} together with an overlaid posture recorded in the wind tunnel at the moment when the skydiver was not performing any maneuvers. It can be seen that the actual neutral posture of the specific participant is very similar to the standard neutral posture. 

%\begin{wrapfigure}{l}{0.4\textwidth}
%\end{wrapfigure}
The trim condition in simulation is achieved when zero is substituted for all derivatives in the equations of motion, and they are solved for posture parameters. Since the neutral posture is symmetrical along the anteroposterior axis (Z-Body), the relevant equations are: forces along Z and Y Body axis, and a pitching moment. The force equation along Y-Body axis determines the terminal velocity. The force equation along Z-Body axis and the pitching moment equation determine the angle of bending the knees and the Body angle of attack. The angle of attack in a trim situation is the Body pitch angle, that should be around 90 degrees (since the posture is belly-to-earth) and, together with degree of bending the knees, is responsible for the balance of the body on 'a column of air': not pitching up or down and not moving forward or backward. The rest of body DOFs are set to the measured values during the trim procedure.

Thus, the three equations can be solved for three most important parameters that describe the neutral fall: terminal velocity, body pitch angle, and the knees bending angle. The solution for these values must be similar to the values measured in the wind tunnel. This indicates that the aerodynamic coefficients are tuned correctly, and that the mass distribution assumed for the skydiver model produces the correct position of the center of gravity.
%\nomenclature{$f$}{[rad/s] frequency}

\subsection{Collected Data and Simulation Results}
The data collected in multiple wind tunnel and free-fall experiments was deposited to the open access data repository \cite{dataset}, along with video recordings of the experiments and data processing summary. %Appendix \ref{sec:exp} contains a detailed description of the participants and results of reconstruction of the basic maneuvers in a belly-to-earth orientation. 
 Videos of the two experiments presented in Figures \ref{fig:tunnel_yr}, \ref{fig:ff_yr} can be viewed on-line via \cite{figshare} (Chapter: \textit{Aerial Rotation Experiments}). Reconstruction of additional basic maneuvers can be found in \cite{Clarke:2021}.
 Reconstruction of advanced maneuvers is described in Section \ref{sec:advanced_maneuvers_ukf}.

\subsection{Insight into the Aerodynamic Model and its Parameters}
\label{sec:aero_params}
The aerodynamic coefficients ($Cd^{max}$,  $Cl^{max}$, $Cm^{max}$), and the damping moment coefficients ($\vec{Cm}_{damp}$) were tuned to the following values: 
\begin{equation}
    \begin{gathered}
    Cd^{max}=1.2 \\
    Cl^{max}=1.8 \\
    Cm^{max}=3.5 \\
    \vec{Cm}_{damp}=[3 \quad 0.5 \quad 6]^T
    \end{gathered}
    \label{eq:params}
\end{equation}
The maximum drag coefficient related to the parallel direction of the local wind ($Cd^{max}$) is responsible for the terminal velocity that the simulated skydiver will converge to. The maximum drag coefficient related to the perpendicular direction of the local wind ($Cl^{max}$) is responsible for the correct amplitude of simulated maneuvers: such that they match the corresponding experiments. The maximum moment coefficient ($Cm^{max}$) is responsible for the correct balance of all aerodynamic forces, ensuring that the transitions between maneuvers match the experiments. The damping moment coefficients are responsible for the correct dynamics of the maneuver (progress in time) and reflects an approximate body resistance (e.g. muscle stiffness) to the developing rotation rates. For this reason the damping moment coefficients depend on the maneuver. The values given in Eq. \eqref{eq:params} match the aerial rotations maneuver. During an intentional turning the body naturally resists much less to yaw rate (as opposed to roll and pitch rate), therefore the yaw damping moment coefficient has the smallest value.

Tuning of the aerodynamic parameters proved to be very simple and straightforward. Initially, we started the parameters tuning according to the following procedure: 
\begin{enumerate}
\item Assume that each body limb $i$ may have different aerodynamic coefficients: $Cl_i^{max}$, $Cd_i^{max}$, $Cm_i^{max}$. The reason is the different amount of flapping clothing on different body parts, while helmet, shoes, and gloves are made of completely different materials.
\item Run Monte-Carlo simulations of the same posture sequence, each time choosing a set of aerodynamic coefficients from a normal distribution around their pre-defined values (initial guess according to published results in similar experiments).
\item Take a set of coefficients that gives the minimum discrepancy between the simulated and measured maneuver (linear and angular velocities). Verify that the simulation gives accurate results reconstructing experiments that were not used for the Monte-Carlo simulations.
\end{enumerate}
It was discovered that such fine tuning is not necessary, and most of the differences between the aerodynamic coefficients of different body segments can be neglected. The simulation provides an accurate matching to recorded maneuvers, if the coefficients given in Eq. \eqref{eq:params} are
used for all body segments. The simulation is not sensitive to small changes in these coefficients: up to 30\% change in any/all coefficients for any/all body segments produces less than 1\% change in the simulation results, i.e. body linear and angular velocities. 

However, the skydiver model is sensitive to the following factors:
\begin{enumerate}
\item \textit{The Center of Gravity}: The skydiver body is modeled by body segments, while each segment has a certain shape, volume, and mass. If the overall center of gravity is modeled incorrectly, the skydiver simulation produces a completely unreasonable behavior. The Xsens system has an internal very detailed biomechanical model of mass distribution in the human body. One of the Xsens outputs is the position of the center of gravity at each instant of time. Therefore, in order to verify the correct modeling, the center of gravity computed by the skydiver model was compared to the one reported by Xsens. In most cases they were almost identical.  
\item \textit{The Derivatives of the Center of Gravity and the Inertia Tensor}: Those derivatives are computed numerically and if not smoothed - cause the skydiver simulation to diverge or produce erroneous results. A low pass filter (second order, relative damping 0.7, frequency 2 Hz) was used to smooth the derivatives.
\item \textit{The Shoe Size}: The shoes are modeled by elliptical cylinders, and it is very important to verify that the ellipse parameters match the actual shoe size of the test-jumper. It is known from the experience of practicing skydivers that the type and size of shoes have a large influence on aerodynamics in free-fall. For example, skydivers report that it is nearly impossible to maintain stability when wearing clown shoes. The simulation fully confirms such observations and reconstructs the shoe-size sensitivity.
\end{enumerate}

\section{Data Processing}\label{sec:dat}
The difference between the wind tunnel and free-fall environments does not seem significant for the Skydiver Model developed in Section \ref{sec:model}. The simulation reconstructs equally well maneuvers performed in the tunnel and in free-fall. This means that a much higher turbulence of the airflow in the tunnel and airflow disruptions caused by wearing a parachute container in free-fall do not have a major effect on the magnitude of aerodynamic forces and moments acting on the body. This fact makes it possible to use wind tunnels as training facilities for skydivers practicing for professional competitions.

\subsection{Interpreting Xsens measurements in terms of a simplified body model}
\begin{figure}
\centering
\begin{subfigure}{.5\textwidth}
  \centering
  \includegraphics[width=.5\linewidth]{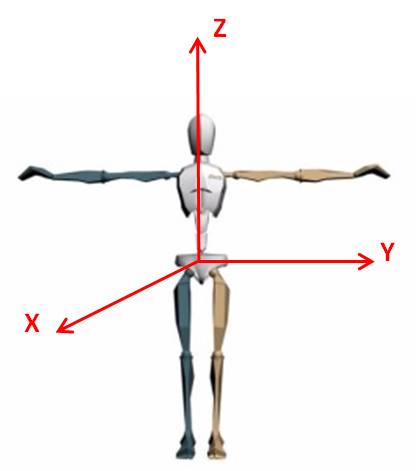}
  \caption{Xsens Coordinate System; Body in T-pose}
  \label{fig:fig3}
\end{subfigure}%
\begin{subfigure}{.5\textwidth}
  \centering
  \includegraphics[width=.7\linewidth]{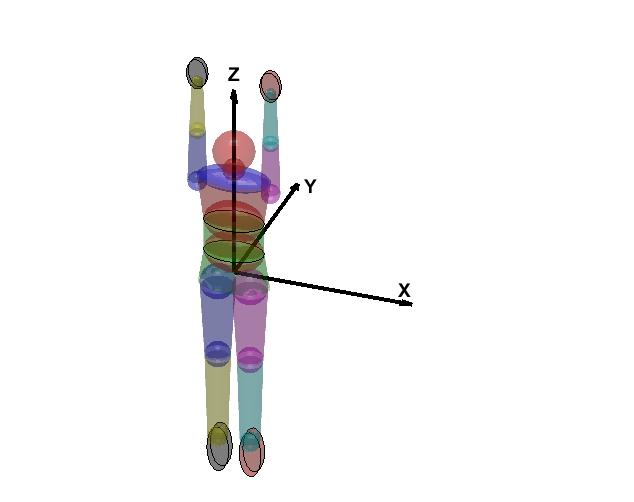}
  \caption{SM Coordinate System; Body in H-pose}
  \label{fig:fig2}
\end{subfigure}
\caption{Coordinate systems and default poses of the Xsens Body Model and the Skydiver Model (SM).}
\label{fig:figs23}
\end{figure}

The Body Model involved in the skydiving simulation consists of 16 segments, with the origin in the pelvis, and each child segment has 3 DOF relative to its parent segment. The input to the simulation is a set of quaternions defining the rotation of all child segments to the matching parent segments in Skydiver Model coordinates relative to the default H-pose, shown in Figure \ref{fig:fig2}. The output of the Xsens measurement system is a set of quaternions of rotation of 23 segments relative to the inertial coordinates and the default T-pose shown in Figure \ref{fig:fig3}. In order to translate Xsens measurements into Skydiver Model modeling framework, it is necessary to perform the following computations: 
\begin{enumerate}
\item For each pair of connected segments: construct a quaternion of relative rotation from two quaternions expressing segment orientation relative to the inertial frame. 
\item Unite segments that are used in a more detailed Xsens model and are not required in the Skydiver Model.
\item Express a transformation between Xsens and the Skydiver Model (SM) coordinate systems.
\item Express a transformation between T-pose and H-pose for the relevant segments (upper arms, forearms, hands, feet)
\end{enumerate}

For example, computation of the rotation quaternion from Right Lower Arm to Right Wrist in H-pose in SM coordinate system $\left( \vec{q}^{\,RLArm-H}_{RWr-H} \right)_{SM}$ is given in Eq. \eqref{eq:qXsens2SM}.
\begin{equation} \label{eq:qXsens2SM}
\begin{gathered}
\left( \vec{q}^{\,RLArm-H}_{RWr-H} \right)_{SM}=   \left( \vec{q}^{\,Xsens}_{SM} \right)^*  \otimes \left( \vec{q}^{\,RLArm-H}_{RLArm-T} \right)_{Xsens} \otimes ... \\ \left( \vec{q}^{\,RLArm-T}_{RWr-T} \right)_{Xsens} \otimes \left( \vec{q}^{\,RWr-T}_{RWr-H} \right)_{Xsens} \otimes \vec{q}^{\,Xsens}_{SM}
\end{gathered}
\end{equation}
where $(\,)^*$ denotes complex conjugate; $\vec{q}^{\,Xsens}_{SM}$ describes rotation from Xsens to SM frame (Eq. \eqref{eq:qXsens2SM1}); $\left( \vec{q}^{\,RLArm-H}_{RLArm-T} \right)_{Xsens}$ describes rotation of the lower arm from H-pose to T-pose in Xsens frame (Eq. \eqref{eq:qXsens2SM2}); $\left( \vec{q}^{\,RWr-T}_{RWr-H} \right)_{Xsens}$ describes rotation of the wrist from T-pose to H-pose in Xsens frame (Eq. \eqref{eq:qXsens2SM3}); and $\left( \vec{q}^{\,RLArm-T}_{RWr-T} \right)_{Xsens}$ describes the rotation from the local lower arm frame to the local wrist frame assuming T-pose and Xsens coordinates (is computed from Xsens measurements, Eq. \eqref{eq:qXsens2SM4}). 
\begin{equation}\label{eq:qXsens2SM1}
    \vec{q}^{\,Xsens}_{SM}=\left[ cos(\frac{\pi}{4}) \quad 0 \quad 0 \quad sin(\frac{\pi}{4}) \right]^T
\end{equation}
\begin{equation}\label{eq:qXsens2SM2}
    \left( \vec{q}^{\,RLArm-H}_{RLArm-T} \right)_{Xsens}=\left[ cos(\frac{\pi}{4}) \quad  -sin(\frac{\pi}{4}) \quad 0 \quad 0 \right]^T
\end{equation}
 \begin{equation}\label{eq:qXsens2SM3}
     \left( \vec{q}^{\,RWr-T}_{RWr-H} \right)_{Xsens}=\left[ cos(\frac{\pi}{4}) \quad 0 \quad sin(\frac{\pi}{4}) \quad 0 \right]^T \otimes \left[ cos(\frac{\pi}{4}) \quad  sin(\frac{\pi}{4}) \quad 0 \quad 0 \right]^T
 \end{equation}
\begin{equation}\label{eq:qXsens2SM4}
\left( \vec{q}^{\,RLArm-T}_{RWr-T} \right)_{Xsens}=\left( \vec{q}^{\,I}_{RLArm-T} \right)^{*}_{Xsens} \otimes \left( \vec{q}^{\,I}_{RWr-T} \right)_{Xsens}
\end{equation}
The Xsens measurements are $\vec{q}^{\,I}_{Limb}$ rotation from the inertial frame to the local frame attached to each limb, assuming T-pose and Xsens axis.

\begin{figure}
    \centering
     \includegraphics[width=\textwidth]{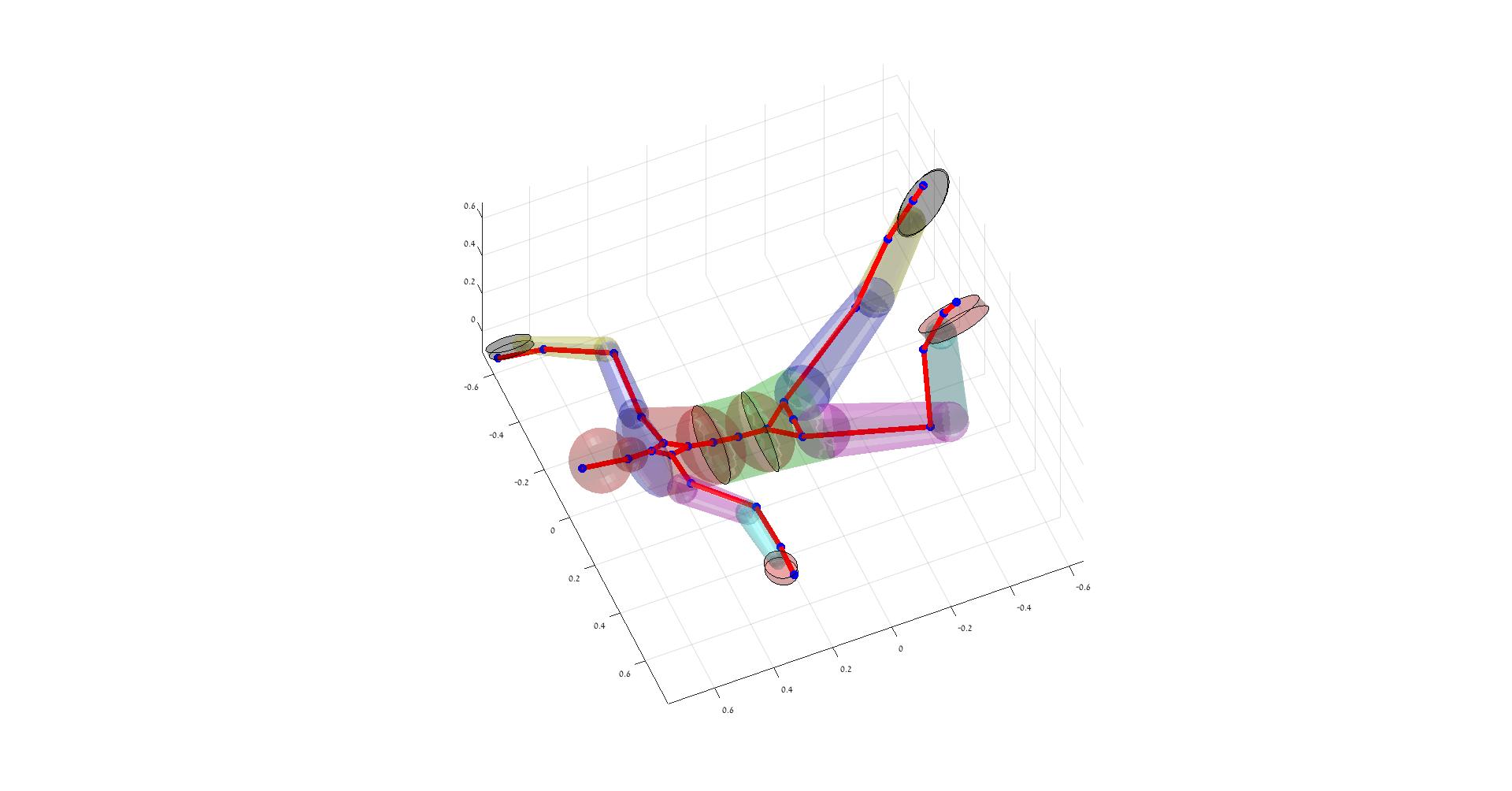}
    \caption{Skydiver body model with an overlaid skeleton constructed from sensors' positions reported by the Xsens system.}
    \label{fig:fig1}
\end{figure}

Processing all other segments in a similar way allows to construct a skydiver posture, an example of which shown in Figure \ref{fig:fig1}, from inertial orientation of segments provided by Xsens.

\subsection{Measurement noise}

The Xsens system output is a result of sensor fusion algorithms, therefore the inertial orientations of body segments are smooth and generally do not require additional filtering. However, they become much noisier in the wind tunnel environment. The reasons might be the magnetic field disturbances due to the metal construction of the tunnel, the high frequency noise sensed by gyros and accelerometers in the free-fall environment, and the inefficiency of the biomecanical walking model central to Xsens algorithms due to not touching the floor during free-fall. 

The skydiver model is most sensitive to noise when computing the numerical derivative of the center of gravity and the tensor of inertia. Therefore, these two signals are passed through low pass filers prior to using them in the motion equations. The body posture is constructed from unfiltered signals, according to the above procedure.

\section{Conscious Control  in Body Flight} \label{sec:conscious}
The skydiving simulation, described in Section \ref{sec:model},  is configured by the means of the following types of parameters:
\begin{enumerate}
    \item \textbf{Body Parameters}: height, weight, shape of every limb, size of helmet, type of jumpsuit, shoes and gloves. These parameters are constant for each experiment.
    \item \textbf{Aerodynamic Constants}: maximum drag and moment coefficients are set such that the simulation exhibits a reasonable terminal velocity and reaction to posture changes. These coefficients are fine tuned in experiments, as explained in Section \ref{sec:aero_params}.
    \item \textbf{Damping Moment Coefficients}: yaw, pitch, and roll damping moment coefficients reflect  an  approximate  body  resistance  (e.g.   muscle stiffness) to the developing rotation rates. These coefficients depend on the trainee and can thus vary depending on the performed maneuver, see Section \ref{sec:aero_params}.
\end{enumerate}
Thus, the only skydiver's input (except, certainly, for the body posture) into the simulator's model is the damping moment coefficients, which have an influence only when rotational rates are developing. However, there are longitudinal maneuvers, e.g. angle flying, which also require a continuous human input. This input is muscle forces that can be applied in any of the limbs, and are perceived by trainees as 'pressing' on the airflow. This muscle input is needed during performing  advanced maneuvers for two purposes:
\begin{enumerate}
    \item \textbf{To change orientation of the body relative to horizon}. E.g. in angle flying the body is initially straight and lies in the horizontal plane, whereas the objective is to keep it straight but pitched down at some angle (say, 45 deg) relative to the horizon. This is executed by applying a muscle force in the upper body that will push it down resisting the aerodynamic force, pushing upwards. The spine might not stay completely straight during this transient movement, but once the desired pitch angle is reached, the body can be returned into its straight posture. The body will be flying at the 45 deg angle relative to the horizon for as long as the skydiver keeps applying the constant muscle force that resists the aerodynamic pitch moment. As soon as the skydiver relaxes his muscle tone, the body will return into the horizontal plane.
    \item \textbf{To change relative orientation of body limbs.} In a belly-to-earth neutral pose most muscles can be relaxed since the airflow is pushing the limbs exactly into a desired arched pose. In more advanced poses a significant muscle effort may be required for moving the limbs into a desired pose and keeping them there. E.g. in sit fly it is hard to place the arms in front of the torso, as the airflow pushes them backwards. 
\end{enumerate}
Therefore, there are actually three forces acting on each individual limb during skydiving maneuvers: gravity, the aerodynamic force, and the internal muscle force that changes the effect of the aerodynamic force.  In some situations the muscle force can be negligible, and in others - very significant.  

It is interesting to find a minimal way to represent these muscle efforts in our model, without a detailed modeling of many body muscles, ligaments and an interaction between the body segments. In Section \ref{sec:advanced_maneuvers_ukf} we explore such a modeling option, adding only one new parameter per axis (roll, pitch, and yaw), however, sufficient for reconstruction of some complex maneuvers. 

These parameters are termed \textit{input moment coefficients}. 
For most of the advanced maneuvers, such as angle flying, rolls, flips, and layouts, they play a significant role in ability of the model to truthfully reconstruct the performed maneuvers. The empirical knowledge acquired over the years of developing skydiving techniques and coaching novices also agrees that, except for moving the limbs, skydivers utilize additional variables for control purposes. The variables related to the \textit{input moments} are usually described as \textbf{engaging with the relative wind}, which is defined as follows:
%\paragraph{}
\begin{quote}
    The conscious action of applying physical resistance against the push of the relative wind so that we can maintain our chosen body position and retain control, \cite[page 31]{newell2020body}.
\end{quote}   
    The variables related to \textit{damping moments} are usually associated by skydiving coaches with \textbf{muscular rigidity}, required for flying in advanced body orientations, e.g.:
\begin{quote}
    The basic foundational body position for back-flying relies heavily on both position symmetry and core strength. We need to be strong enough to maintain some amount of muscular rigidity in order to reduce unintentional movement at the core, \cite[page 100]{newell2020body}. 
    
    Similarly, for head-down flying \textit{the maintenance of a strong core position is of paramount importance}, \cite[page 186]{newell2020body}.
\end{quote}    
    These two types of inputs, namely the input moments and damping moments, can be tightly coupled especially during flying in advanced body orientations and performing challenging maneuvers. For instance:
\begin{quote}    
    The basic head-up foundational position, or 'sit fly' position as it is more commonly known, relies heavily on both symmetry and core strength. This is most regularly achieved through the proper \textbf{engagement} of the upper back and shoulders, \cite[page 144]{newell2020body}.
\end{quote}    
    These observations, verified by our work on estimation of these additional user inputs from experiments, suggest that these are control variables that drive the skydiver's dynamics along with her body configuration (posture). In \cite[Section 3.5]{Clarke:2021} %\ref{sec:advanced_maneuvers_rnn} and \ref{sec:advanced_maneuvers_utc} 
    two control algorithms are suggested for incorporating input and damping moment coefficients into a control system hierarchy. It is shown how the additional control variables allow to track advanced maneuvers and maintain various body orientations.

\section{Modeling and Estimation of Conscious Control Components  from Experiments} \label{sec:advanced_maneuvers_ukf}

\subsection{Modeling of the Input Moments}
The idea is high level modeling. Low level modeling would start from the contraction of each individual muscle, relating the force it is producing to a specific segment, considering the forces transferred to each segment from its parent segment, and then summing up all the forces to compute the overall moments that can contribute to changing the body orientation in the 3D space. Instead, we start from this possible moment contribution and utilize our prior knowledge of its purpose: resisting the aerodynamic moments (which are already modeled for each segment in Section \ref{sec:model}) during the transient and probably matching them in the steady state. We also approximately know from empirical evidence what body segments are most actively engaged for maneuvering in each axis: yaw, pitch, and roll. If the representation of the overall muscle moment contribution is truthfully formulated, a model for forces producing these moments should be straightforward.   

Thus, the following moment is added to the model:
\begin{equation} \label{eq1}
\begin{gathered}
    \vec{M}_{muscle \atop input}=[M_{pitch}, M_{yaw}, M_{roll}]^T \\
    M_{pitch}=-in_{pitch} \left (M_{atot} \right)_x^{center}\\ 
    M_{yaw}=-in_{yaw}\left [ \left (M_{atot} \right)_y^{up \atop prox} +  \left (M_{atot} \right)_y^{low \atop prox} + \left (M_{atot} \right)_y^{right \atop dist} + \left (M_{atot} \right)_y^{left \atop dist} \right ] \\ 
    M_{roll}=-in_{roll}\left [ \left (M_{atot} \right)_z^{left} + \left (M_{atot} \right)_z^{right} \right ]
    \end{gathered}
\end{equation}
where 
\begin{equation}
    \begin{gathered}
        \left (M_{atot} \right)_x^{center}=\sum_{i=1}^{N_{center}}\left(\vec{M}^i_{atot}\right)_x \\
        \left (M_{atot} \right)_y^{up \atop prox}=\sum_{i=1}^{N{up \atop prox}}\left(\vec{M}^i_{atot}\right)_y \qquad
         \left (M_{atot} \right)_y^{low \atop prox}=\sum_{i=1}^{N{low \atop prox}}\left(\vec{M}^i_{atot}\right)_y \\
         \left (M_{atot} \right)_y^{right \atop dist}=\sum_{i=1}^{N{right \atop dist}}\left(\vec{M}^i_{atot}\right)_y \qquad
         \left (M_{atot} \right)_y^{left \atop dist}=\sum_{i=1}^{N{left \atop dist}}\left(\vec{M}^i_{atot}\right)_y \\
         \left (M_{atot} \right)_z^{left}=\sum_{i=1}^{N_{left}}\left(\vec{M}^i_{atot}\right)_z \qquad
         \left (M_{atot} \right)_z^{right}=\sum_{i=1}^{N_{right}}\left(\vec{M}^i_{atot}\right)_z. \\
    \end{gathered}
\end{equation}
The forces that produce the input moments are
\begin{equation} \label{eq1_1}
\begin{gathered}
    \vec{F}_{muscle \atop input}=[F_x, F_y, F_z]^T \\
    F_x=-in_{yaw}\left [\frac{1}{l_1} \left (M_{atot} \right)_y^{up \atop prox} -  \frac{1}{l_2} \left (M_{atot} \right)_y^{low \atop prox} \right ] \\ 
    F_y
    =\frac{in_{pitch}}{l_1}\left (M_{atot} \right)_x^{center}-\frac{in_{roll}}{l_3}\left [ \left (M_{atot} \right)_z^{left} - \left (M_{atot} \right)_z^{right} \right ]\\ 
    F_z
    =-\frac{in_{yaw}}{l_4}\left [  \left (M_{atot} \right)_y^{right \atop dist} - \left (M_{atot} \right)_y^{left \atop dist} \right ] 
    \end{gathered}
\end{equation}
where $in_{pitch}$, $in_{yaw}$, $in_{roll}$ are dimensionless coefficients specifying the skydiver's muscle input; $N{up \atop prox}$, $N{low \atop prox}$, $N_{right}$, $N_{left}$, $N{right\atop dist}$, $N{left\atop dist}$ are the number of limbs used by the skydiver to produce the relevant input, depending on the performed maneuver. For example, for back-to-earth tracking, described later in this section, it holds that   
\begin{itemize}
    \item upper body proximal limbs: thorax, head, upper arms - $N{up\atop prox}=4$
    \item lower body proximal limbs: hips - $N{low\atop prox}=2$
    \item right/left distal limbs: legs, forearms - $N{right\atop dist}=N{left\atop dist}=2$
    \item right/left limbs: hips, legs, forearms, upper arms - $N_{right}=N_{left}=4$.
\end{itemize}
$l_1$, $l_2$, $l_3$, $l_4$ [m] are the characteristic lever arms, and, 
finally, $\vec{M}^i_{atot}$ [Nm] is the total aerodynamic moment acting on limb $i$ expressed in Body coordinate system and computed as in Eq. \eqref{eq:sum_m}:
\begin{equation}
    \vec{M^i_{atot}}=\displaystyle\sum_{i=1}^{N_{limbs}}\left(\vec{r_{cg}^i} \times \vec{F_a}^i + \vec{M_a}^i \right).
\end{equation}

%\nomenclature{$in_{pitch},in_{yaw},in_{roll}$}{[-] pitch, yaw, and roll input moment coefficients}
%\nomenclature{$\vec{M}_{muscle \atop input}=[M_{pitch}, M_{yaw}, M_{roll}]^T$}{[Nm] muscle input moments}
%\nomenclature{$\vec{F}_{muscle \atop input}=[F_x, F_y, F_z]^T$}{[N] muscle input forces}
%\nomenclature{$l_i$}{[m] characteristic length of limb $i$}

Notice the minus sign in Eq. \eqref{eq1}: the input moment is resisting the aerodynamic moment acting on the limbs. The input coefficient value determines how much of the aerodynamic moment is compensated, specifically if an input coefficient is equal to 1  means that the muscles are imposing on the limbs the same moment (but in the opposite direction) as imposed by the airflow.  In this situation the skydiver's current orientation in space is expected to be maintained, as in angle flying. The case of negative input coefficients is treated later in this section.

%this is exactly what happens in angle flying - the skydivers feel their head and thorax have to 'press' on the airflow with a constant force in order to keep a constant angle of their body towards the horizon. If this pressure is released the airflow 'pushes' their body back into a horizontal plane (due to a pitch moment).

From the experiments with different skydivers it was observed that the more proficient skydivers use less limbs and less effort to produce those input moments. The skydiver who performed the maneuvers described later in this section was more proficient in a belly-to-earth orientation than being back-to-earth. Therefore, the model for tracking belly-to-earth excluded legs and forearms from the yaw moment equation and legs, hips, upper arms, and forearms from the roll moment equation. 

\subsection{Estimation of the Input Moment Coefficients}

Altogether the skydiving simulator includes six dimensionless coefficients related to user input: three (yaw, pitch, roll) damping moment coefficients ($\vec{Cm}_{damp}$ in Eq. \eqref{eq:sum_m}) and three input moment coefficients ($in_{pitch}$, $in_{yaw}$, $in_{roll}$ in Eq. \eqref{eq1}). For the simple maneuvers (e.g. rotations described in Section \ref{sec:model}) performed by experienced skydivers the input moments were neglected (not modeled) and the damping moment coefficients were modeled as constants and tuned in simulation. However, while performing complex maneuvers, the skydiver changes his inputs and the resistance to developing angular rates during the experiment. Thus, these six coefficients in Eq. \eqref{eq:ukf_state}  can be estimated from the recorded data as functions of time. 

The measurements available from the experiment are: body (pelvis) inertial heading, pitch and roll angles measured by the Xsens system, and the horizontal velocity component computed from the latitude and longitude measurements of the GNSS system that was synchronized with the Xsens.    

The chosen estimation framework is the Unscented Kalman Filter (UKF) \cite{julier1997new} with some modifications, explained below. The main reason for this choice is the high non-linearity of the observation model. An Unscented Transform seems to be a better alternative to the linearization involved in e.g.the Extended Kalman Filter, and less complicated than a Particle Filter. Moreover, in our case, to derive analytical expressions for Jacobians would be highly complex.    

%\vspace{0.5cm}
\section{The Modified Unscented Kalman Filter  for the estimation of moment coefficients}\label{sec:modifiedukf}
%\textbf{Summary of the Unscented Kalman Filter for the estimation of moment coefficients}
\subsection{The state}
%\vspace{0.5cm}
%\textit{State}
%\vspace{0.25cm}

The states are input moment coefficients and damping moment coefficients:
\begin{equation}
\label{eq:ukf_state}
    \vec{X}=[in_{yaw}, in_{pitch}, in_{roll}, Cm_{damp}^{yaw}, Cm_{damp}^{pitch}, Cm_{damp}^{roll}]^T
\end{equation}
The state dimension is $n=6$. We do not have any {\it {\`a} priori}  knowledge about dynamics of these coefficients, therefore, it is assumed:
\begin{equation}
    \begin{gathered}
    \vec{X}_{k+1}=\vec{X}_k+\vec{w}_k, \qquad
    \vec{w}_k \sim N(0,Q) 
    \end{gathered}
\end{equation}
where $\vec{w}_k$ is the process noise with diagonal covariance matrix whose value was chosen by tuning to be  $Q_{i,i}=[0.25, 0.25, 0.25, 1, 1, 1] \cdot dt, \quad i=1,2,..6$, and $dt$ [s] is the simulation step. 
\vspace{0.5cm}
%\nomenclature{$\vec{X}_k,\vec{Z}_k,\vec{Z}pred_k$}{state, measurement, and predicted measurement vectors at step $k$ in the UKF framework}
%\nomenclature{$Q,R$}{constant covariance matrices of the process and measurement noise, respectively, used in the UKF framework}
%\nomenclature{$\vec{Skydiver}_k$}{state of the skydiver dynamics at step $k$ used in the UKF framework}

\subsection{Measurements}
%\vspace{0.25cm}
The measurement vector (for the tracking experiment) includes the pelvis inertial orientation and the horizontal velocity component: 
\begin{equation} \label{eq:meas}
    \vec{Z}_k=[V_{hor}, heading, pitch, roll]^T_k, \quad k=1,...t_{end} \cdot 240
\end{equation}
The measurement dimension is $m=4$. The length of the skydive is $t_{end}$, the measurements and the skydiver's posture are recorded at 240 Hz. From sensor data, the measurement noise covariance is a diagonal matrix $R$:  $R_{i,i}=[0.1,0.05,0.01,0.01]^2$, $i=1,..4$.
%\vspace{0.5cm}

\subsection{Initial Conditions}
%\vspace{0.25cm}
Initial state and covariance matrix (diagonal) for step $k=0$ are defined as:
\begin{equation}
    \begin{gathered}
      \vec{X}_{k/k}=[0, 0, 0, 6, 6, 6]^T \\
      \left (P_{k/k} \right )_{i,i}=[0.25, 0.25, 0.25, 6, 6, 6]^2, \quad i=1,..6
    \end{gathered}
\end{equation}
In addition to the UKF state $\vec{X}$, we initialize the state of the skydiver dynamics: skydiver's initial orientation (expressed by a quaternion $\vec{q}^{\, B}_I$ from Body to Inertial frame), and angular and linear velocity ($\vec{\Omega}$ and $\vec{V}$, respectively). 
\begin{equation}
    \vec{Skydiver}_k=[\vec{q}^{\, B}_I, \vec{\Omega}, \vec{V}]^T
\end{equation}
The state $\vec{X}$ along with the recorded skydiver's postures are the inputs for the skydiver model when it is propagated forward in time. 
%\vspace{0.5cm}
\subsection{Sigma Points}
%\vspace{0.25cm}
At each simulation step we choose $2n+1$ sigma points $\vec{X_i}$ and their associated weights $W_i$ in the following way, where the value of $W_0$ was chosen by tuning:
\begin{equation} \label{eq:sigpukf}
    \begin{gathered}
      \vec{X_0}=\vec{X}_{k/k} \\
      \vec{X_i}=\vec{X}_{k/k} \pm \sqrt{\frac{n}{1-W_0}}\vec{S}_j, \qquad  i=1,..2n, \quad j=1,..n \\
      W_i=\frac{1-W_0}{2n}, \quad W_0=0.5
    \end{gathered}
\end{equation}
where $\vec{S}_j$ is column $j$ of matrix $S$ that satisfies $S \cdot S = P_{k/k}$.
\vspace{0.5cm}

%\nomenclature{$\vec{X}_{\cdot/\cdot}, P_{\cdot/\cdot}=S \cdot S$}{estimated state vector and its covariance matrix in the UKF framework}
%\nomenclature{$\vec{X}_i, W_i, \vec{Z}_i$}{Sigma points, their associated weights, and Sigma points propagated through the observation model. Used in the UKF framework}
%\nomenclature{$\vec{S}_j$}{column $j$ of matrix $S$ used for computing the Sigma points}

\subsection{State Constraints}
%\vspace{0.25cm}
After the sigma points are selected they are bounded by the minimum and maximum values of the skydiver's parameters under investigation:
\begin{equation} \label{eq:minmax}
    \begin{gathered}
\left(\vec{X_i}\right)_{min}=[0, 0, 0, 0.01, 0.01, 0.01]^T \\
\left(\vec{X_i}\right)_{max}=[6.5, 6.5, 6.5, 24, 24, 24]^T
 \end{gathered}
\end{equation}
Additionally, in contrast to the suggestion in \cite{kandepu2008constrained}, we prevent two sigma points from being identical, which may happen after imposing the constraints, by modifying the relevant elements of sigma points. If an element $j$ of $ \vec{X}_{k/k}$ equals to its min or max boundary and $\vec{X_i}(j)<\left(\vec{X_i}\right)_{min}(j)$ or $\vec{X_i}(j)>\left(\vec{X_i}\right)_{max}(j)$, respectively, then this element is moved to the other side of the boundary,
\begin{equation} \label{eq:movepukf}
            \vec{X_i}(j)= \vec{X}_{k/k}(j)-k_i(\vec{X_i}(j)-\vec{X}_{k/k}(j)),
\end{equation}
%\begin{equation} \label{eq:movepukf}
%            \vec{X_i}(j)=1.5 \cdot \vec{X}_{k/k}(j)-0.5 \cdot \vec{X_i}(j)
%\end{equation}
where the coefficent $k_i$ was tuned to 0.5.
%\vspace{0.25cm}

\subsection{Observation Model}
%\vspace{0.25cm}
Each sigma point drives the skydiver model, thus each $\vec{X_i}$ has a matching skydiver state $\vec{Skydiver}(\vec{X_i})=[\vec{q}^{\, B}_I, \vec{\Omega}, \vec{V}]^T$. This state is used to compute the predicted measurement $\vec{Z_i}$ for each sigma point. First, the Body axes are computed in the Xsens coordinate system: \begin{equation} \label{eq:xyz}
    \begin{gathered}
 \vec{q}_z=[cos\frac{\pi}{4}, 0, 0, -sin\frac{\pi}{4}]^T \\
        \vec{X}_{body}=\vec{q}_z \otimes (\vec{q}^{\, B}_I \otimes [1, 0, 0]^T) \\
        \vec{Y}_{body}=\vec{q}_z \otimes (\vec{q}^{\, B}_I \otimes [0, 1, 0]^T) \\
        \vec{Z}_{body}=\vec{q}_z \otimes (\vec{q}^{\, B}_I \otimes [0, 0, 1]^T)
        \end{gathered}
\end{equation}
%\nomenclature{$\vec{X}_{body},\vec{Y}_{body},\vec{Z}_{body}$}{unity vectors representing the Body axes in the Xsens coordinate system}
Next, the body heading, pitch and roll angles are computed for each sigma point $i$:
\begin{equation} \label{eq:ypr}
    \begin{gathered}
        heading^i=arctan\frac{\vec{Z}_{body}(2)}{ \vec{Z}_{body}(1)} \\
        pitch^i=arctan\frac{\vec{Z}_{body}(3)}{\sqrt{ \vec{Z}_{body}(1)^2 +  \vec{Z}_{body}(2)^2 }} \\
        roll^i=sign(\vec{Y}_{body}(3)) \cdot arctan\frac{\vec{X}_{body}(3)}{\sqrt{ \vec{X}_{body}(1)^2 +  \vec{X}_{body}(2)^2 }} + \pi \cdot (\vec{Y}_{body}(3)<0)
           \end{gathered}
\end{equation}
The expression for $roll$ is valid also for back-to-earth flying, and can be used during transitions from belly to back and vice-a-versa. Finally, the horizontal velocity component is computed as:
\begin{equation} \label{eq:vel}
\begin{gathered}
\vec{V_I}=\vec{q}^{\, B}_I \otimes \vec{V} \\
V_{hor}^i=\sqrt{ \vec{V_I}(1)^2+ \vec{V_I}(2)^2}
\end{gathered}
\end{equation}
Summarizing the above equations, we get $\vec{Z_i}=[V_{hor}^i, heading^i, pitch^i, roll^i]^T$.
\vspace{0.5cm}       
%\nomenclature{$dt$}{simulation step, in this work 1/240 s}
%\nomenclature{$N_{pred}, N_{solve}$}{number of prediction steps, and number of solution steps in the UKF framework}

\subsection{Prediction Window and Solution Window}
%\vspace{0.25cm}
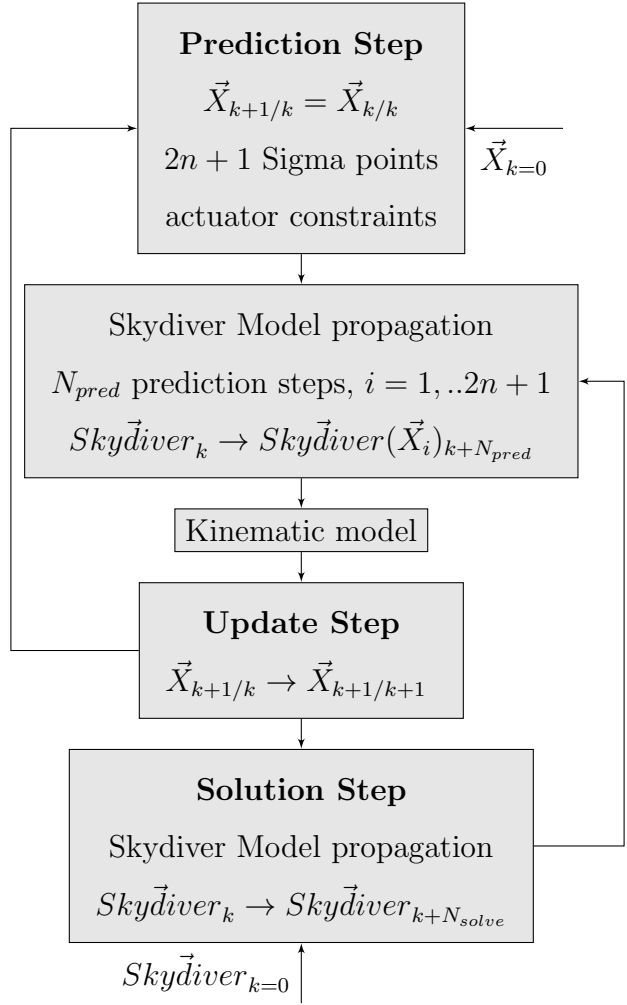
\begin{figure}[!ht]
\begin{center}
%\begin{wrapfigure}{l}{0.5\textwidth}
    \centering
    \tikzstyle{blockgr} = [draw, fill=gray!20, rectangle, 
    minimum height=0.5em, minimum width=4em]
\tikzstyle{input} = [coordinate]

 \begin{tikzpicture}[auto, node distance=3.7cm, >=latex']
\node [input](in1) {};
\node [blockgr, left=0.9cm of in1] (b1) { \begin{tabular}{c} \textbf{Prediction Step} \\ {$\vec{X}_{k+1/k}=\vec{X}_{k/k}$} \\ {$2n+1$} Sigma points \\ actuator constraints  \end{tabular}};
\node [input, right=1.3cm of b1](in2) {$\vec{X}_{k=0}$};
\node[blockgr, below=0.4cm of b1](b2){ \begin{tabular}{c}  Skydiver Model propagation \\ $N_{pred}$ prediction steps, {$i=1,..2n+1$} \\ {$\vec{Skydiver}_k \rightarrow \vec{Skydiver}(\vec{X_i})_{k+N_{pred}}$} \end{tabular}};
\node [blockgr, below=0.4cm of b2] (b3) { Kinematic model};
\node [blockgr, below=0.4cm of b3] (b4) { \begin{tabular}{c} \textbf{Update Step} \\ {$\vec{X}_{k+1/k} \rightarrow \vec{X}_{k+1/k+1}$ }  \end{tabular}};
\node[blockgr, below=0.4cm of b4](b5){ \begin{tabular}{c}  \textbf{Solution Step} \\ Skydiver Model propagation \\ {$\vec{Skydiver}_k \rightarrow \vec{Skydiver}_{k+N_{solve}}$}  \end{tabular}};
\node [input, below=0.8cm of b5](in3) {$\vec{Skydiver}_{k=0}$};

\draw [->] (b1.south) -- node {} (b2.north);
\draw [->] (b2.south) -- node {} (b3.north);
\draw [->] (b3.south) -- node {} (b4.north);
\draw [->] (b4.south) -- node {} (b5.north);
\path [draw,->] (b5.east) -- ($(b5.east) + (1.2, 0)$) |- (b2.east);
\path [draw,->] (b4.west) -- ($(b4.west) - (1.7, 0)$) |- (b1.west);
\draw [->] (in2) -- node {{$\vec{X}_{k=0}$}} (b1.east);
\draw [->] (in3) -- node {{$ \vec{Skydiver}_{k=0} $}} (b5.south);
    \end{tikzpicture}
    \caption{Block diagram of the modified Unscented Kalman Filter.}
    \label{fig:ukf_diag}
    %\vspace{-0.5cm}
    \end{center}
\end{figure}
In the conventional UKF the prediction is computed one step forward, then the predicted measurements computed for each sigma point are compared to $\vec{Z}_k$. However, in order for the sigma points to reflect the influence of skydiver's muscles input on his motion in the 3-D space, more than one step of $dt=1/240$ s is needed. Thus, we take into account the skydiver's dynamics time constant by computing the prediction during $N_{pred}$ steps. During the prediction 'window' the parameter state $\vec{X}$ remains constant (and thus the sigma points) and the skydiver's motion is propagated for each sigma point and according to the recorded postures that match the window time. Next, we compare the measurement that matches the last step of the prediction window to the skydiver's velocity and orientation, computed for each sigma point. The best results were obtained for $N_{pred}=0.25 \cdot 240$. This means that if a given set of sigma points drives the skydiver dynamics for 0.25 s, it is possible to make meaningful conclusions regarding the influence of muscles inputs on the body inertial motion. 

Another difference from the conventional UKF scheme is the 'solution' step, as shown in Figure \ref{fig:ukf_diag}. Whereas normally the UKF has prediction and update steps, we introduce an additional step, which propagates the model of the skydiver in time using the current state $\vec{X}$ as input. The model can be propagated during $dt$ (one step) or $N_{solve}$ steps. Its dynamics state (skydiver's orientation, angular/linear velocity) is then used as the initial condition for the skydiver models that run during the next prediction step for each sigma point. Since the muscle input parameters are not changing at 240 Hz it is possible to save computation time and use a longer 'solution' window. We used 3-4 steps for skydives that included highly dynamic maneuvers (flips and rolls) and 10-12 steps otherwise.   
%\vspace{0.5cm}

\subsection{Prediction Step}
    \begin{enumerate}
        \item \textbf{Selection of sigma points and enforcement of state constraints}
        
        Sigma points are selected according to Eq. (\ref{eq:sigpukf}), constrained according to Eqs. (\ref{eq:minmax}), (\ref{eq:movepukf}), and the predicted state is computed as:
        \begin{equation}
          \vec{X}_{k+1/k}=\sum_{i=0}^{2n}W_i\vec{X_i}
         \end{equation}
        \item \textbf{Propagation of the Skydiver model for each sigma point}
        
        Each $\vec{X_i}$ drives the skydiver model during $N_{pred}$ steps starting from $\vec{Skydiver}_k$, and the final skydiver's state is saved and coupled to its matching sigma point: $\vec{Skydiver}(\vec{X_i})$. 
        \item \textbf{Propagation of covariance and the sigma points through the observation model}
        
         \begin{equation}
          P_{k+1/k}=Q+\sum_{i=0}^{2n}W_i(\vec{X_i}-\vec{X}_{k+1/k})\cdot(\vec{X_i}-\vec{X}_{k+1/k})^T
         \end{equation}
         The predicted measurements $\vec{Z_i}$ are computed for each sigma point according to Eqs. (\ref{eq:xyz})-(\ref{eq:vel}), and summarized as:
         \begin{equation}
          \vec{Z}pred_{k}=\sum_{i=0}^{2n}W_i\vec{Z_i}
         \end{equation}
    \end{enumerate}
    
    \subsection{Update Step}
    \begin{enumerate}
        \item \textbf{Covariance of innovation and cross covariance computation}
        \begin{equation}
            C_z=R+\sum_{i=0}^{2n}W_i(\vec{Z_i}-\vec{Z}pred_{k})\cdot(\vec{Z_i}-\vec{Z}pred_{k})^T
        \end{equation}
         \begin{equation}
            C_{xz}=\sum_{i=0}^{2n}W_i(\vec{X_i}-\vec{X}_{k+1/k})\cdot(\vec{Z_i}-\vec{Z}pred_{k})^T
        \end{equation}
        \item \textbf{Data assimilation}
        \begin{equation}
            \begin{gathered}
            K=C_{xz}C_z^{-1} \\
            \vec{X}_{k+1/k+1}=\vec{X}_{k+1/k}+K(\vec{Z}_{k}-\vec{Z}pred_{k}) \\
            P_{k+1/k+1}=P_{k+1/k}-KC_zK^T
            \end{gathered}
        \end{equation}
        The updated state $\vec{X}_{k+1/k+1}$ is bounded according to Eq. (\ref{eq:minmax}).
    \end{enumerate}
\vspace{0.5cm}
%\nomenclature{$C_z,C_{xz}$}{covariance and cross covariance matrices in the UKF framework}
%\nomenclature{$K$}{gain matrix used in the UKF and UTC frameworks}

\subsection{Solution Step}
%\vspace{0.25cm}
The Skydiver state $\vec{Skydiver}_k$ is propagated during the solution window $N_{solve}$ using the current estimates for parameters $\vec{X}_{k+1/k+1}$, which remain constant during $N_{solve}$. The obtained Skydiver state 
\begin{equation*}
 [\vec{Skydiver}_k, \vec{Skydiver}_{k+1},.. \vec{Skydiver}_{k+N_{solve}-1}]
 \end{equation*}
is saved for error analysis (presented next), while the last state $\vec{Skydiver}_{k+N_{solve}-1}$ becomes the new initial condition for the next prediction step. 

Notice that the next step is not $k+1$, but rather $k+N_{solve}$. Since the values of $\vec{X}$ are held constant during the solution window, the state and covariance obtained in the update step are used as is:
\begin{equation}
            \begin{gathered}
            \vec{X}_{k+N_{solve}/k+N_{solve}}=\vec{X}_{k+1/k+1} \\
            P_{k+N_{solve}/k+N_{solve}}=P_{k+1/k+1}
\end{gathered}
        \end{equation}

\section{Experimental Results} \label{sec:experimentalresults}
\subsection{The Tracking Maneuver}
\begin{figure}[!h]
    \includegraphics[width=1\textwidth]{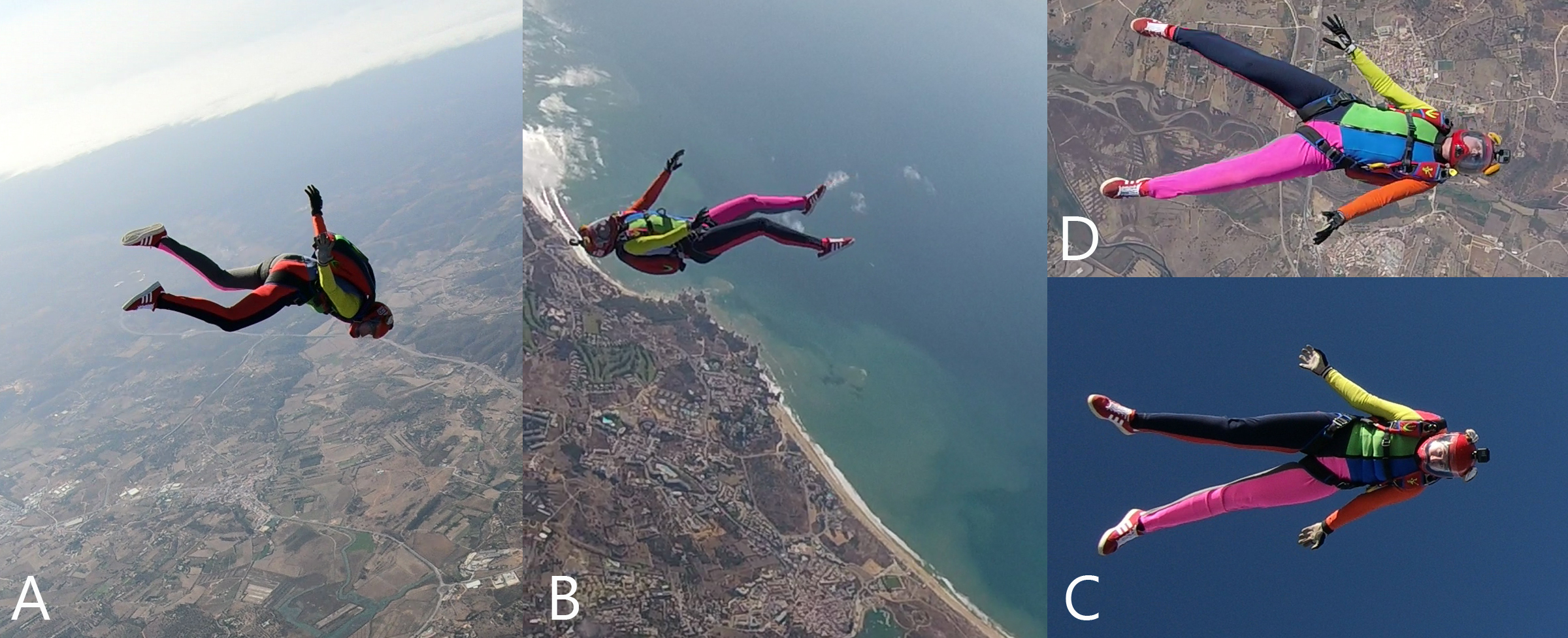}
    \caption{Side and top views of the belly-to-earth (A,C) and back-to-earth (B,D) typical tracking postures. Notice that the view (D) was taken by a photographer flying on belly above the skydiver, and the view (C) was taken by a photographer flying on back below the skydiver.}
    \label{fig:trackpic}
\end{figure}

The tracking maneuver, i.e. developing a significant horizontal velocity component,  is one of the most challenging for modeling and reconstruction in simulation. The reasons are the following:
\begin{enumerate}
    \item Tracking starts immediately after clearing the aircraft, i.e. before reaching the terminal velocity, and while being still influenced by the velocity of the aircraft.
    \item In order to track the skydiver flies at an angle towards the earth, rather than in a belly-to-earth pose. This is called \textit{angle flying}.
    \item This angle is maintained by a continuous user input from the skydiver's part: 'pressing' with his upper body on the airflow such that the pitching moment produced by his internal muscle effort is sufficient to keep the body angled towards the horizon.
    \item This effort is not constant and can even change dramatically during the experiment since this skydiver is novice at the angle flying discipline and experiences pitch oscillations and stability loss.
    \item The experiment includes a transition to back-to-earth pose via a barrel roll maneuver, tracking on the back in the same general direction, then returning to belly via a barrel roll, and again continuing the track. See Figure \ref{fig:trackpic}.
\end{enumerate}

\begin{figure}[htb]
    \centering
    \includegraphics[width=1\textwidth]{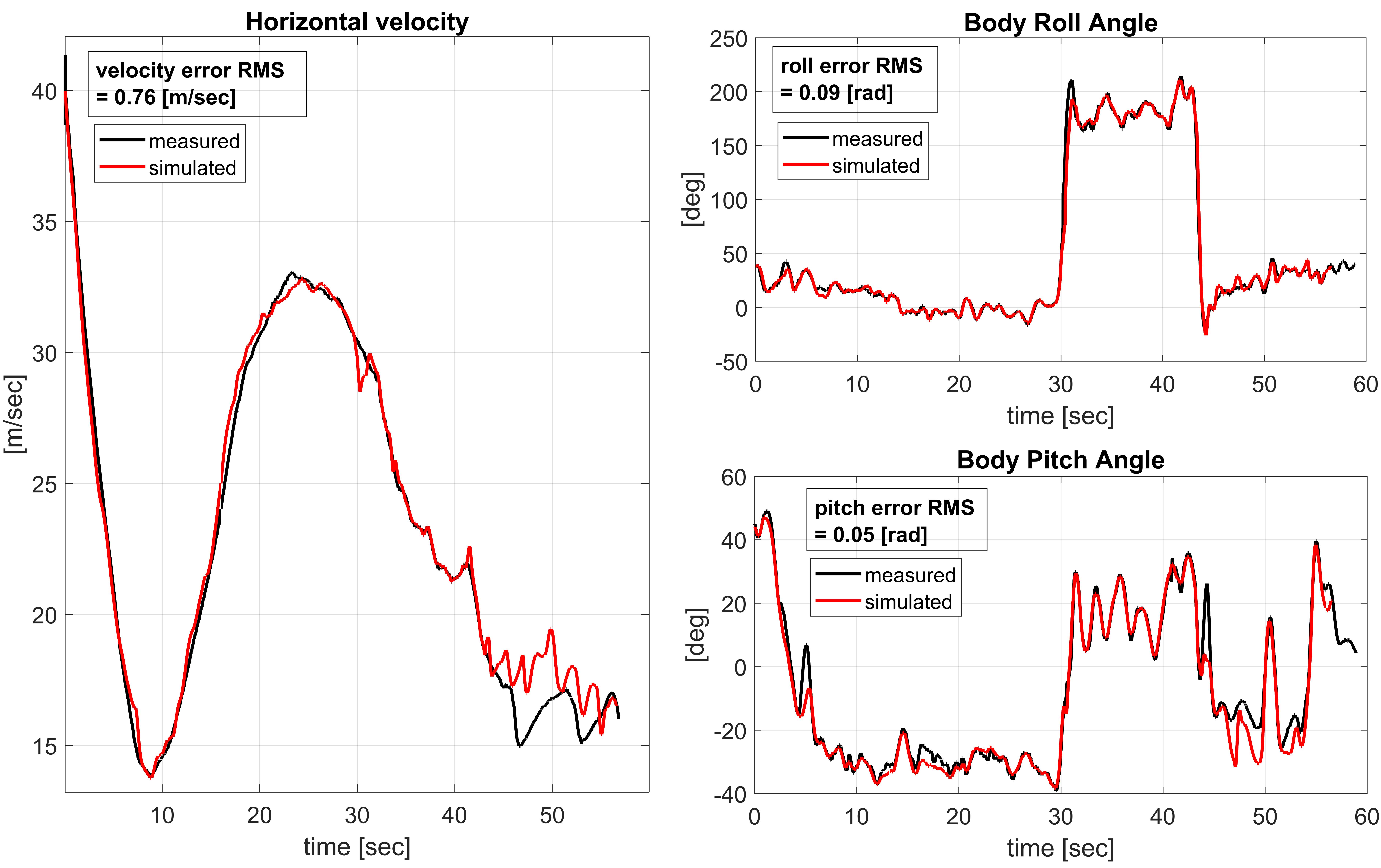}
    \caption{Reconstruction of the tracking experiment in simulation.}
    \label{fig:track}
\end{figure} 
\begin{figure}[!htb]
    \centering
    \begin{subfigure}{.48\textwidth}
    \centering
    \includegraphics[width=0.98\textwidth]{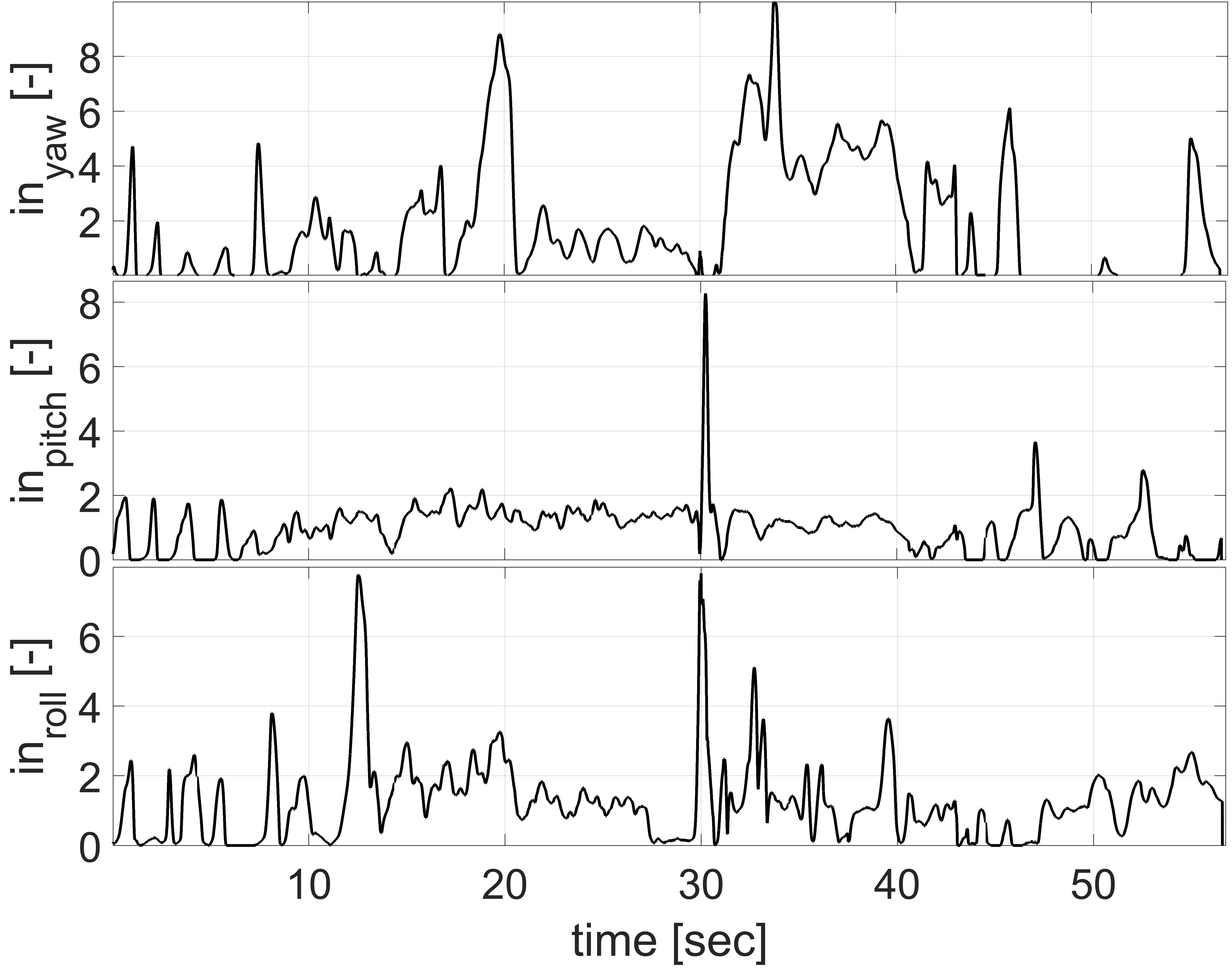}
    \caption{input moment coefficients}
    \label{fig:trackin}
\end{subfigure} 
\begin{subfigure}{.48\textwidth}
    \centering
    \includegraphics[width=0.98\textwidth]{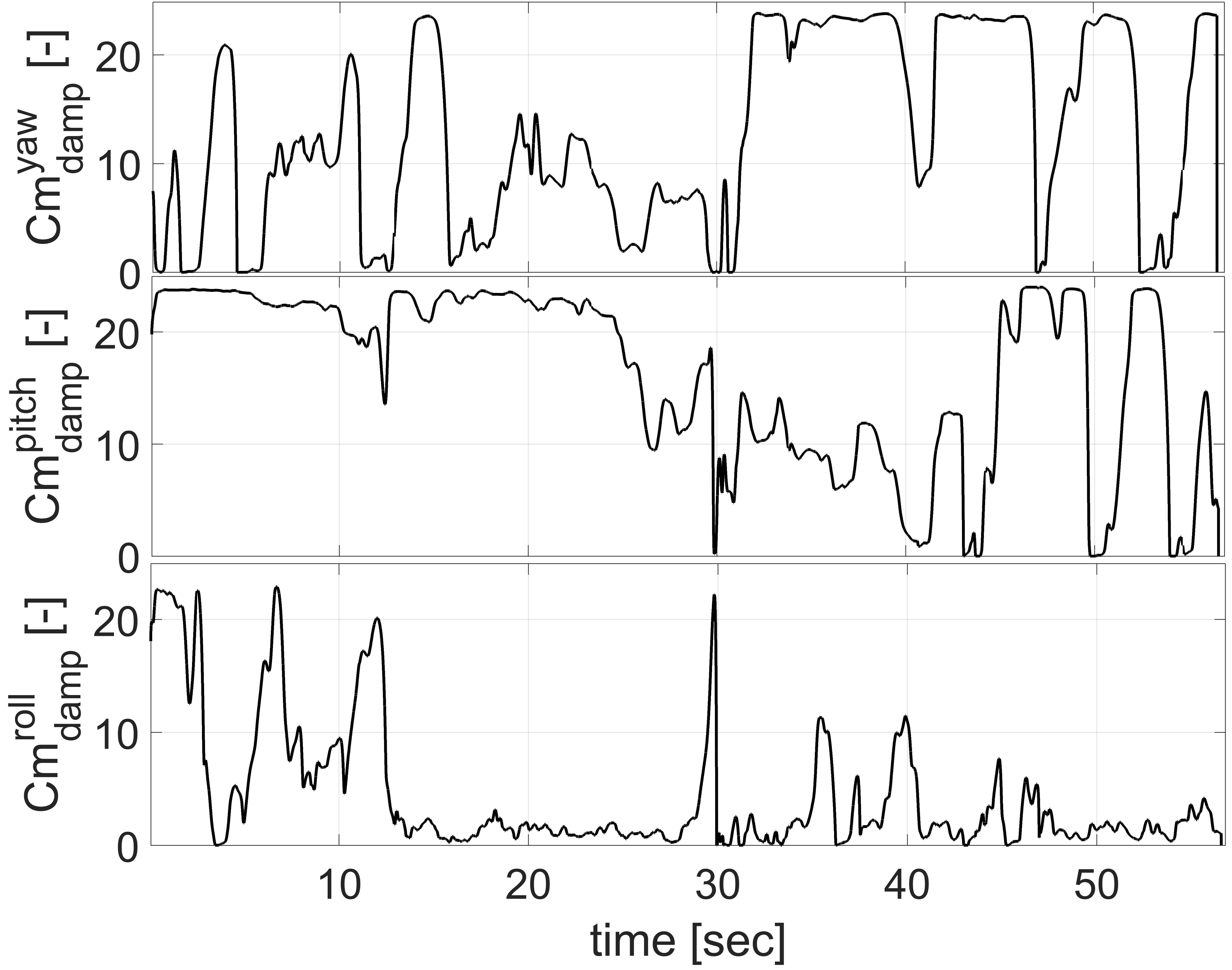}
    \caption{damping moment coefficients}
    \label{fig:trackdamp}
    \end{subfigure}
    
    \caption{Skydiver inputs during the tracking experiment, as estimated by the modified Unscented Kalman Filter.}
    \label{fig:trackcoeffs}
\end{figure} 

Reconstruction results are summarized in Figure \ref{fig:track}: it presents the measured and reconstructed in simulation body roll and pitch angles, and  horizontal velocity. Notice, the different stages of the experiment: 

    \textbf{The 'exit' stage} (0-10 s): the skydiver exits with the horizontal velocity of the aircraft, and high body pitch angle (about half way between an up-right and a belly-to-earth orientation). The horizontal velocity and the pitch angle decrease as the skydiver clears the aircraft and gets into a tracking pose.
    
    \textbf{The angle flying stage} (10-30 s): The skydiver tracks forward while the body is pitched down and his horizontal velocity increases until reaching its maximum at about 33 m/s. At this stage the skydiver maintains approximately a constant pressure on the airflow applied through his head and thorax: the pitch moment input coefficient is between 1 and 2, see Figure \ref{fig:trackin}. This means he is pressing on the airflow to match/ double the moment imposed on the head and thorax by the airflow. Also, notice that the pitch moment damping coefficient is quite high, see Figure \ref{fig:trackdamp}. This is needed to resist the pitching motion and maintain a constant pitch angle. Probably, with experience this skydiver will be able to maintain the same pitch angle by applying less pressure on the airflow and less resistance to the pitch rate.
    
    %\begin{wrapfigure}{l}{0.45\textwidth}
    \begin{figure}
    \centering
    \includegraphics[width=0.45\textwidth]{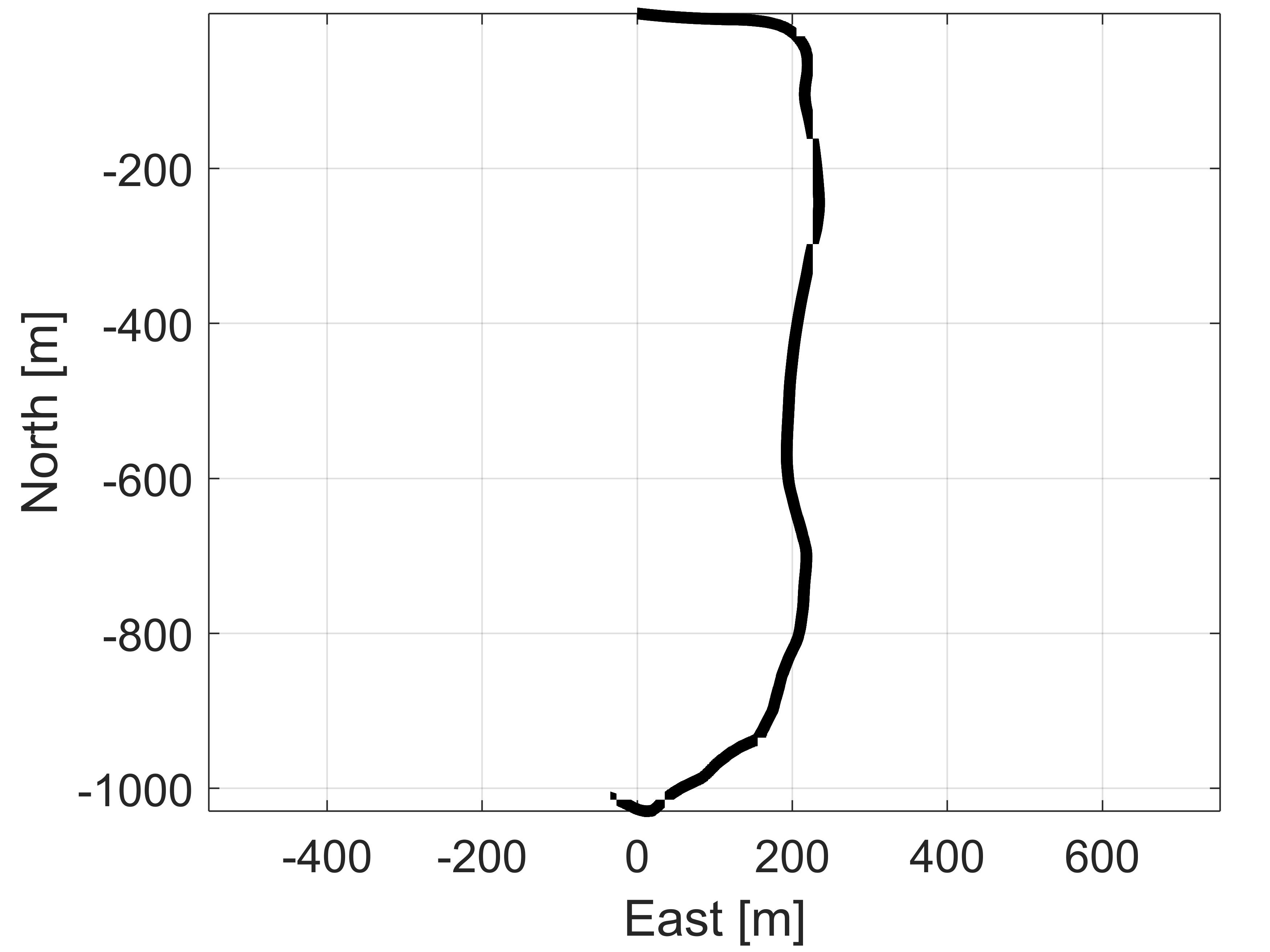}
    \caption{Top view of skydiver's trajectory during the tracking experiment, as reconstructed by simulation.}
    \label{fig:tracktop}
    %\vspace{-0.3cm}
    \end{figure}
%\end{wrapfigure} 
    \textbf{Transition to back tracking} ($\sim$30 s): The skydiver rolls from belly to back through the right side of the body. The transition is achieved by applying a strong roll input moment and resisting the roll motion while the equivalent reaction from the airflow is building up (at that time the body slightly rolls in the opposite direction to the desired), and then, releasing the resistance so that the aerodynamic moment acting on the body rolls it into a back-to-earth orientation. See Figures \ref{fig:trackin}, \ref{fig:trackdamp} for the timing of these inputs. Again, it seems that the skydiver is applying too much effort, and with experience will be able to significantly decrease both inputs.

    \textbf{The back tracking stage} (30-43 s): The skydiver continues the track in a back-to-earth orientation. Notice that in this pose the skydiver is less proficient and thus experiences pitch and roll oscillations (typical for novices), which are accurately reconstructed in simulation. The tracking maneuver becomes less efficient and the skydiver's horizontal velocity becomes smaller, see Figure \ref{fig:track}.
    
    \textbf{Transition to belly tracking} (43 s): The skydiver rolls back into a belly-to-earth orientation. Since this is a more accustomed pose, the transition is smoother than before and no excessive effort is applied: the roll moment input coefficient is between 1 and 2, and it is sufficient to drop the resistance to roll movement in order to let the transition happen. See Figures \ref{fig:trackin}, \ref{fig:trackdamp}.
    
    \textbf{Tracking belly-to-earth} (43-55 s): The skydiver continues the track in a belly-to-earth orientation, while preparing for stopping and deploying the parachute. For this reason the body pitch angle is not as steep as during the angle-flying stage, and the horizontal velocity remains small. The skydiver checks the altimeter around 50 s, what causes coming out of the angle-flying orientation for a while. Also, the skydiver is adjusting the tracking direction, see Figure \ref{fig:tracktop}, in order to acquire a view of the dropzone. Notice that until this stage the tracking direction was first - with the velocity of the aircraft, and once the exit was cleared - perpendicular to the direction of the aircraft, which is the tracking safety rule.

\subsection{Transition Maneuvers}

Another extremely challenging scenario for reconstruction in simulation is performing various transition maneuvers. Transitions allow the skydiver to switch between belly-to-earth and back-to-earth orientations, and can be performed in many different ways. The skydiver who performed the transitions in the experiment was more proficient at certain transition types, and novice at other types. This adds difficulty to reconstruct these maneuvers in simulation, since his inputs were moderate and smooth at some parts of the experiment and very unpredictable during other parts. A few times the skydiver was very close to loosing stability, i.e. initiating an uncontrollable spin/flip/roll. 

\begin{figure}[!htb]
    \centering
    \includegraphics[width=1\textwidth]{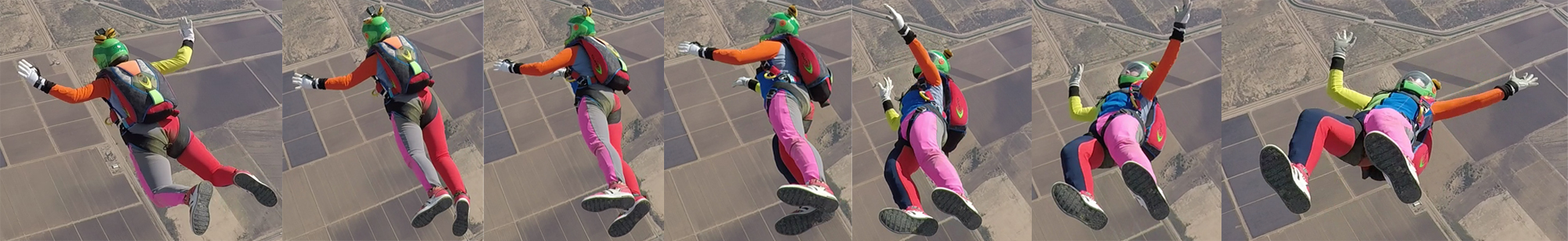}
    \caption{Barrel roll maneuver: transition from belly to back through the right side. Time sequence from left to right.}
    \label{fig:tr1}
    \vspace{-0.25cm}
\end{figure}

\begin{figure}[!htb]
    \centering
    \includegraphics[width=1\textwidth]{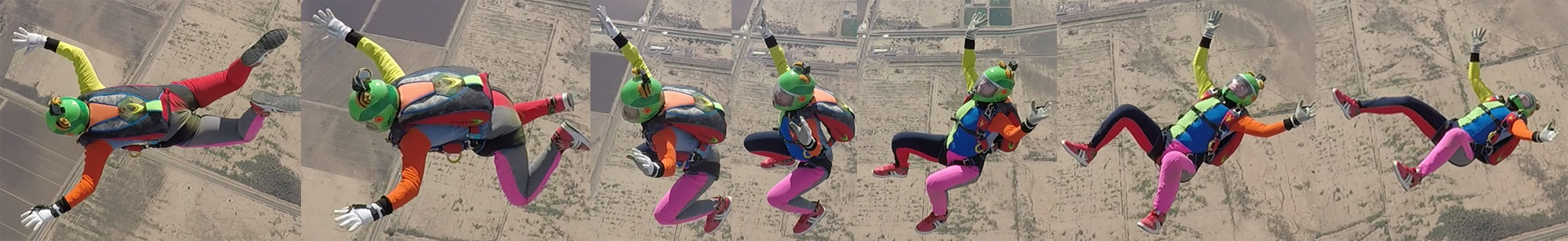}
    \caption{Transition from belly to back through the legs. Time sequence from left to right.}
    \label{fig:tr2}
    \vspace{-0.25cm}
\end{figure}

\begin{figure}[!htb]
    \centering
    \includegraphics[width=1\textwidth]{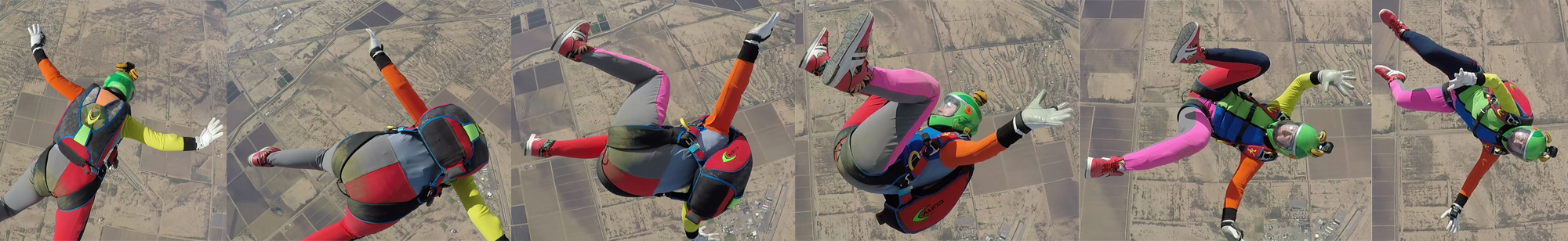}
    \caption{Front loop maneuver: transition from belly to back through the head. Time sequence from left to right. Notice, that the performance was not accurate, and the transition occurred mostly through the right side rather than the head.}
    \label{fig:tr3}
    \vspace{-0.25cm}
\end{figure}

\begin{figure}[!htb]
    \centering
    \includegraphics[width=1\textwidth]{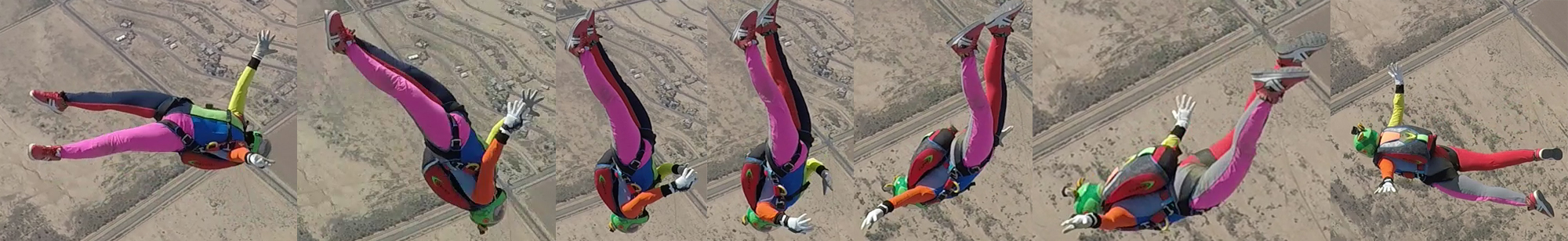}
    \caption{Layout maneuver: transition from back to belly through the head. Time sequence from left to right.}
    \vspace{-0.25cm}
    \label{fig:tr4}
\end{figure}

\begin{figure}[!htb]
    \centering
    \includegraphics[width=1\textwidth]{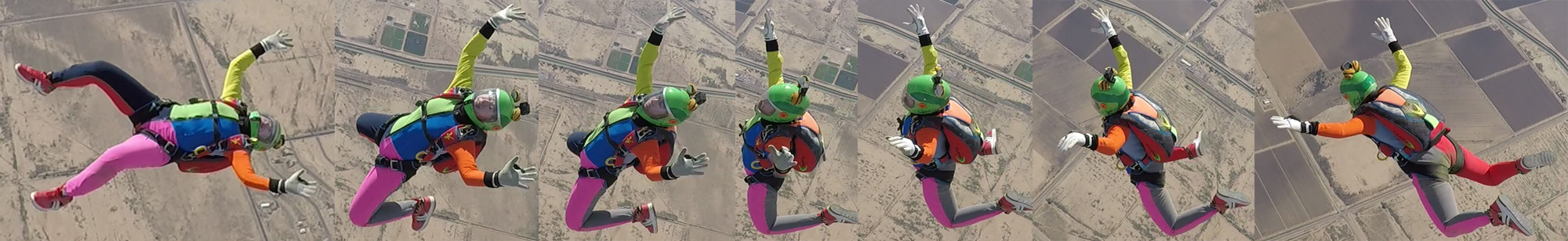}
    \caption{Transition from back to belly through the legs. Time sequence from left to right.}
    \label{fig:tr5}
\end{figure}

Altogether, during 45 seconds of 'working time', the skydiver performed 14 transitions in the following order:
\begin{multicols}{2}
\begin{enumerate}
    \item Back flip from belly to back (Figure \ref{fig:tr2})
    \item Barrel roll from back to belly
    \item Front loop (Figure \ref{fig:tr3})
    \item Barrel roll from back to belly
    \item Back flip from belly to back
    \item Back layout that didn't succeed and turned into a barrel roll
    \item Back flip from belly to back
    \item Front flip: back to belly (Figure \ref{fig:tr5})
    \item Barrel roll: belly to back (Figure \ref{fig:tr1})
    \item Back layout (Figure \ref{fig:tr4})
    \item Back flip from belly to back
    \item Back layout
    \item Back flip from belly to back
    \item Back layout
\end{enumerate}
\end{multicols}
\begin{figure}[!htb]
    \centering
    \includegraphics[width=1\textwidth]{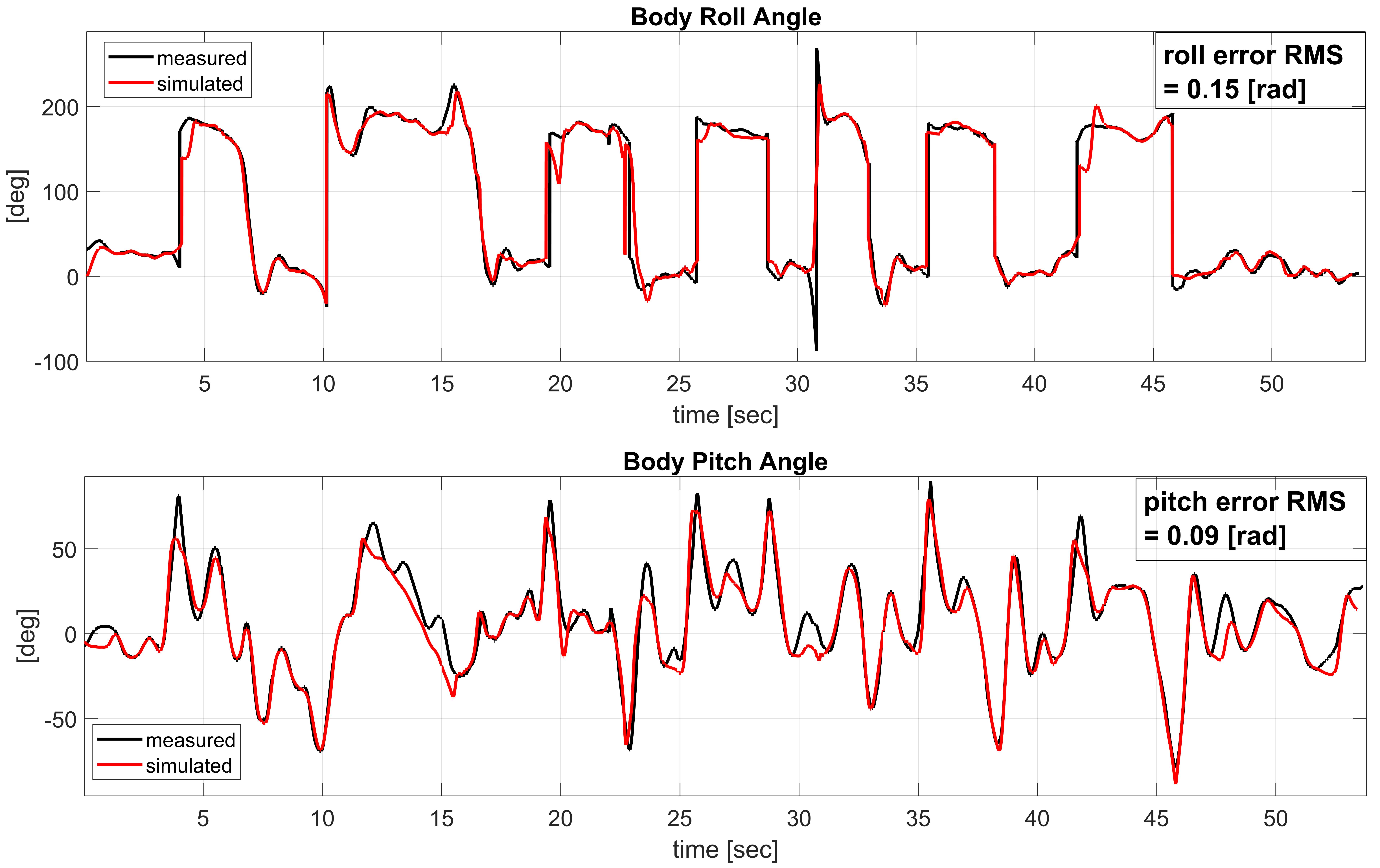}
    \caption{Reconstruction of the transition maneuvers in simulation. Damping moment coefficients are limited to $[0.01, 24]$.}
    \label{fig:tran}
\end{figure} 

\begin{figure}[!htb]
    \centering
    \begin{subfigure}{.48\textwidth}
    \centering
    \includegraphics[width=0.98\textwidth]{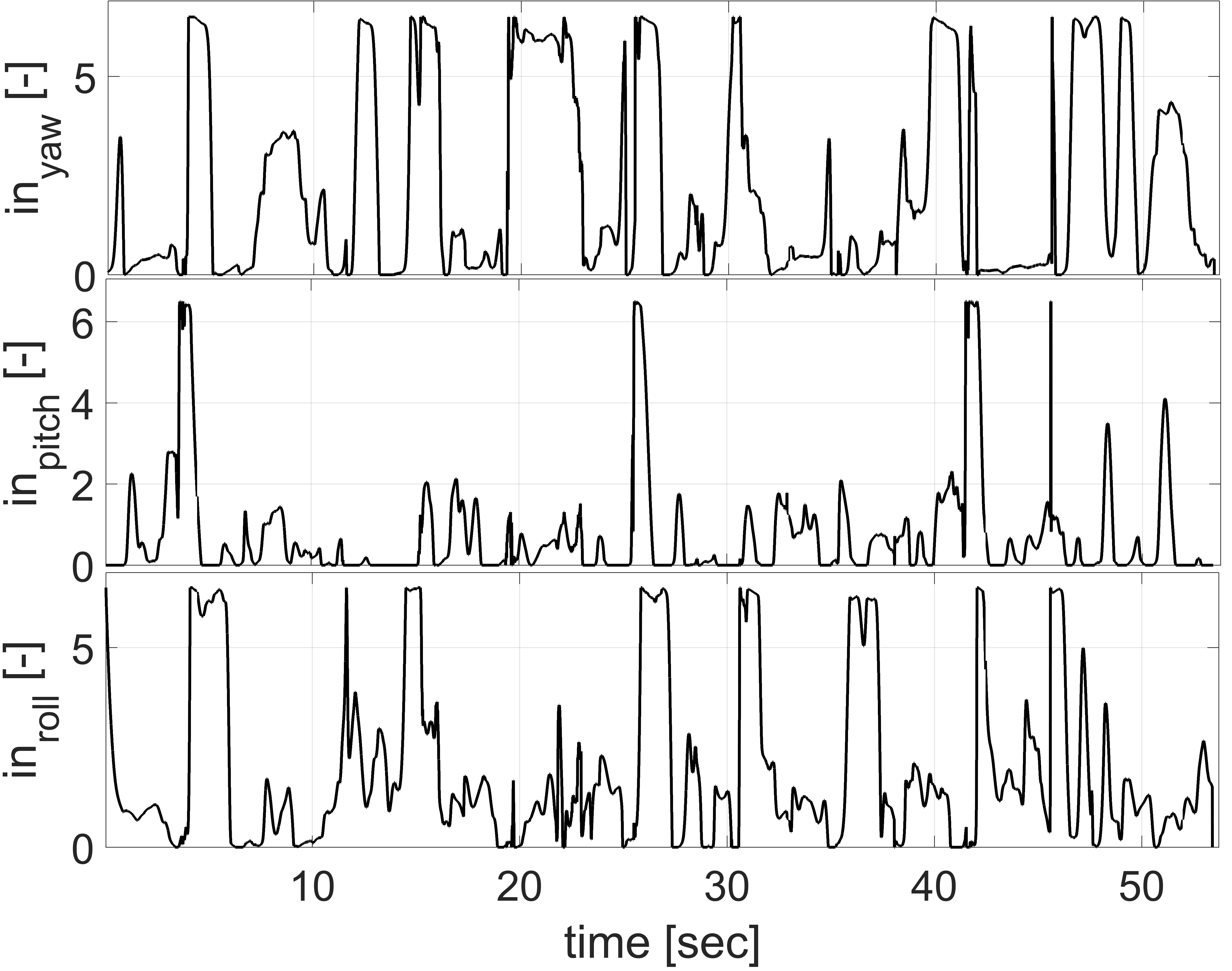}
    \caption{input moment coefficients}
    \label{fig:tranin}
\end{subfigure} 
\begin{subfigure}{.48\textwidth}
    \centering
    \includegraphics[width=0.98\textwidth]{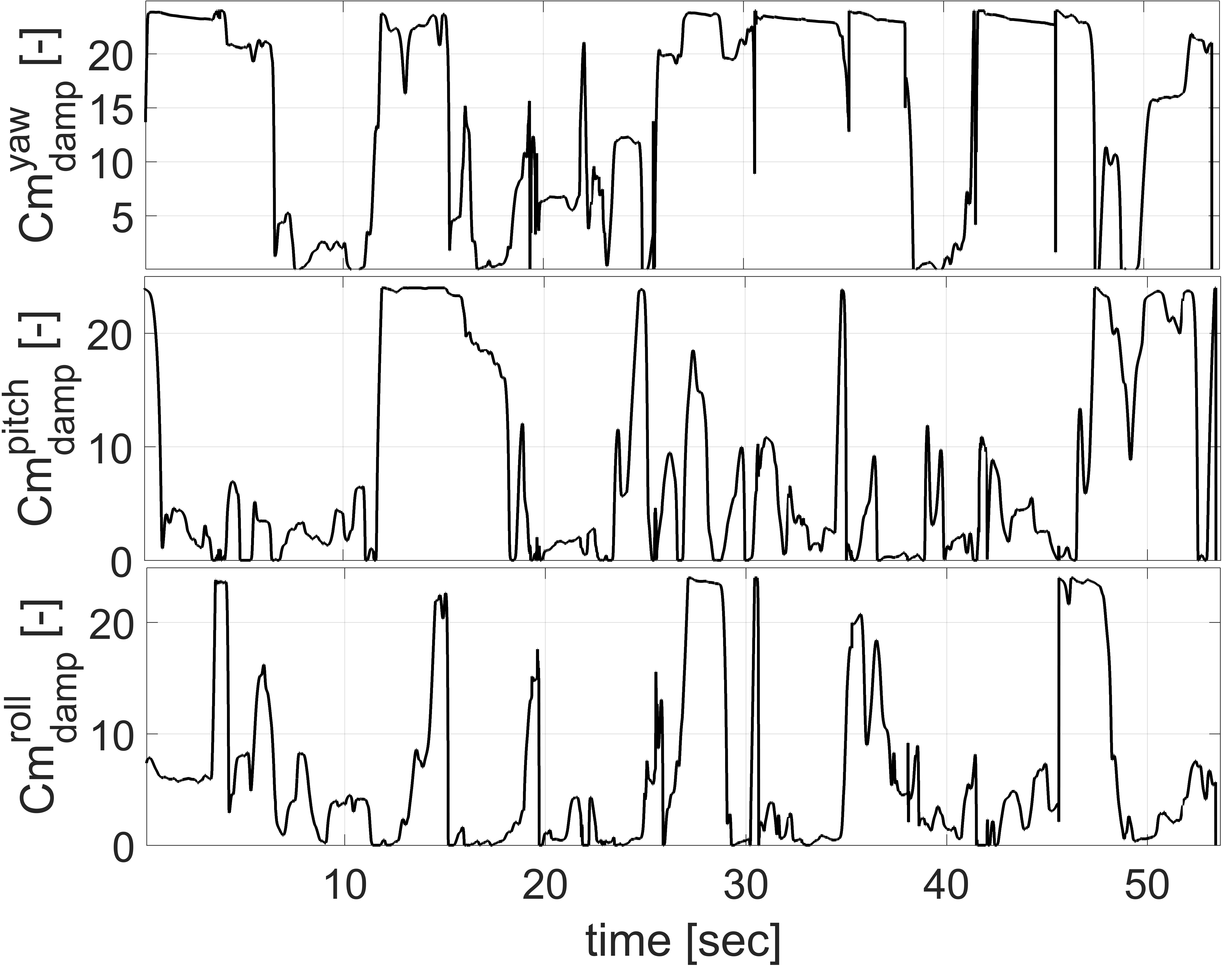}
    \caption{damping moment coefficients}
    \label{fig:trandamp}
    \end{subfigure}
    
    \caption{Skydiver inputs during the transition maneuvers, as estimated by the modified Unscented Kalman Filter. Damping moment coefficients are limited to $[0.01, 24]$.}
    \label{fig:trancoeffs}
\end{figure}

The reconstruction results of the transition maneuvers are summarized in Figures \ref{fig:tran}, \ref{fig:trancoeffs}. The results were obtained using the same skydiver's inputs model as for the tracking maneuver: Eq. \eqref{eq1}, \eqref{eq1_1}.  The measurement vector, however, did not include the horizontal velocity, since the GNSS was not available during this experiment. Thus, Eq. \eqref{eq:meas} was modified as follows:
\begin{equation} \label{eq:meas_flips}
    \vec{Z}_k=[heading, pitch, roll]^T_k, \quad k=1,...t_{end} \cdot 240
\end{equation}
the measurement dimension was $m=3$, and the noise covariance - a diagonal matrix $R$:  $R_{i,i}=[1.5,0.01,0.01]^2$, $i=1,..3$. Notice, that the heading measurement is not accurate without the aid of GNSS.

\begin{figure}[!htb]
    \centering
    \includegraphics[width=1\textwidth]{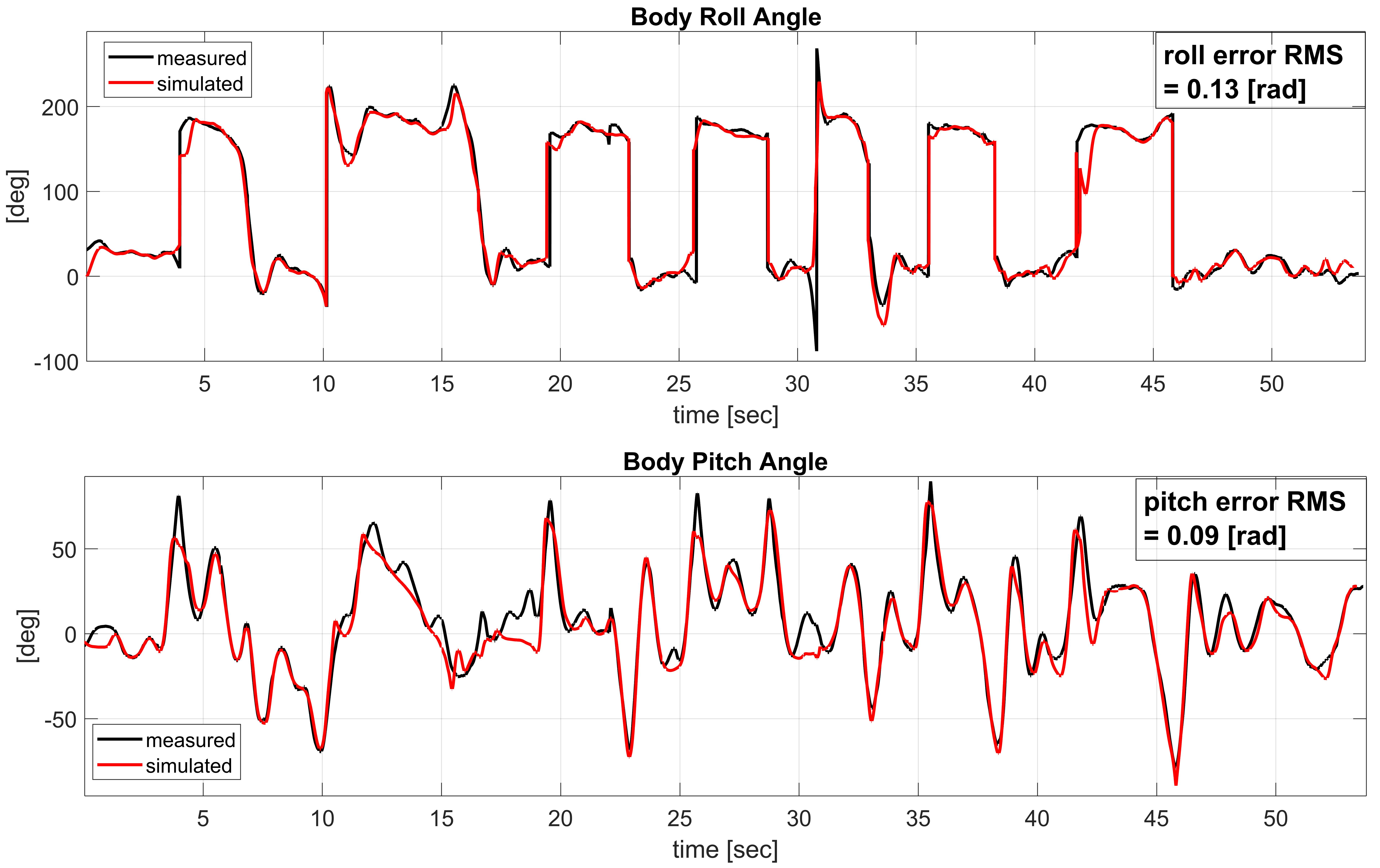}
    \caption{Reconstruction of the transition maneuvers in simulation. Damping moment coefficients are limited to $[-0.25, 24]$.}
    \label{fig:tran_im}
\end{figure} 

\begin{figure}[!htb]
    \centering
    \begin{subfigure}{.48\textwidth}
    \centering
    \includegraphics[width=0.98\textwidth]{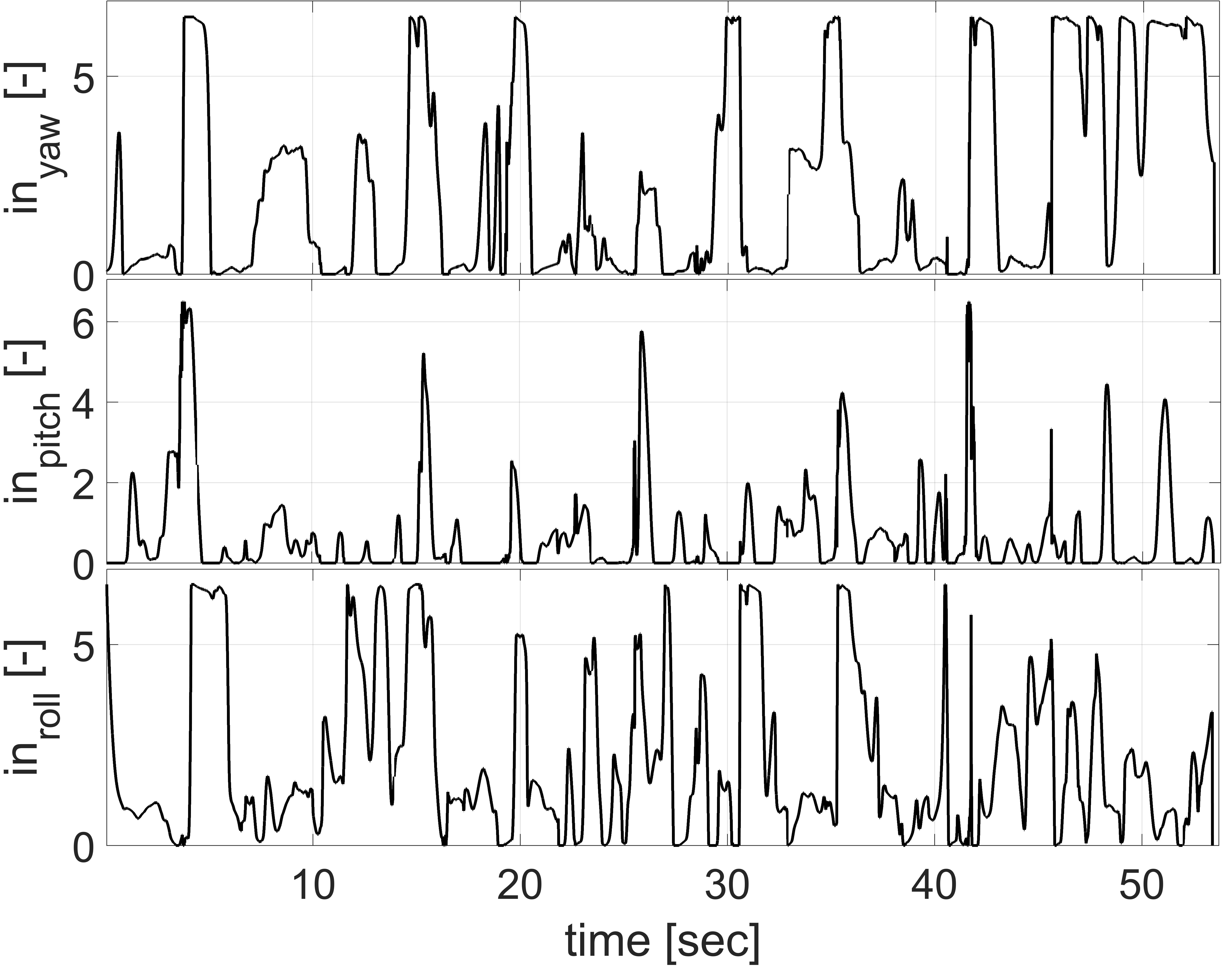}
    \caption{input moment coefficients}
    \label{fig:tranin_im}
\end{subfigure} 
\begin{subfigure}{.48\textwidth}
    \centering
    \includegraphics[width=0.98\textwidth]{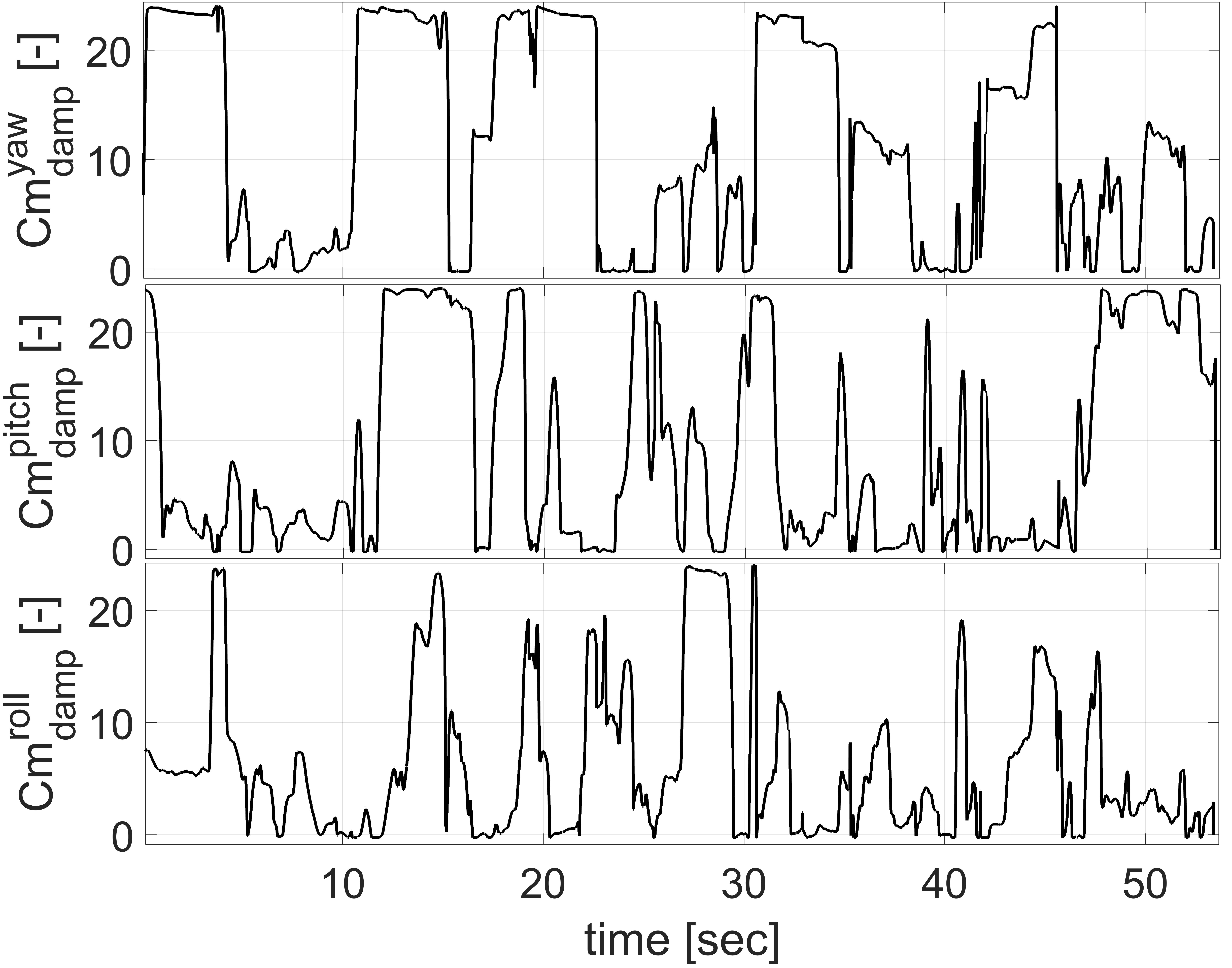}
    \caption{damping moment coefficients}
    \label{fig:trandamp_im}
    \end{subfigure}
    
    \caption{Skydiver inputs during the transition maneuvers, as estimated by the modified Unscented Kalman Filter. Damping moment coefficients are limited to $[-0.25, 24]$.}
    \label{fig:trancoeffs_im}
\end{figure}

Notice, that some of the transitions were reconstructed less accurately, for example the maneuver number 6, around 23 s (see Figure \ref{fig:tran}). It is possible to further improve the simulation accuracy by incorporating the following observation. 

Naturally, the skydiver's body resists the developing angular rates, adding the damping moment (notice the minus sign of the last term) to the moments equation Eq. \eqref{eq:sum_m}. 
However, during transition maneuvers the goal is the opposite: instead of decaying the oscillation it is desired to encourage it in order to flip/roll over. Thus, it is desired to produce a negative damping effect, also sometimes called \textit{negative resistance}, such as when playing musical instruments. For example, while bowing, the vibration of the string is increased instead of decreasing. This is achieved by musicians through finding by trial and error the right way to play the instrument. In skydiving the situation is similar: skydivers develop a way to relax and tense up certain muscles such that a negative damping is produced. 

This can be taken into account in our model by simply setting the lower bound for damping coefficients in Eq. \eqref{eq:minmax} to -0.25 instead of 0.01. The new results, obtained due to this change, are shown in Figures \ref{fig:tran_im},  \ref{fig:trancoeffs_im}. The improvement of reconstruction of the previously problematic maneuver number 6 is clearly seen. This maneuver started as a layout, however the skydiver felt an increasing instability and turned it into a barrel roll. Notice in Figure \ref{fig:trandamp_im} that during this transition (around 23 s) the pitch and yaw damping moment coefficients were estimated as negative values. The former may explain the feeling of layout going out of control, whereas the latter may be responsible for the fast initiation of a barrel roll. The video of the experiment with a superimposed simulation recording allows to visually compare the actual and reconstructed maneuvers, see \cite[Chapter  \textit{Reconstruction of the Transitions Maneuvers in the Skydiving Simulation}]{figshare}.     

It is interesting to compare the accuracy of maneuvers reconstruction with and without estimation of skydiver conscious inputs. Advanced maneuvers, such as angle flying and transitions described above, can not be reconstructed without estimating the input and damping moment coefficients as a function of time. However, for simple maneuvers in a belly-to-earth orientation, reconstructed in Section \ref{sec:model}, it was assumed that the input moment coefficients are zero and the damping moment coefficients have constant values. In Section \ref{sec:ukf} reconstruction of aerial rotations with and without estimation of the skydiver inputs is compared.

\subsection{Skydiver Conscious Inputs During Basic Maneuvers}\label{sec:ukf}
\begin{figure}[!htb]
    \includegraphics[width=1\textwidth]{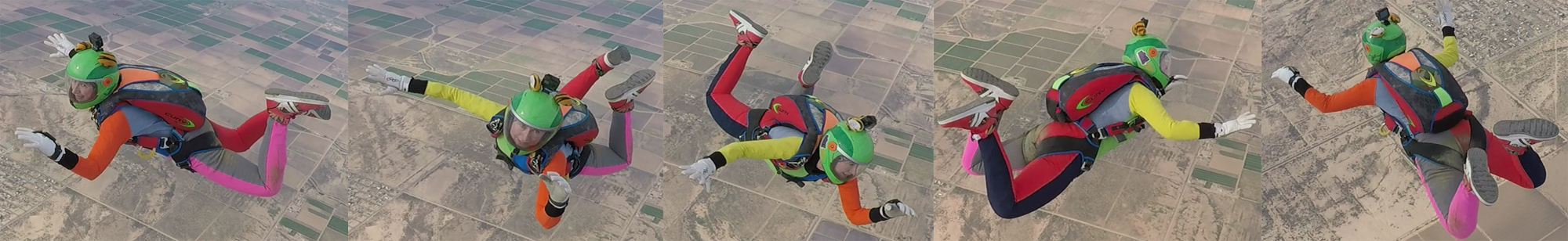}
    \caption{Snapshots of one left turn from the turning experiment video. Time sequence from left to right.}
    \label{fig:turnpic}
\end{figure} 

\begin{figure}[!htb]
    \centering
    \begin{subfigure}{.48\textwidth}
    \centering
    \includegraphics[width=0.98\textwidth]{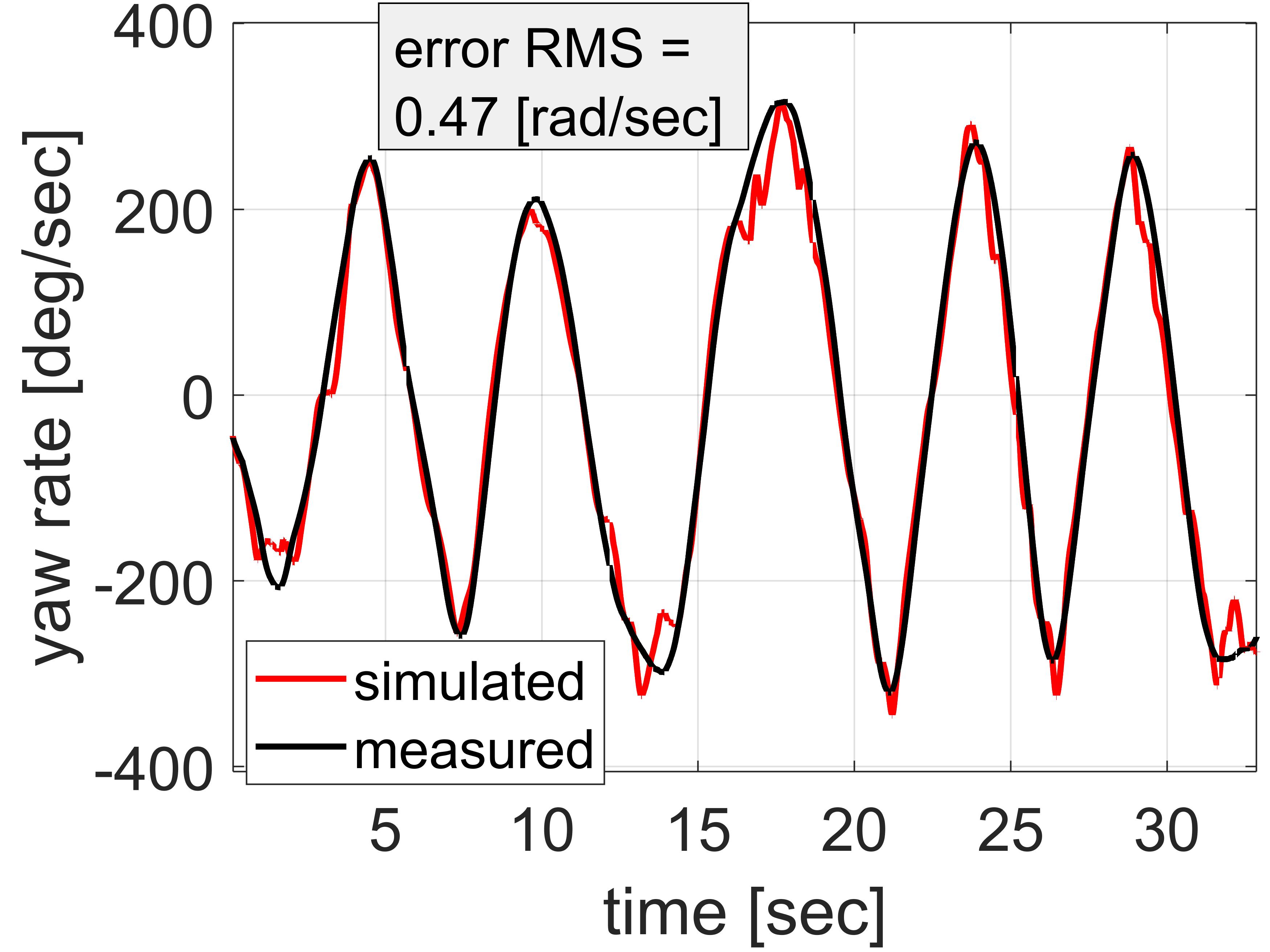}
    \caption{yaw rate}
    \label{fig:turn}
\end{subfigure} 
\begin{subfigure}{.48\textwidth}
    \centering
    \includegraphics[width=0.98\textwidth]{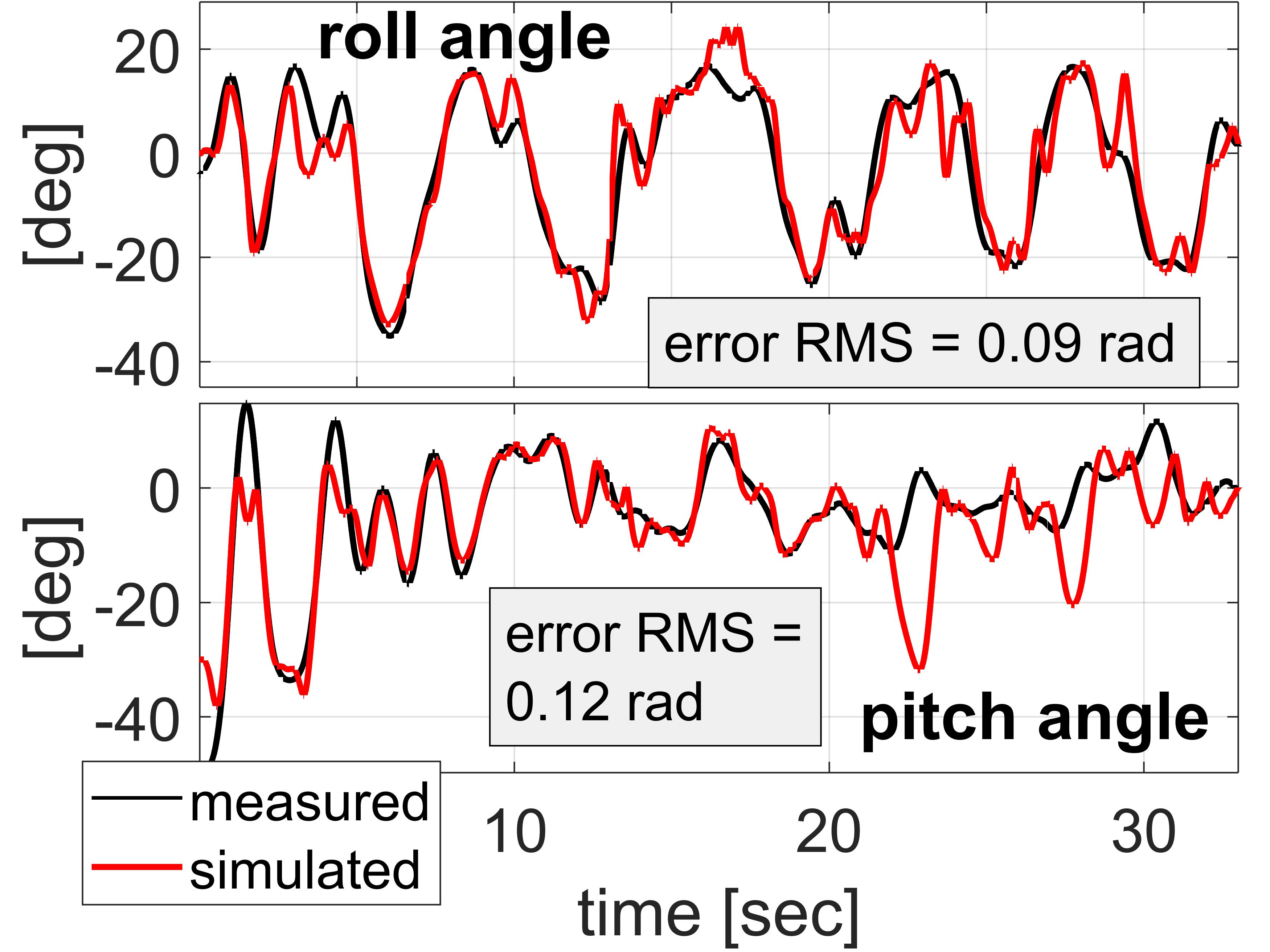}
    \caption{body orientation}
    \label{fig:turnpr}
    \end{subfigure}
    
    \caption{Reconstruction of the turning experiment in simulation.}
    \label{fig:turnypr}
\end{figure} 
The turning experiment, i.e. 360 degrees turns to the right and to the left, can be easily reconstructed in simulation if performed by an experienced skydiver. The  damping moment coefficients can be assumed constant and tuned in simulation and the input moment coefficients can be assumed zero, the reconstruction of the turning maneuver (the yaw rate) is still quite accurate, see Figure \ref{fig:taff_turn_noUKF}. The reason is that when skydivers experienced in belly-flying perform this maneuver the turn is initiated while their body is in the horizontal plane: The change in body posture produces aerodynamic forces in the body coronal plane that generate the yaw moment. As the turn progresses, the body develops roll and pitch angles due to evolving aerodynamic moments and input moments from the part of the skydiver. Therefore, as seen from Figure \ref{fig:taff_turn_noUKF},  yaw rate of the maneuver can be reconstructed without a precise estimation of body roll and pitch angles. Certainly, the body angles and the yaw rate can be estimated more accurately if the input moment and damping coefficients are estimated by the UKF, see Figures \ref{fig:taff_turn_UKF}, \ref{fig:taff_turncoeffs}. 

%perform this basic maneuver so smoothly, the limbs movements are small, and the muscles in all limbs remain relaxed, i.e. not generating an excessive pressure on the airflow. 

However, the turns can also be performed aggressively, at the edge of stability: The change in body posture and skydiver's input moments produce aerodynamic forces in the body sagittal and transverse planes, thus generating roll and pitch moments. The body  develops roll and pitch angles, what exposes the torso to the airflow and generates a yaw moment, initiating a fast and sometimes uncontrollable turn. In this case, it is not sufficient to assume zero input moment coefficients and constant damping moment coefficients for maneuver reconstruction. First, it is necessary to estimate these coefficients. Such experiment was conducted (see Figure \ref{fig:turnpic}) and reconstructed in simulation, see results in Figure \ref{fig:turnypr}.

\begin{figure}[!htb]
    \centering
    \begin{subfigure}{.48\textwidth}
    \centering
    \includegraphics[width=0.98\textwidth]{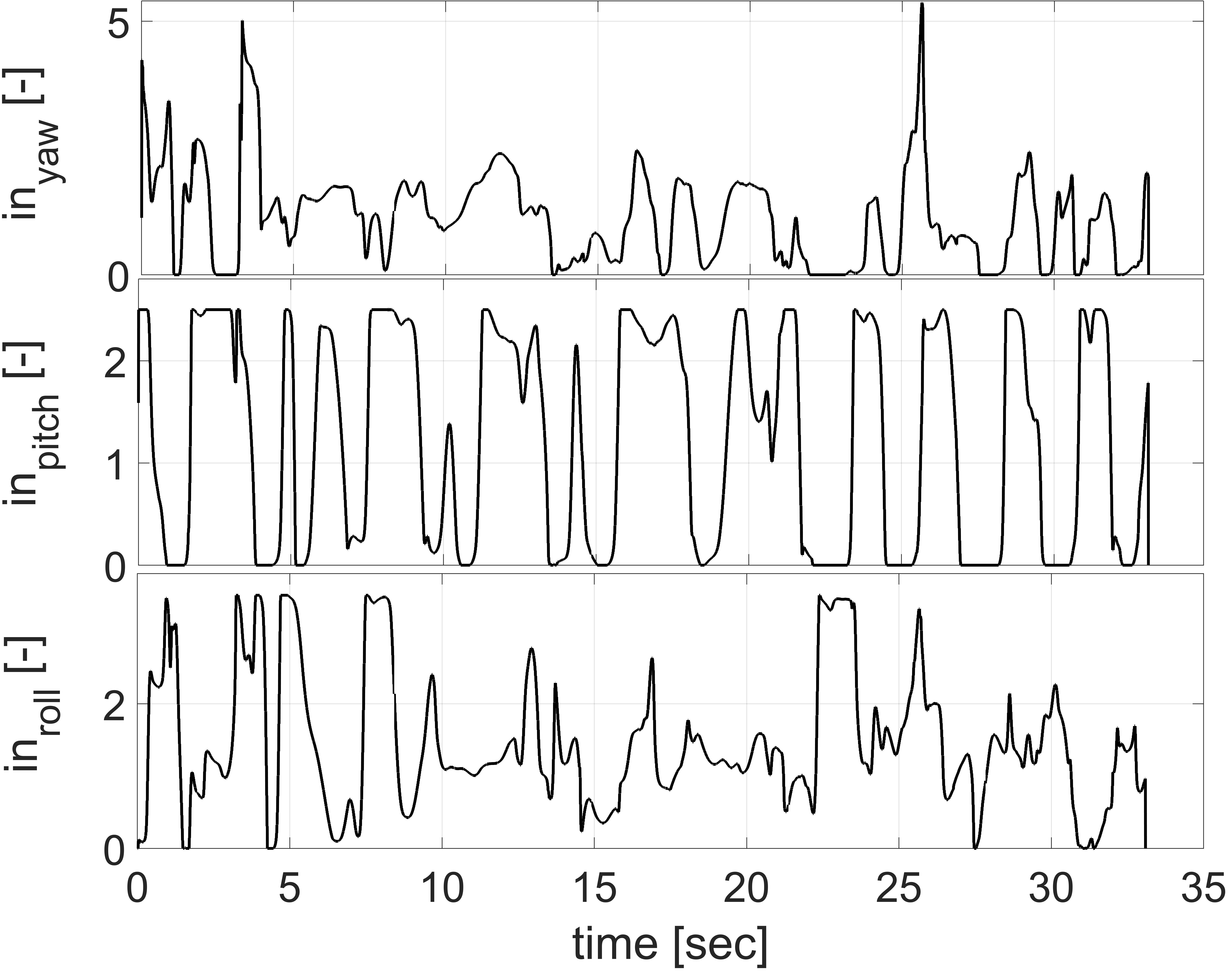}
    \caption{input moment coefficients}
    \label{fig:turnin}
\end{subfigure} 
\begin{subfigure}{.48\textwidth}
    \centering
    \includegraphics[width=0.98\textwidth]{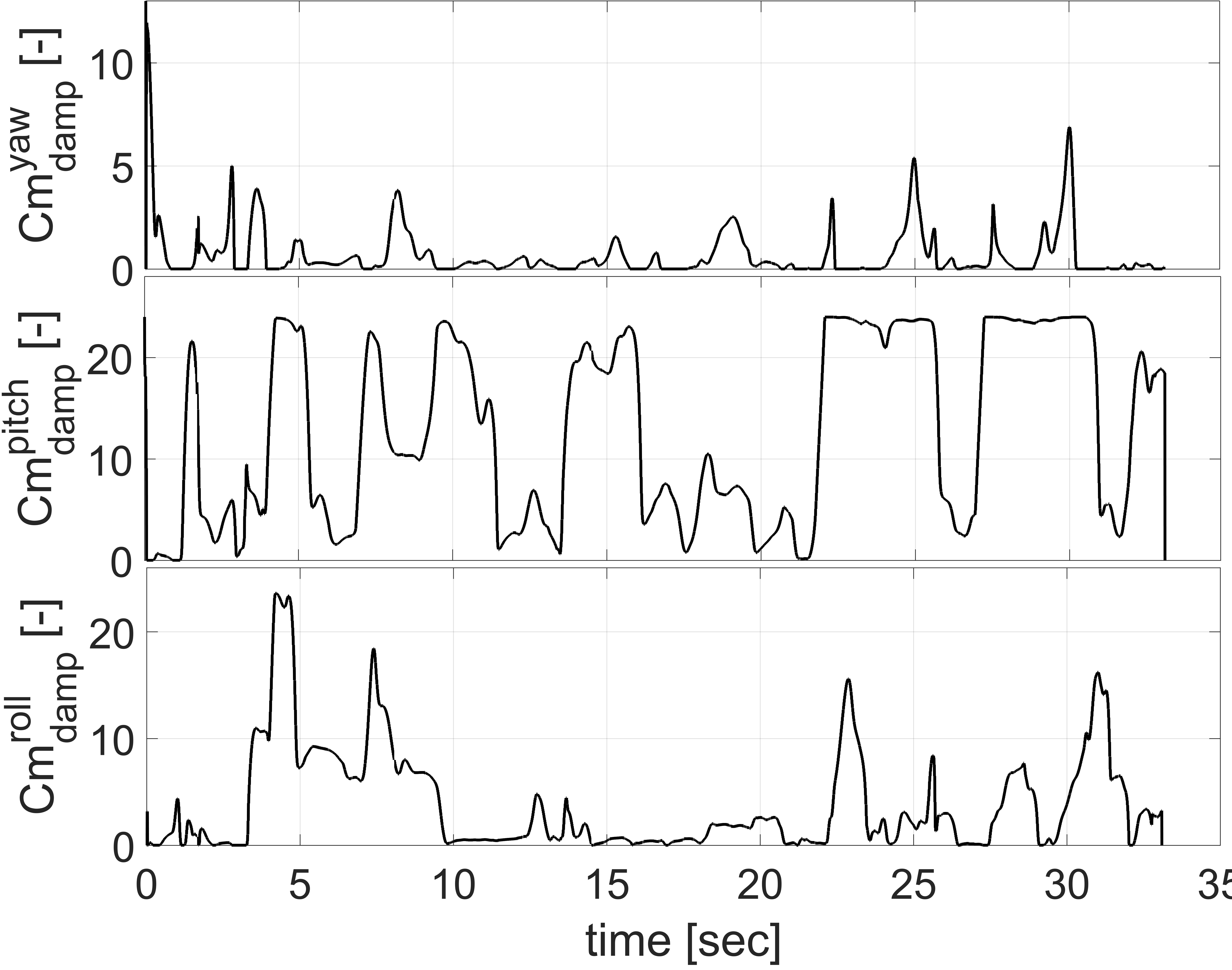}
    \caption{damping moment coefficients}
    \label{fig:turndamp}
    \end{subfigure}
    
    \caption{Skydiver inputs during the turning experiment, as estimated by the modified Unscented Kalman Filter.}
    \label{fig:turncoeffs}
\end{figure}

\begin{figure}[!htb]
    \centering
    \includegraphics[width=1\textwidth]{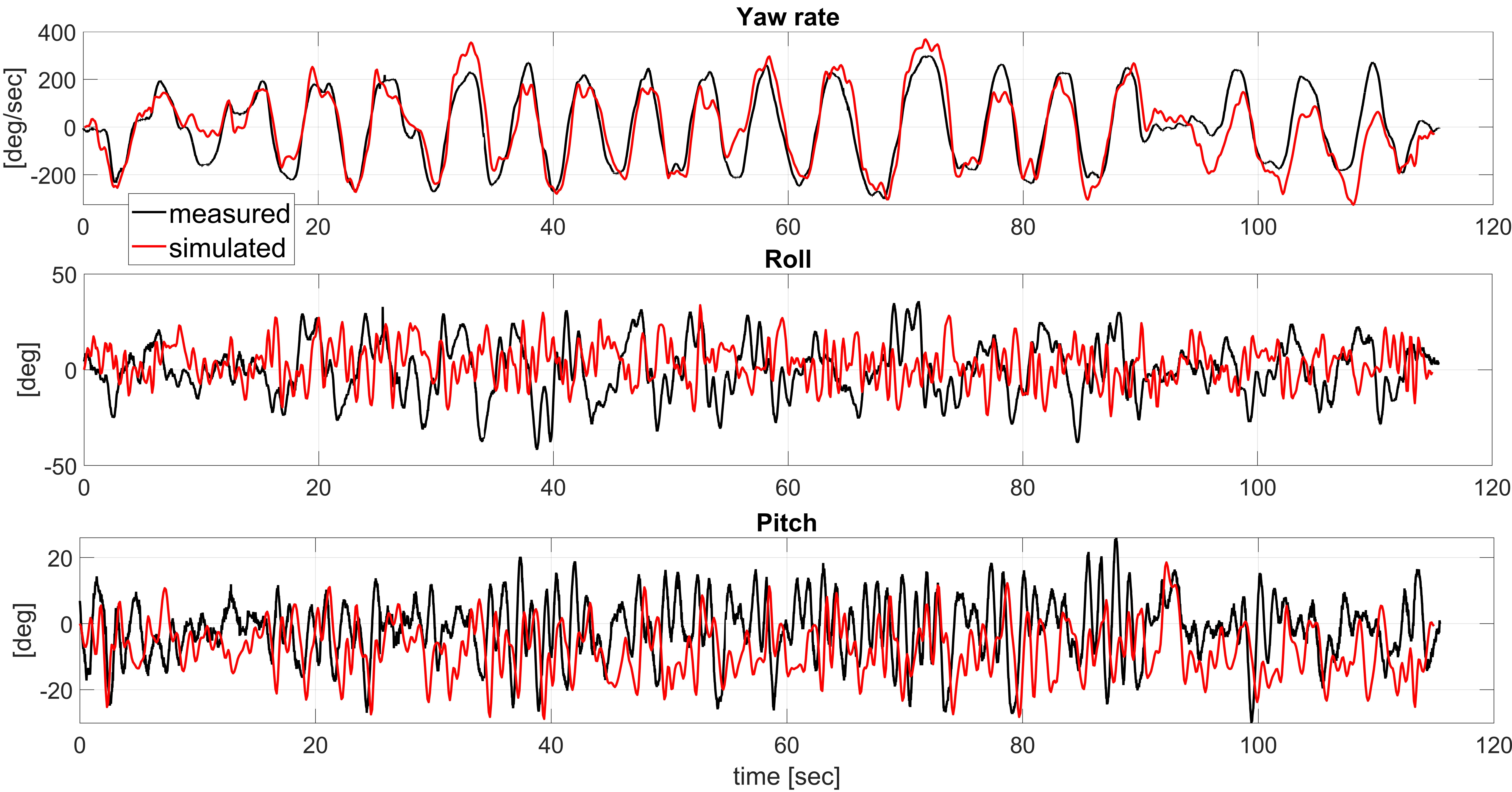}
    \caption{Reconstruction of the turning wind tunnel experiment in simulation. The turns were performed by a highly experienced skydiver, input moment coefficients were set to zero, and yaw, pitch, roll damping moment coefficients were set to 0.5, 3, 6, respectively.}
    \label{fig:taff_turn_noUKF}
\end{figure} 

\begin{figure}[!htb]
    \centering
    \includegraphics[width=1\textwidth]{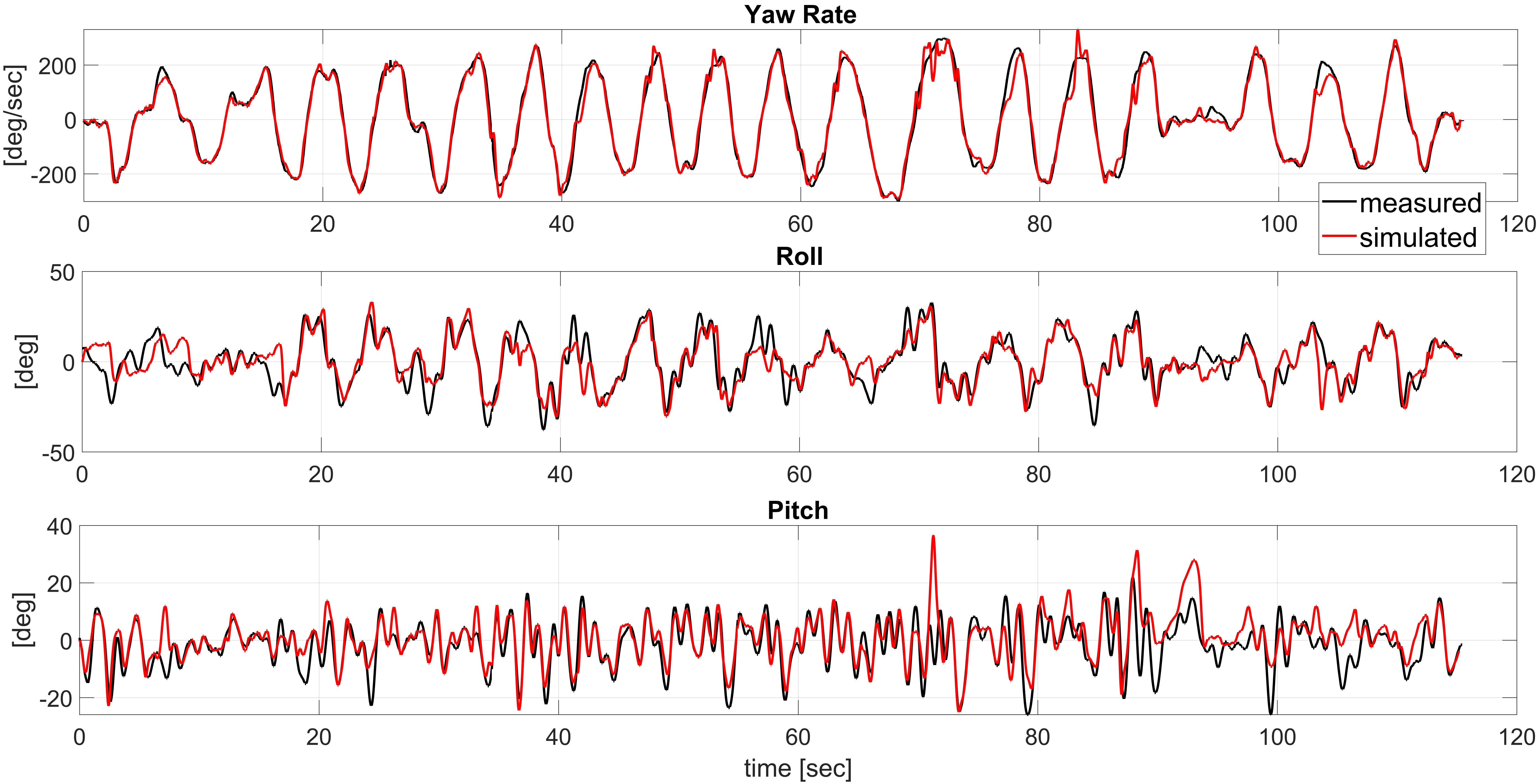}
    \caption{Reconstruction of the turning wind tunnel experiment in simulation. The turns were performed by a highly experienced skydiver, the input and damping moment coefficients were estimated by the modified Unscented Kalman Filter.}
    \label{fig:taff_turn_UKF}
\end{figure}

\begin{figure}[!htb]
    \centering
    \begin{subfigure}{.48\textwidth}
    \centering
    \includegraphics[width=0.98\textwidth]{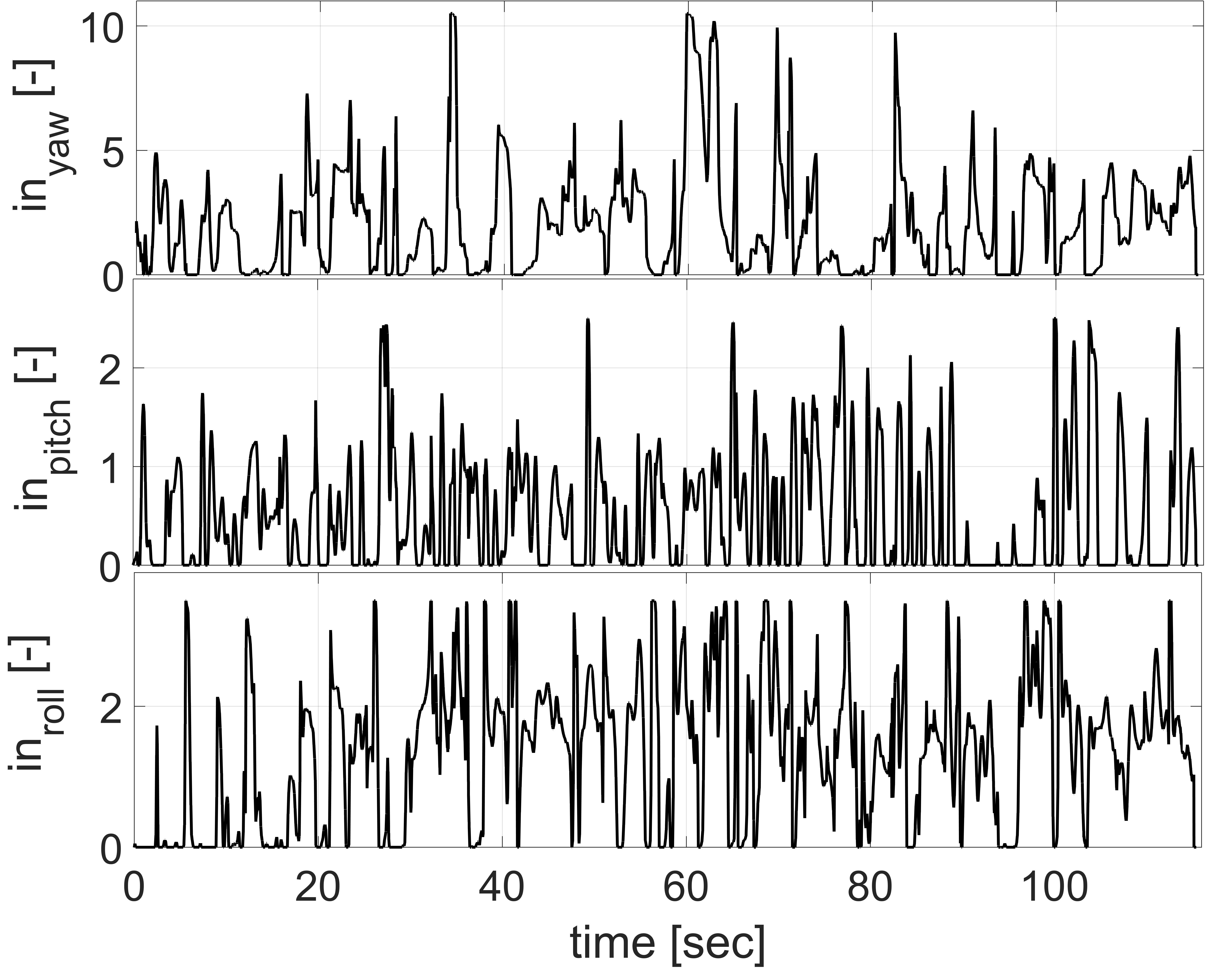}
    \caption{input moment coefficients}
   \label{fig:taff_turnin}
\end{subfigure} 
\begin{subfigure}{.48\textwidth}
    \centering
    \includegraphics[width=0.98\textwidth]{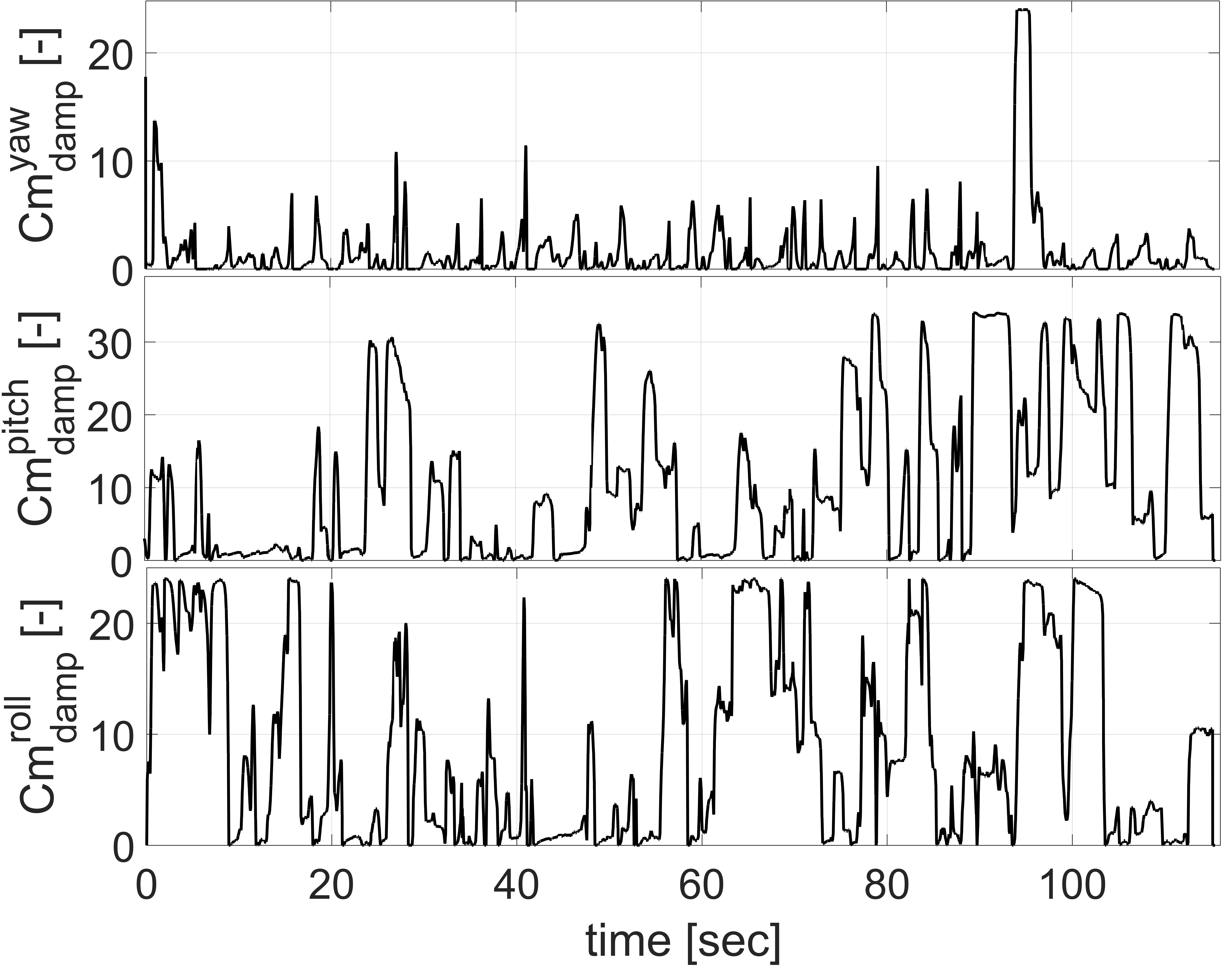}
    \caption{damping moment coefficients}
   \label{fig:taff_turndamp}
    \end{subfigure}
    
    \caption{Skydiver inputs during the wind tunnel turning experiment, performed by a highly experienced skydiver, as estimated by the modified Unscented Kalman Filter.}
    \label{fig:taff_turncoeffs}
\end{figure}

The measurement vector of the turning experiments included yaw rate (the smoothed derivative of the measured heading angle), thus Eq. \eqref{eq:meas} was modified as follows:
\begin{equation} \label{eq:meas_turns}
    \vec{Z}_k=[yaw \, rate, pitch, roll]^T_k, \quad k=1,...t_{end} \cdot 240
\end{equation}
the measurement dimension was $m=3$, and the noise covariance - a diagonal matrix $R$:  $R_{i,i}=[0.01,0.01,0.01]^2$, $i=1,..3$. Also, the upper bound of the state vector (Eq. \eqref{eq:minmax}) was adjusted to reflect the goal of the maneuver - turning: 
\begin{equation}
    \left(\vec{X_i}\right)_{max}=[10.5, 3.5, 2.5, 24, 24, 24]^T
\end{equation}
The aerodynamic model for the skydiver's input moments was the same as for back-to-earth tracking. 
\begin{figure}[!htb]
    \centering
    \includegraphics[width=0.8\textwidth]{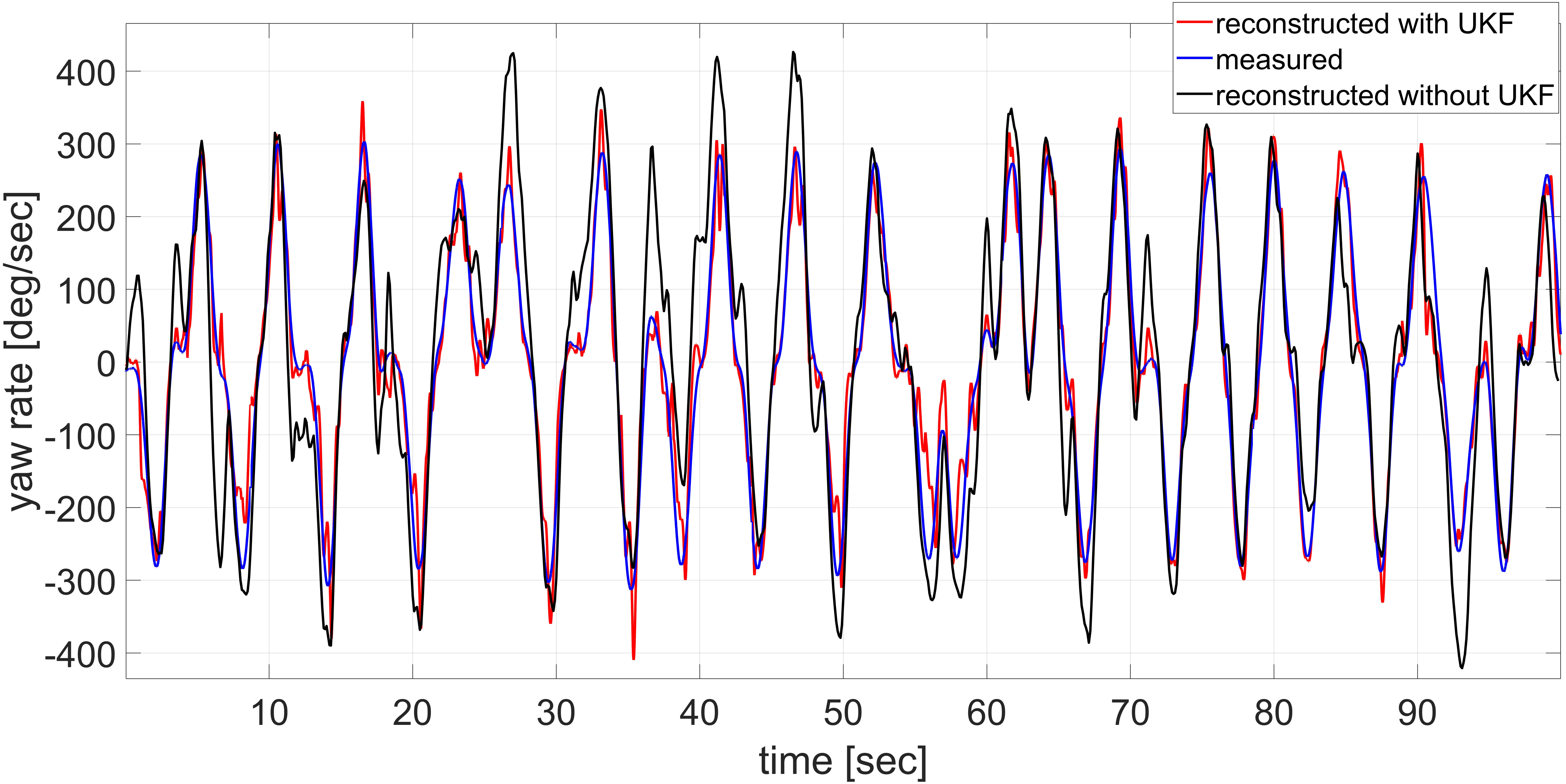}
    \caption{Reconstruction of the turning wind tunnel experiment in simulation. The turns were performed by the Elite Skydiver, the input and damping moment coefficients were: 1. estimated by the modified Unscented Kalman Filter (UKF); 2. input moment coefficients set to zero, and yaw, pitch, roll damping moment coefficients - set to 0.5, 3, 6, respectively.}
    \label{fig:jen_turn}
\end{figure}
\begin{figure}[!htb]
    \centering
    \begin{subfigure}{.48\textwidth}
    \centering
    \includegraphics[width=0.98\textwidth]{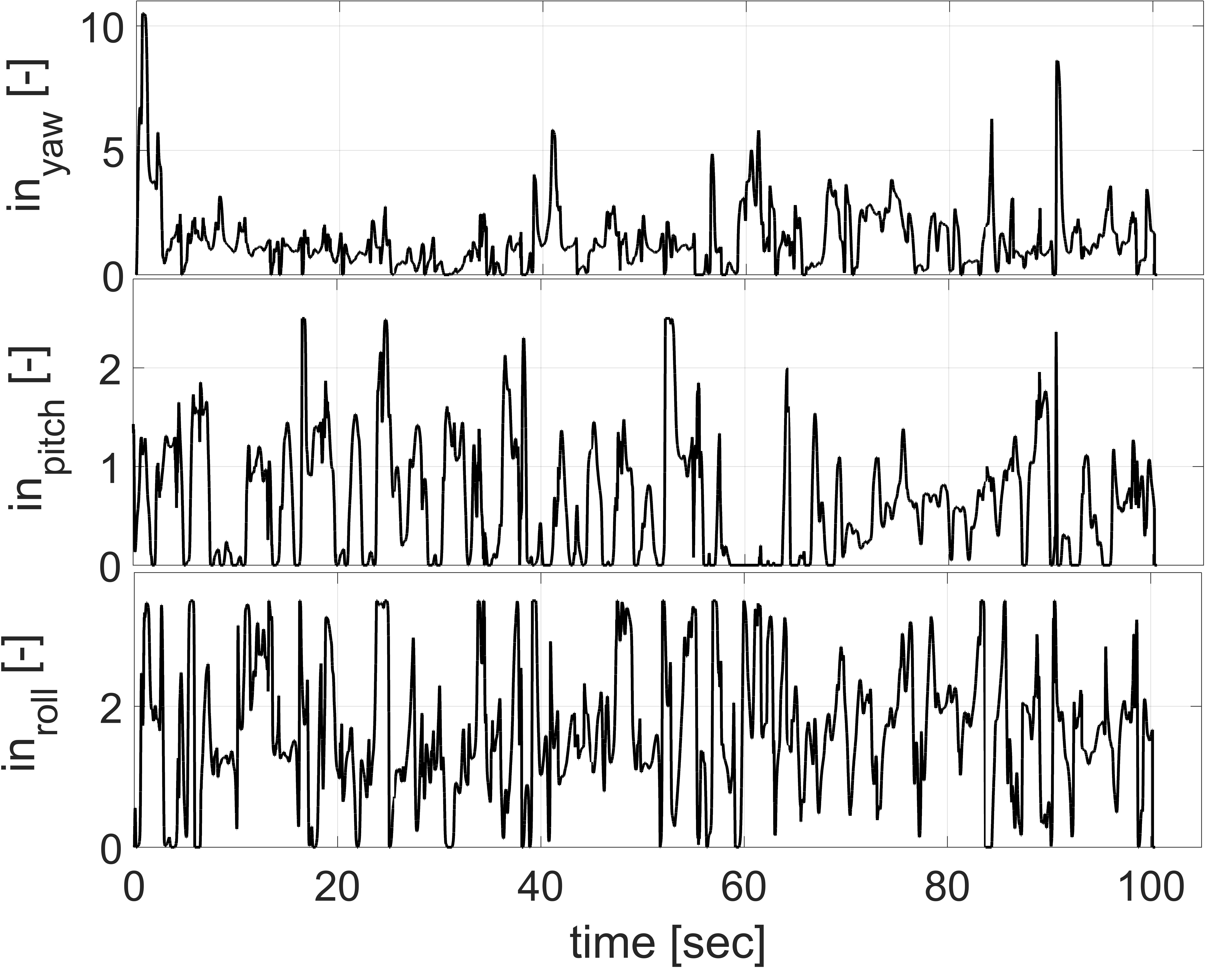}
    \caption{input moment coefficients}
   \label{fig:jen_turnin}
\end{subfigure} 
\begin{subfigure}{.48\textwidth}
    \centering
    \includegraphics[width=0.98\textwidth]{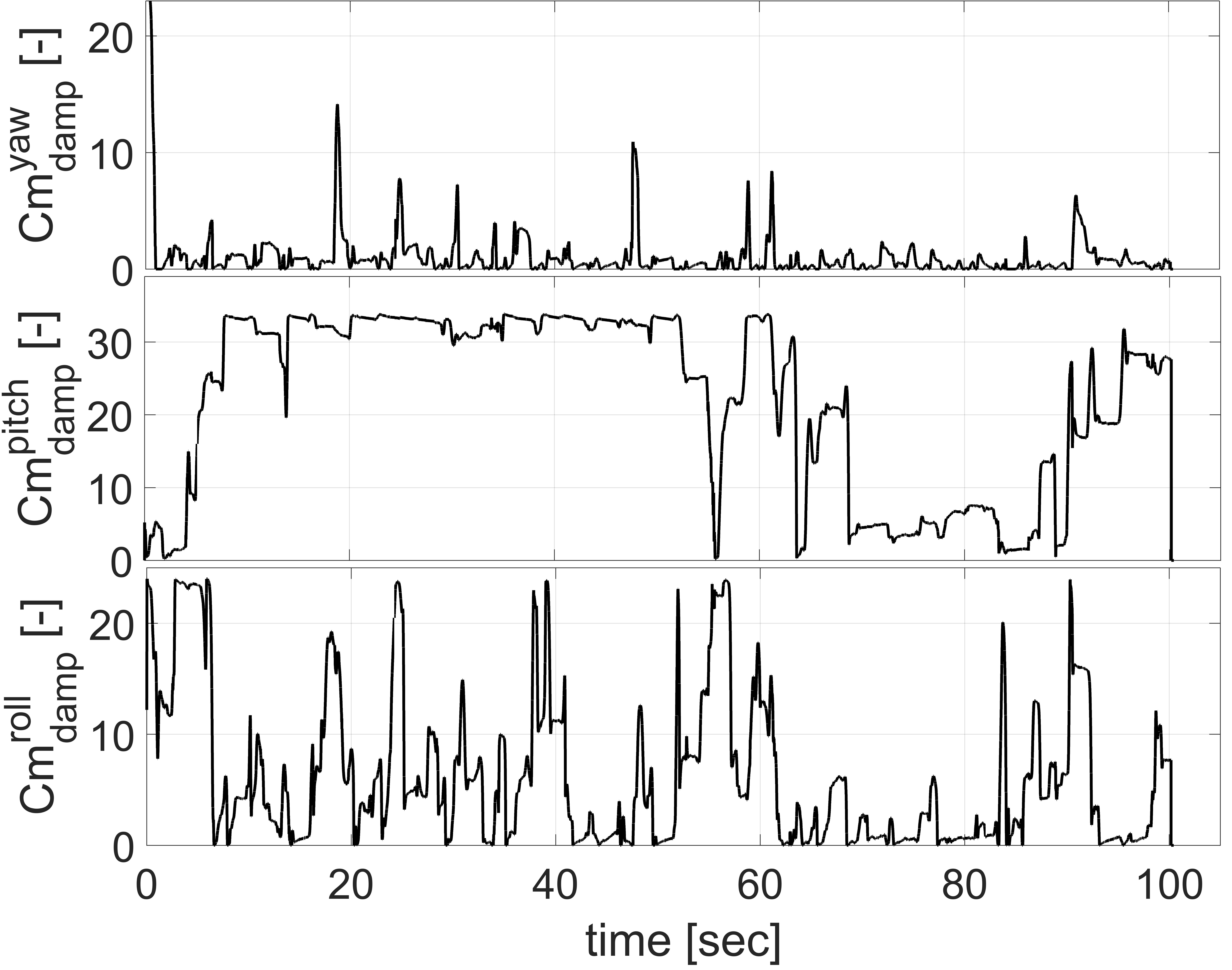}
    \caption{damping moment coefficients}
   \label{fig:jen_turndamp}
    \end{subfigure}
    
    \caption{Skydiver inputs during the wind tunnel turning experiment, performed by the Elite Skydiver, as estimated by the modified Unscented Kalman Filter.}
    \label{fig:jen_turncoeffs}
\end{figure}

For the experienced skydiver (shown on Figures \ref{fig:taff_turn_UKF}, \ref{fig:taff_turncoeffs}), however, the model was simplified since with experience less limbs are used to produce the desired inputs: only the most efficient ones. Thus, the legs, upper arms and forearms were excluded from the yaw moment equation; and the legs, hips, upper arms, and forearms were excluded from the roll moment equation (in Eq. \eqref{eq1}). Additionally, during belly-to-earth maneuvers the more experienced skydiver is more likely to engage the hips and legs for pitch moment inputs, as these surfaces are most aerodynamically efficient. Therefore, the pitch moment equation included hips and legs instead of head and thorax.  

It is interesting that when performing maneuvers in an RW suit (Relative Work suit with booties) moment inputs in all axes (roll, pitch, and yaw) are applied by the means of thighs and legs. This was found by estimating the input and damping moment coefficients via UKF (see Figure \ref{fig:jen_turncoeffs}) from the data recorded in the wind tunnel turning experiment performed by an RW competitor wearing such a suit. This result is expected, since the booties provide a large aerodynamic surface, which should become the major tool for skydivers specializing in the RW discipline. 

Figure \ref{fig:jen_turn} shows the comparison of turning reconstruction when assuming constant input and damping moment coefficients, and when estimating them via UKF. It can be clearly seen that some turns can not be truthfully reconstructed without dynamically estimating the user inputs.

\section{Insight into Implementation Mechanisms} \label{sec:advanced_maneuvers_mech}

For the practical application of the above findings it is necessary to understand how the human body actually produces certain damping moment coefficients, in particular the negative ones.
Keeping in mind that these coefficients are the representation of the unmodeled internal body biomechanics, the following variables could be involved in forming the damping, i.e. resistance to angular rates:
\begin{enumerate}
    \item \textbf{Flexibility of some segments modeled as rigid}, e.g. thorax - the chest can be rolled inwards or widely open without significantly changing the state of the thorax-abdomen or shoulders joints. This small chest movement may have a very large influence on the airflow since it causes qualitative changes of the flow patterns, like arching. Another example is the hands - can change width, length, and even shape if closed into fists.
    \item \textbf{True shape of segments modeled as truncated cones}: Forearms, upper arms, legs and thighs are modeled as truncated cones, meaning identical projections into a sagittal and coronal planes. However, thighs and legs, for example, are usually wider in the sagittal plane. This fact is widely used in practicing sit flying and layouts: when a more efficient input from the legs is desired they are turned with the wider side towards the airflow.
    \item \textbf{Variability in shape and volume of limbs}: The model assumes a constant shape and volume of all segments, which, however, can fluctuate depending on muscle tension. For example, changing the shape of the abdomen area can have a similar affect to arching.
    \item \textbf{Muscle tension}. It is known from empirical evidence that extensive muscle tension can influence the patterns of the airflow around the body: The patterns become less optimal, producing more turbulence and thus parasitic drag. This creates unexpected aerodynamic forces and moments in different directions. That's the reason that skydiving students are instructed to relax the body tension as much as possible. Letting the airflow slightly move the limbs (in amplitudes below the noise level of the inertial sensors) allows a more optimal arrangement of the airflow patterns, mitigating the natural bilateral asymmetry of the human body.
    \item \textbf{Stretching}: An ability of the human body to stretch is not modeled, whereas, it may be involved in flying some advanced poses. In particular, it seems that achieving stability in head-up and head-down poses requires isometric stretching with flexion proportional to the transverse loading coming from the airflow.  
\end{enumerate}

Considering the above features of human biomechanics, we can make the following hypothesis, which should be experimentally verified or rejected:

The easiest way to control damping is by the means of tensing and relaxing the muscles. Tense body produces less damping, relaxed body - more damping. Thus, to stop undesirable rotations it is sufficient, in most cases, to relax the muscle tension. 

Negative damping that encourages the oscillations and is used for performing e.g. layouts, which require to bypass a stable equilibrium in the vertical orientation, can be achieved in the following way. A skydiver can roll in the chest around the frontal axis (making it produce less drag, i.e. fall faster) and roll the legs around the longitudinal axis exposing the wider surface to the airflow (producing more drag). This is true for both possible initial orientations: back-to-earth and belly-to-earth. 

When the rotation begins the body can be stretched and tensed up to keep the damping low through the transition. After the flip occurs and the rotation should be stopped, it is possible to relax the body to increase the damping, and, once again, roll the legs around the longitudinal axis such that a wider surface is exposed to the airflow to produce more drag in the lower body, and thus the pitching moment in the opposite direction. 

\subsection{Reconstruction of layouts in simulation}
\label{sec:layouts_rec}
\begin{figure}[htb]
    \centering
    \includegraphics[width=1\textwidth]{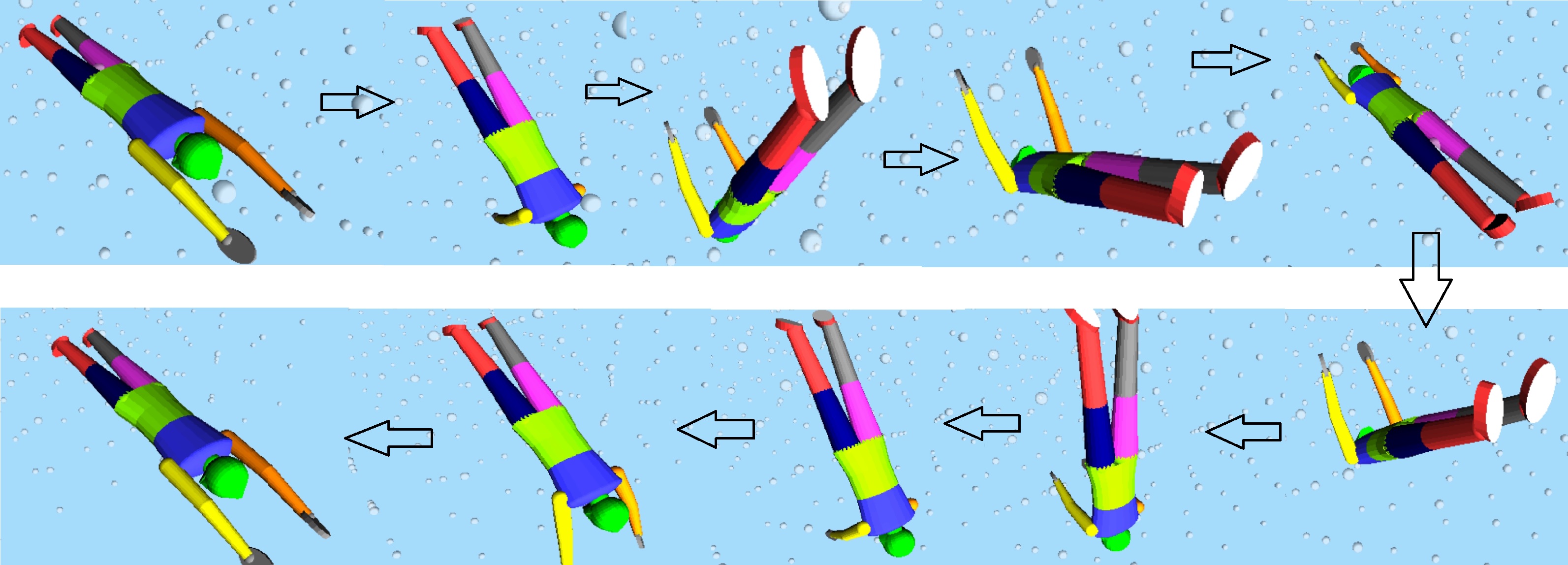}
    \caption{Snapshots of simulation of front and back layouts.}
    \label{fig:layouts_snap}
\end{figure} 

\begin{figure}[htb]
    \centering
    \includegraphics[width=0.9\textwidth]{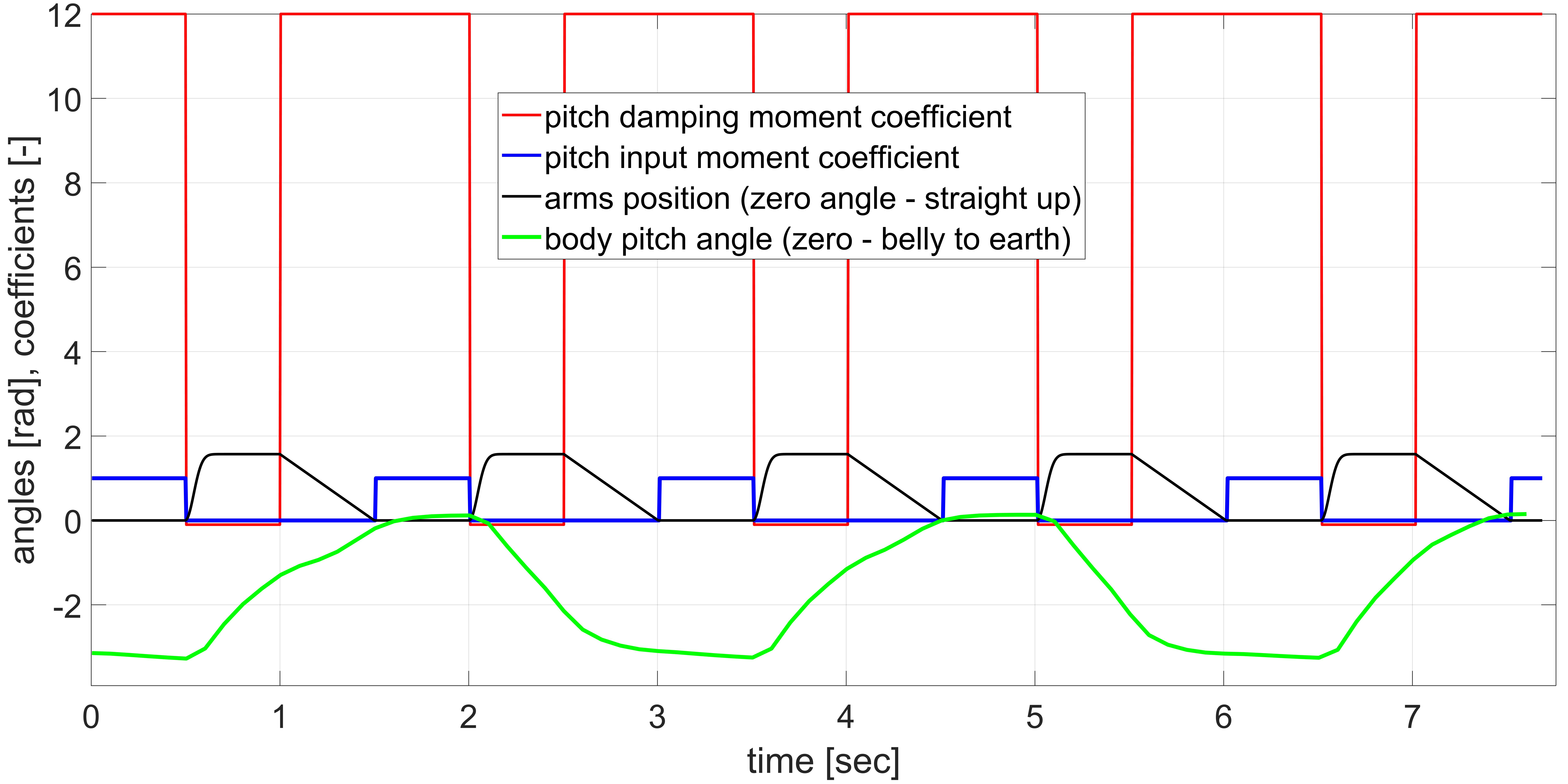}
    \caption{Simulation of front and back layouts.}
    \label{fig:layouts}
\end{figure}

Figure \ref{fig:layouts_snap} shows in snap shots how front and back layouts happen in simulation, while the inputs were chosen to be as simple as possible, in order to reveal the minimum changes that allow to perform layouts:
\begin{itemize}
    \item The default body posture was chosen completely straight (all DOFs equal zero angles). Only arms can move between the default position and being perpendicular to the torso.
    \item The pitch input moment coefficient can be either 1 or 0
    \item The pitch damping moment coefficient can be either 12 or -0.1
\end{itemize}
With the correct timing (shown in Figure \ref{fig:layouts}), these three inputs were sufficient to produce front and back layout maneuvers. 
The execution can be described by the four steps:
\begin{enumerate}
    \item \textbf{Preparation:} Applying the pitch input moment via the legs and thighs, while keeping the pitch damping high so that the movement in the opposite direction to the desired is very small: the body stays nearly horizontal.
    \item \textbf{Onset:} Releasing the pitch input and moving the arms to the position where they are perpendicular to the torso. This triggers the pitch movement.
    \item \textbf{Transition:} Keeping the pitch damping negative to increase the amplitude of the movement in order to flip over.
    \item \textbf{Stopping:} Increasing the pitch damping. The arms can be slowly returned to a default position in order to prepare for the next layout.
\end{enumerate}
%Considering the discussion in the previous subsection, the simplest execution of layouts can be 'translated' into coaching instructions as follows:
%\begin{enumerate}
    %\item \textbf{Preparation:} Relax, apply lateral rotation to the hips joints, and press on the airflow with thighs and legs.
   % \item \textbf{Onset:} Release the pressure and move the arms in front of the torso, letting the chest roll inwards.
   % \item \textbf{Transition:} Stretch the body and tense up until the flipping over occurs.
    %\item \textbf{Stopping:} Relax, apply lateral rotation to the hips joints.
%\end{enumerate}

\subsection{Pitch and Roll oscillations experienced by novices}

Any asymmetrical change of body posture produces aerodynamic moments that rotate the body into a different inertial orientation. If the damping moment  is large - the body will slowly transition into a new equilibrium position, if the damping moment is small - oscillations will develop. Depending on the damping moment value and the initial posture their amplitude will decay, stay constant, or increase, i.e. the new asymmetrical body posture will result into a new stable equilibrium, a limit cycle, or become unstable.

Novice skydivers often experience oscillations in pitch in a belly-to-earth pose. Such oscillations are readily reconstructed in simulation: starting from a stable neutral posture and small pitch damping, moving the arms downwards (using the shoulders internal rotation) onsets the oscillations. At a certain shoulders' joint angle a Hopf bifurcation occurs: the oscillations have a constant amplitude. The frequency of oscillations depends on the initial neutral posture and body parameters: the less body area exposed to the airflow - the higher the oscillations frequency.  

The oscillations in roll are reconstructed in a similar way: one arm and one leg on the same side are moved downwards. If the roll damping moment is sufficiently small - roll oscillations develop. 

Practically, it is impossible to maintain a perfectly symmetrical posture. Therefore, a novice skydiver will necessarily oscillate if his body is tensed up, since this decreases the damping moment. The 'key' to a stable belly-to-earth pose is relaxing the whole body, while the lumbar spine is extended to provide the arch.

\section{Discussion} \label{sec:discussion}
This paper presents the complete and comprehensive development of a dynamic model of a skydiver in free fall, substantially extending the state of the art in    \cite{skydive_dietz2011cfd, skydive_myers2009FAST, skydive_sit, skydive_cardona2011measurement, bsd_kwon1994kwon3d}. The model covers the movement of the skydiver around stable equilibria, as well as unstable transitions, and high amplitude maneuvers.  The input to the model is the skydiver posture as a function of time, i.e. position and orientation of the modeled limbs, and the skydiver conscious control input  i.e. muscle tensions and pressure on the airflow, represented in a simplified manner by damping moment and input moment coefficients, as a function of time. The model output is the skydiver position,   and orientation in 3D space, as well as inertial linear and angular velocities, all as a function of time.

Aerodynamic and other model parameters were tuned, {\it inter alia} depending on the body size and mass of the modeled skydiver. The model was successfully validated by comparing with measurements on real skydivers performing maneuvers in wind tunnels and in free fall. The posture and position/velocity/orientation measurements were taken from sensors on the Xsens Technologies  motion capture suit \cite{sensor_roetenberg2009xsens} that the skydivers were wearing. Depending on maneuver, the conscious control inputs were either tuned to constants, or estimated from the measurements and the postulated skydiver model using an Unscented Kalman Filter modified for this purpose.

The goodness of fit between measurements and model outputs can be appreciated from the following: The Xsens accuracy of dominant joint angles (whose maximum range is less than 180 deg) is less than 5 deg RMS, and the accuracy of the Xsens inertial measurements is, for body roll and pitch angles 0.01-0.1 rad RMS, for angular velocity 0.01-0.05 rad/s RMS, for horizontal linear velocity (with integrated GNSS) 0.1 m/s RMS,  and for vertical linear velocity 1 m/s RMS;  the differences between the Xsens measurements, and the skydiver model inertial orientation and velocities are, for body roll and pitch angles 0.05-0.15 rad RMS, for angular velocity 0.15 rad/s RMS, for horizontal linear velocity 0.45 m/s RMS, and for  vertical linear velocity 1.5 m/s RMS, with the amplitudes of body angles, angular velocity, and horizontal and vertical linear velocity being  3 rad, 7 rad/s, 15 m/s, and 65 m/s, respectively. The RMS values of the difference between the Xsens measurements and the skydiver model outputs are reported in the relevant figures.

Using known aerodynamic principles, previous research mentioned just above, and a profound understanding of the various skydiving disciplines, an attempt was made to construct an as simple as possible model, that sufficiently faithfully matched the measurements. An indication of the possible parsimony of the model is that  the moment coefficients that initially were tuned to be constant had to be estimated as time-varying in order to model advanced maneuvers. However, parsimony should be studied further. E.g. could certain limbs be eliminated, or do more limbs have to be included? Should the limbs be modeled differently from rigid truncated cones  (Section~\ref{sec:advanced_maneuvers_mech})?  Sensitivity to some parameters was investigated, mainly qualitatively (Section~\ref{sec:aero_params}), but a comprehensive study of parameter sensitivity should be undertaken, see e.g. \cite{est_ioslovich2010dominantpar}.

 The experiments reported in this paper made it possible also to gain  novel insight into the possible mechanisms of performing advanced skydiving maneuvers, and to appreciate the final stage of skill acquisition, defined by Bernstein  \cite{sport_bernstein1967co} as: "\textit{Exploiting the environmental dynamics to the full extent}". We may hypothesize that it means, in the context of skydiving, 
maintaining postures that create an \textit{opportunity} for the maximum pressure input, which can be consciously applied to the airflow. The reason is that the pressure on the airflow, as well as muscle tension, proved to be suited for control purposes, since they can be very efficient even when binary, as was shown in Section \ref{sec:layouts_rec}. 
   Simulations of layouts - one of the most advanced elements of body flight -  show that the three skydiver inputs (i)  postural change, (ii) muscle tension, and (iii) pressure on the airflow 
    have to be resolved as a synergy. Each of these inputs may have a simple activation profile, but as a synergy they provide  outstanding maneuverability.

Our skydiver model can be seen as  a virtual skydiver: an avatar in a computer simulation that 'learns' to move in free-fall, performs maneuvers, improves its technique, and develops a movement repertoire. 
In \cite{clarkeGutman2017, MEDarticle} feedback controllers for an 'autonomous' skydiver are reported. In \cite{Qlearning} an optimal body actuation strategy (\textit{movement pattern}) is derived utilizing reinforcement learning. In \cite{Clarke:2021}, the skydiver model dynamics is analyzed as a function of a given movement pattern. Also in \cite{Clarke:2021},  the skydiver model is suggested to be the heart of a Kinesthetic Training Module (KTM) as a tool for skydiver instructors to teach and train skydivers at various skill levels.
Hence, in addition to the refining of the skydiver model, further work may include building and testing a KTM prototype, \cite{clarke2018computerized}, and to apply the skydiver modeling principles to other sports.  

\section*{Declaration of competing interest}
The authors declare that they have no known competing financial interests or personal relationships that could have appeared to influence the work reported in this paper. The authors declare that they have a relevant US Patent Application US20180025664A1 \citep{clarke2018computerized}.
 \bibliographystyle{elsarticle-num} 
 \bibliography{cas-refs}





% \end{thebibliography}
\end{document}